\documentclass[11pt]{article}

\usepackage[preprint]{acl}

\usepackage{times}
\usepackage{latexsym}

\usepackage[T1]{fontenc}

\usepackage[utf8]{inputenc}

\usepackage{microtype}

\usepackage{inconsolata}

\usepackage{graphicx}

\usepackage[utf8]{inputenc} 
\usepackage[T1]{fontenc}    
\usepackage{hyperref}       
\usepackage{url}            
\usepackage{booktabs}       
\usepackage{amsfonts}       
\usepackage{mathtools}

\usepackage{nicefrac}       
\usepackage{microtype}      

\usepackage{colortbl}
\usepackage{amsmath,amssymb,amsthm}
\usepackage{float}
\usepackage{enumitem}
\usepackage[dvipsnames]{xcolor}
\usepackage{inconsolata}
\usepackage{multirow}
\usepackage{longtable}
\usepackage{stfloats}
\usepackage{placeins}
\usepackage{subcaption}
\usepackage{wrapfig}
\usepackage{soul}
\usepackage{verbatim}
\usepackage{bm}
\usepackage{epsf}
\usepackage{dsfont}
\usepackage{subcaption}
\usepackage[export]{adjustbox}
\usepackage{tablefootnote}
\usepackage{tabularx}

\usepackage{times}
\usepackage{latexsym}
\usepackage[most]{tcolorbox}

\theoremstyle{plain}

\theoremstyle{remark}

\usepackage[most]{tcolorbox}

\tcbset{
  mytheoremstyle/.style={
    colback=green!5!white,
    colframe=green!70!black,
    fonttitle=\bfseries,
    coltitle=black,
    boxrule=0.8pt,
    arc=4pt,
    left=4pt,
    right=4pt,
    top=4pt,
    bottom=4pt,
    before skip=10pt,
    after skip=10pt,
  }
}

\makeatletter
\newenvironment{theorembox}[1][]{%
  \refstepcounter{theorem}
  \begin{tcolorbox}[mytheoremstyle,
    title={Theorem~\thetheorem\if\relax\detokenize{#1}\relax\else~(\textnormal{#1})\fi}]
  \itshape
}{%
  \end{tcolorbox}
}

\makeatother

\title{Quantize What Counts: More for Keys, Less for Values}

\author{
  Mohsen Hariri$^{1}$, 
  Alan Luo$^{1}$, 
  Weicong Chen$^{1}$, 
  Shaochen Zhong$^{2}$, 
  Tianyi Zhang$^{2}$, 
  Qifan Wang$^{3}$, \\
  \textbf{Xia Hu}$^{2}$, 
  \textbf{Xiaotian Han}$^{1}$, 
  \textbf{Vipin Chaudhary}$^{1}$ \\[1ex]
  ${}^{1}$Case Western Reserve University; 
  ${}^{2}$Rice University; 
  ${}^{3}$Meta \\[1ex]
  \texttt{mohsen.hariri@case.edu}
}

\def\mn{{\em Key-Prioritized Quantization}}
\def\op{{\em Key-Value Norm Disparity}}

\begin{document}
\maketitle

\begin{abstract}
Large Language Models (LLMs) suffer inference-time memory bottlenecks dominated by the attention Key-Value (KV) cache, which scales with model size and context length. While KV-cache quantization alleviates this cost, bit allocation between keys and values is often tuned heuristically, lacking theoretical grounding and generalizability. 
This paper proposes two theorems that anchor mixed-precision KV quantization in the intrinsic geometry of Transformer models. First, key projections systematically have larger spectral and Frobenius norms than value matrices, implying higher information density along the key path. Second, for any given memory budget, prioritizing precision for keys over values strictly reduces quantization error and better preserves accuracy. Empirical evaluations across various prominent LLMs and benchmarks show that key-favored allocations (e.g., 4-bit keys, 2-bit values) retain up to 98.3\% accuracy compared to uniform allocations (e.g., 4-bit for both), while conserving memory. These results transform bit allocation from ad hoc tuning into a theoretically grounded, geometry-driven design principle for efficient LLM inference. 
Source code is available at \url{https://github.com/mohsenhariri/spectral-kv}.

\end{abstract}

\section{Introduction}\label{sec:intro}

Large Language Models (LLMs) have rapidly scaled in recent years, driving major advances in generative capabilities and reasoning performance~\cite{transformers, seq2seq}. Model size has increased by several orders of magnitude: GPT has grown from 117M parameters in GPT-1~\cite{gpt1} to 1.5B in GPT-2~\cite{gpt2}, 175B in GPT-3~\cite{gpt3}, and 1.8T in GPT-4~\cite{gpt4}. Open-source models have followed a similar trajectory, with Llama reaching 2T parameters~\cite{llama4}, Mistral Large scaling to 123B~\cite{mistrallarge2}, and DeepSeek V3 to 671B~\cite{deepseekv3}.

\begin{figure}[t]
    \centering
    \begin{subfigure}[h]{0.27\textwidth}
        \centering
        \includegraphics[width=\linewidth]{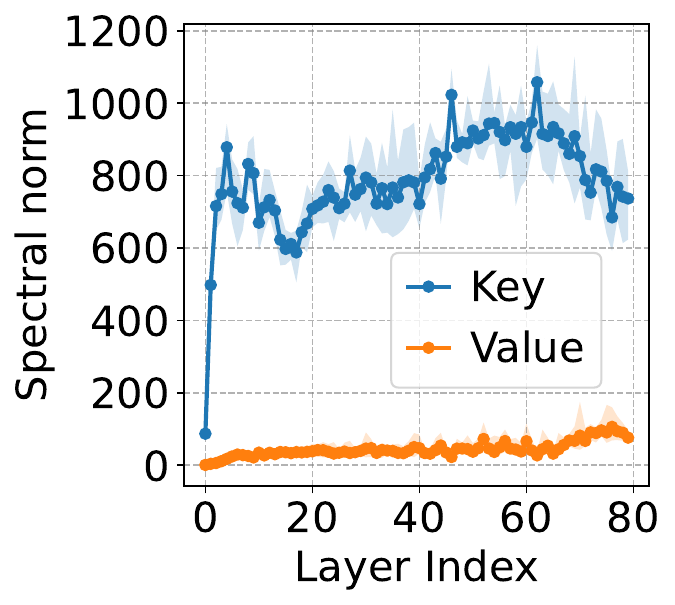}
    \end{subfigure}
    \hfill
    \begin{subfigure}[h]{0.20\textwidth}
        \centering
        \resizebox{!}{0.60in}{ 
        \setlength{\tabcolsep}{2pt}
        \begin{tabular}{lcc}
         \toprule
         Size & {\color{NavyBlue}K}\textsubscript{2}{\color{orange}V}\textsubscript{4} & \cellcolor[gray]{0.9}{\color{NavyBlue}K}\textsubscript{4}{\color{orange}V}\textsubscript{2} \\
         \midrule
         1B   & 0.06 & \cellcolor[gray]{0.9}0.34 \\
         8B   & 0.55 & \cellcolor[gray]{0.9}0.75 \\
         14B  & 0.78 & \cellcolor[gray]{0.9}0.91 \\
         70B  & 0.76 & \cellcolor[gray]{0.9}0.87 \\
         \bottomrule
        \end{tabular}
        }
        \vspace{10pt}
    \end{subfigure}
    \vspace{-8pt}
    \caption{\textbf{Key cache needs more bits.} (\textbf{Left}):
Spectral norms of the key cache (\textcolor{NavyBlue}{blue}) and value cache (\textcolor{orange}{orange}) across layers in Llama3.3-70B show that key caches consistently exhibit higher norms. 
(\textbf{Right}): GSM8k accuracy for two schemes: \(\text{K}_{2}\text{V}_{4}\), representing 2-bit allocation for the K cache and 4-bit allocation for the V cache and \(\text{K}_{4}\text{V}_{2}\), representing 4-bit allocation for the K cache and 2-bit allocation for the V cache, demonstrates that allocating more bits to the key cache maintains strong performance, confirming the efficacy of norm-aware, mixed-precision quantization.}
\vspace{-2ex}
    \label{fig:result_examples}
\end{figure}

However, this rapid growth has introduced severe \emph{inference-time memory bottlenecks}, primarily due to the Key-Value (KV) cache~\cite{cot}. As parameter counts increase, context lengths must also grow to support more complex reasoning, which further expands the KV cache and strains GPU memory~\cite{pmlr-v202-sheng23a}. Modern systems already support extremely long contexts~\cite{geminiflash2, openaio1, openaio3mini, deepseekr1}, reaching up to 10 million tokens~\cite{meta_llama4_blog_2025}, making memory-efficient methods essential.

\emph{KV cache quantization} refers to reducing the precision of KV tensors (e.g., BF16 to INT4), which can provide substantial memory savings while maintaining \emph{controlled accuracy degradation}, provided applied strategically~\cite{deepcompression, whitepaperquant}. Although many KV quantization methods have been proposed, most determine key–value bit splits through ad hoc hyperparameter tuning on cache statistics (e.g., inference-time activations

) rather than grounding them in intrinsic model properties (e.g., model weights). This raises a fundamental question: \textbf{\textit{How should bits be allocated in a principled and generalizable way?}}

To answer, Figure~\ref{fig:result_examples} illustrates two key observations. First, key caches (\(\mathbf{K}\)) consistently have larger spectral norms than value caches (\(\mathbf{V}\)). Second, assigning more bits to keys (e.g., \(\mathrm{K}_4\mathrm{V}_2\)) improves accuracy across multiple models compared to key-underprovisioned allocations (\(\mathrm{K}_2\mathrm{V}_4\)). These motivate a deeper investigation into the distinct roles of key-value weights (i.e., \(\mathbf{W}^\mathbf{K}\) and \(\mathbf{W}^\mathbf{V}\)) in attention mechanisms and their implications for quantization performance. Toward this, our contributions are:

\begin{itemize}[leftmargin=0.5cm, itemsep=0cm, topsep=0.1cm]
    \item We propose the \textbf{\op{}} theorem, proving that the expected spectral and Frobenius norms of \(\mathrm{W}^\mathrm{K}\) predominantly exceed those of \(\mathrm{W}^\mathrm{V}\) across prominent LLM models (i.e., Llama3 and Mistral herds).
    We then derive the \textbf{\mn{}} theorem, establishing the theoretical foundation on why assigning higher precision to \(\mathrm{K}\) than \(\mathrm{V}\) strictly reduces quantization error, enabling greater KV-cache compression while maintaining accuracy.
    
    \item We corroborate and operationalize these theorems across a diverse set of models (Llama-3.2-1B/3B/8B, Llama-3.3-70B, Phi-4-14B, Qwen3-0.6B/1.7B/4B/8B, DeepSeek-R1, Mistral-0.3-7B), datasets (C4, MMLU, GSM8K, EQ-Bench, CoQA, and LongBench\footnote{We purposefully select generative tasks rather than commonsense reasoning tasks to better isolate quantization effects; see Section~\ref{sec:why_generative_tasks} for details.}), and two quantization backends (Optimum Quanto and HQQ). 

    Notably, \(\mathrm{K}_4\mathrm{V}_2\) retains \textbf{98.3\%} (1-shot) and \textbf{94.1\%} (5-shot) accuracy of \(\mathrm{K}_4\mathrm{V}_4\) accuracy while reducing KV-cache memory by \textbf{25\%}, demonstrating both the theoretical soundness and practical effectiveness of the proposed strategy.

    \item Owing to its efficient one-off tunability, we show that our geometry-driven mixed-precision strategy is \emph{orthogonal} to existing inference-time KV quantization methods and can be seamlessly integrated to yield synergistic gains. In a case study with rotation-based outlier redistribution, combining a key-prioritized quantization (\(\mathrm{K}_4\mathrm{V}_2\)) with \emph{key-only rotation} outperforms \(\mathrm{K}_4\mathrm{V}_4\) by \textbf{4.4-18\%} in accuracy across tasks. In contrast, rotating value caches is unnecessary and sometimes detrimental

\end{itemize}

\section{Background and Related Work}
\label{sec:relwork}

\paragraph{KV Quantization.} 
Quantization methods for LLMs can be categorized by timing: \emph{training-time} and \emph{post-training} (PTQ)~\cite{gholami2021surveyquantizationmethodsefficient}. Training-time quantization integrates quantization into model training, typically achieving higher accuracy by quantizing weights or activations during the optimization process. However, it requires labeled data and incurs significant training overhead. PTQ applies quantization after training, avoiding retraining costs and labeled data requirements~\cite{whitepaperquant}, but sometimes yields lower accuracy.

PTQ can target different model components: weights, activations, or the KV cache. Weight-only quantization~\cite{gptq, obq, awq} achieves strong accuracy but does not reduce activation or KV memory. Weight-activation quantization~\cite{dettmers2022llmint8, smoothquant, omniquant} reduces overall memory but often sacrifices accuracy. KV quantization offers the best of both: it targets the rapidly growing KV cache~\cite{efficientlyscalingtransformerinference}, providing activation-level memory savings while maintaining the accuracy of weight-only methods~\cite{wkvquant}.

\paragraph{Existing KV Quantization Schemes.} 
KV quantization methods can be categorized by how they treat keys and values~\cite{li2025surveylargelanguagemodel}:
\begin{itemize}[leftmargin=0.5cm, itemsep=0cm, topsep=0.1cm]
    \item \textbf{Outlier redistribution.} These methods smooth or relocate outliers (i.e., unusually large activation values that dominate quantization ranges) in KV tensors, e.g., SmoothQuant~\cite{smoothquant}, AWQ~\cite{awq}, and OmniQuant~\cite{omniquant}.
    \item \textbf{Fixed-precision.} A single bit-width is used for both keys and values, ignoring their different roles and statistical properties~\cite{zeroquant, flexgen, kivi}.
    \item \textbf{Mixed-precision.} Different bit-widths are assigned to different parts of the cache~\cite{kvquant, wkvquant, snapkv, qaq, skvq, csr, kvtuner}.
\end{itemize}

While mixed-precision schemes are the most flexible, existing methods have not systematically explored \emph{asymmetric bit allocation between keys and values}. KVQuant~\cite{kvquant}, for example, focuses on vector-wise outlier handling rather than analyzing keys and values separately. Only a few methods have attempted to address this issue, and even then, only superficial insights have been gained.

\paragraph{Needs and Gaps.}
Despite rapid progress in KV quantization, several needs from the LLM research community remain unmet.  
\textbf{First}, there is a need for \emph{principled strategies} to guide bit allocation. Current frameworks either treat the key-value split as a hyperparameter tuned through grid search \cite{kvtuner} 
or rely on heuristics derived from cache statistics (e.g., activation ranges or distributions collected at inference time) \cite{h2o, pm-kvq, mikv}. These approaches are costly, model- or data-specific, and provide little theoretical insight into the inherent differences between keys and values.  
\textbf{Second}, there is a need to \emph{understand and exploit key-value asymmetry}. Existing works such as KVTuner~\cite{kvtuner}, SKVQ~\cite{skvq}, and QAQ~\cite{qaq} briefly observe that allocating more bits to keys can preserve accuracy, but none explain \emph{why} or propose a generalizable strategy. KVTuner reports differences in attention vs. perplexity errors across bit pairs without analysis; SKVQ finds asymmetric allocations (e.g., 2-bit keys, 1.5-bit values) through hyperparameter search rather than model structure; and QAQ focuses on different data types for keys and values rather than bit-width asymmetry.  
\textbf{Third}, there is a need for \emph{lightweight, modular methods that integrate seamlessly with existing KV quantization frameworks}. The studies mentioned above often involve complex, runtime-dependent procedures that are difficult to generalize or combine, limiting their drop-in applicability and composability with other techniques.

\paragraph{Our Perspective.}
We address these needs by deriving bit allocation strategies \emph{directly from model weights}, analyzing spectral and Frobenius norms of key and value matrices. Our approach is \textbf{lightweight}, incurring only a one-off analytical cost per model without any inference-time introspection; \textbf{generalizable}, since weight statistics are invariant across inputs and tasks; and \textbf{principled}, grounding allocation in linear algebraic properties rather than heuristic search. Moreover, our mix-precision quantization oracle is \textbf{orthogonal} to existing KV quantization techniques, paving a foundation that others can build upon. For example, pairing our mixed-precision allocation with rotation-based outlier redistribution techniques yields complementary improvements in both accuracy and memory savings. In this way, we elevate bit allocation from an ad hoc design choice to a geometry-informed building block for future KV quantization frameworks.
\section{Norm Dynamics of KV Weights}
\label{theory}

We now establish a theoretical foundation for mixed-precision KV-cache quantization by analyzing the intrinsic geometry of the \emph{key} and \emph{value} projection weights. Our analysis proceeds in two steps. First, we prove that key weights systematically exhibit larger spectral and Frobenius norms than value weights (\op{}), a property that is preserved in the resulting key and value caches. Second, we demonstrate that this norm gap directly impacts quantization error, providing a formal guarantee that assigning higher bit precision to keys yields strictly lower distortion and higher inference accuracy than symmetric or inverted allocations (\mn{}).

\subsection{\op{}: Key Weights Dominate in Norm}
\label{sec:kvnd}

The key weight matrix \( \mathrm{W}^K \) maps hidden states into the key cache, whereas the value weight matrix \( \mathrm{W}^V \) determines the representations retrieved during attention. Because quantization error scales approximately with the dynamic range of the signal, the relative magnitudes of \( \|\mathrm{W}^K\| \) and \( \|\mathrm{W}^V\| \), e.g., their Frobenius or spectral norms, indicate which cache is more sensitive to quantization.

\begin{figure}[htbp]
  \centering
  \includegraphics[width=\columnwidth]{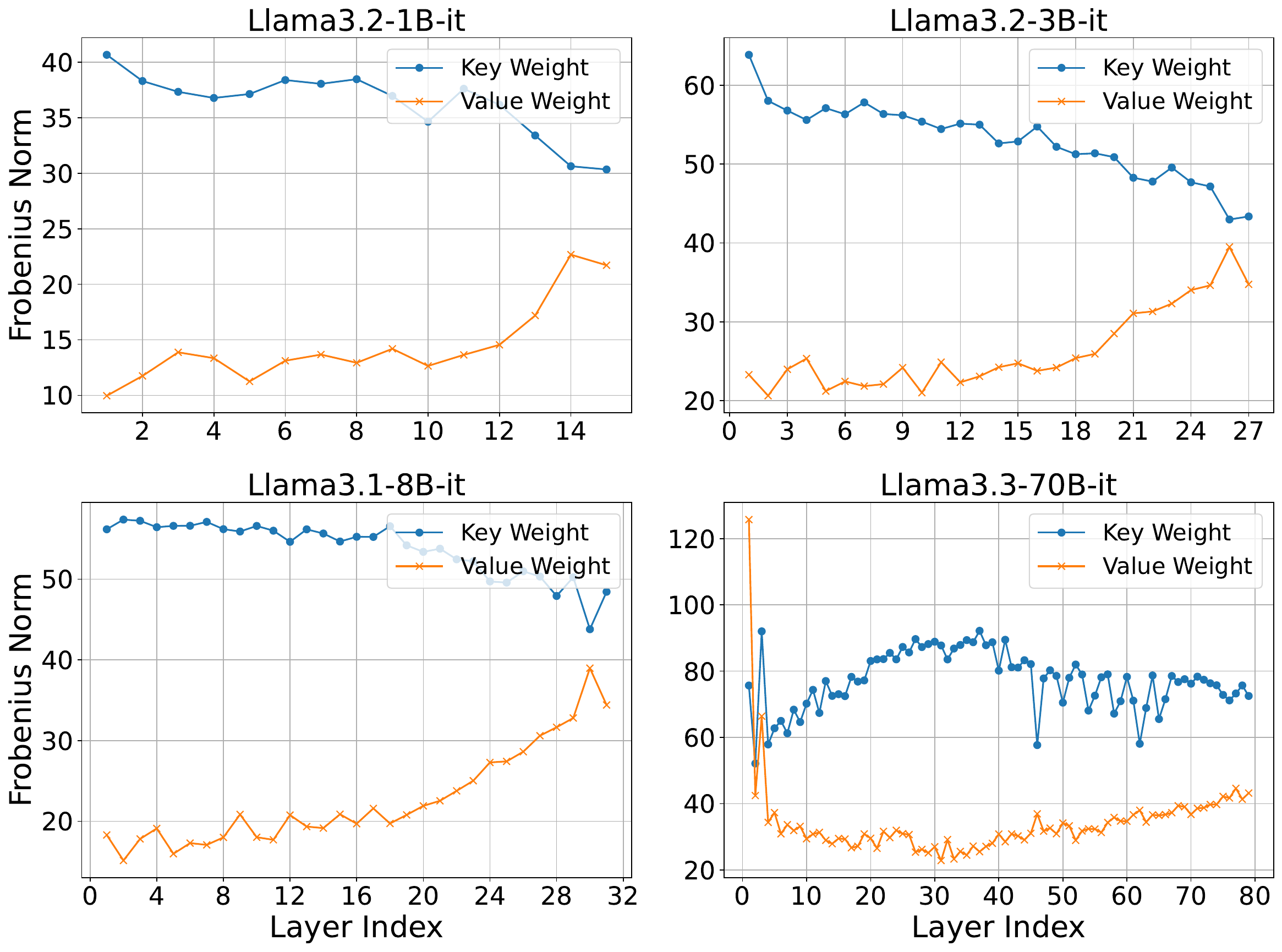}
  \caption{\textbf{Frobenius norms of key and value weight matrices across the Llama 3 family.} \( \|W^K\|_F \) consistently exceeds \( \|W^V\|_F \) across nearly all layers, with the exception occurring in early layers of the 70B variant.}
  \label{fig:frobweights_llama}
\end{figure}

Empirical measurements across four Llama-3 herds (Figure~\ref{fig:frobweights_llama}) show that the Frobenius norm of \( \mathrm{W}^K \) consistently exceeds that of \( \mathrm{W}^V \) in nearly every layer; the same ordering holds for spectral norms. This persistent gap motivates the following theorem.

\begin{theorembox}[\op{}]
\label{thm:op}
Let \( \mathrm{W}^K \) and \( \mathrm{W}^V \) denote the key and value projection matrices in a Transformer. Then
\[
\mathbb{E}\bigl[\|\mathrm{W}^K\|_{F}\bigr] > \mathbb{E}\bigl[\|\mathrm{W}^V\|_{F}\bigr].
\]
\end{theorembox}

The detailed proof is provided in Appendix~\ref{append:theorem1}. The core idea here is to examine how the Frobenius norms of the key and value weight matrices evolve during training. The analysis begins with Xavier initialization~\cite{pmlr-v9-glorot10a}, where all projection matrices have identical expected norms, and tracks their evolution under stochastic gradient descent (SGD). Intuitively, \(\mathrm{W}^K\) play a dual role: they shape the attention map and determine what representations are stored in the cache. Each input is multiplied by \(\mathrm{W}^K\) to produce keys that interact multiplicatively with the queries derived from \(\mathrm{W}^Q\) (i.e., the query projection). As training proceeds, \(\mathrm{W}^Q\) typically grows to sharpen attention, amplifying the gradient signals backpropagated into \(\mathrm{W}^K\). By contrast, \(\mathrm{W}^V\) only influences post-attention representations, so its gradients lack this multiplicative amplification. This architectural asymmetry causes \(\mathrm{W}^K\) to receive systematically larger updates, leading to persistently larger norms over time.

This phenomenon is ubiquitous. Appendix~\ref{appendix:kvw} offers analogous results for Mistral family~\cite{mistrallarge2}, demonstrating the generality of the pattern. We next show that this norm disparity has direct consequences for quantization.

\subsection{\mn{}: Key-Favored Allocation Minimizes Quantization Error}
\label{sec:kdq}

Theorem~\ref{thm:op} implies that, on average, \(\mathrm{W}^K\) and their resulting activations, \(\mathrm{K}\), have larger magnitude than their value counterparts. Since quantization error under uniform scalar quantization scales with the signal energy, assigning equal bit precision to both is sub-optimal.

Consider an additive-residual Transformer block (layer normalization omitted for clarity~\cite{elhage2021mathematical}):
\[
h_{l+1} = h_l + W_l h_l.
\]
Quantizing \( W_l \) to \( \tilde W_l = W_l + \Delta_l \) introduces an error bounded by
\[
\|h_{l+1} - \tilde h_{l+1}\|_2
= \|\Delta_l h_l\|_2
\le \|\Delta_l\|_2 \|h_l\|_2.
\]
After \(L\) layers, the worst-case deviation accumulates multiplicatively:
\[
\|h_L - \tilde h_L\|_2
\le \Bigl(\prod_{i=1}^{L} \|W_i\|_2\Bigr)
\max_{1\le i \le L}\|\Delta_i\|_2.
\]
Because the expected norms satisfy
\[
\mathbb{E}\bigl[\|W^K\|_{F}^{2}\bigr]
>
\mathbb{E}\bigl[\|W^V\|_{F}^{2}\bigr],
\]
any quantization noise injected along the key path is amplified more strongly through the network.

Let \( X \in \mathbb{R}^{\text{seq}\times d_{\text{model}}} \) be the hidden-state matrix at a given layer. The same input generates both caches via
\[
K = X W^K, \qquad V = X W^V.
\]
Multiplying the previous inequality by \( X \) and applying sub-multiplicativity of the Frobenius norm yields
\[
\mathbb{E}\bigl[\|K\|_{F}^{2}\bigr] >
\mathbb{E}\bigl[\|V\|_{F}^{2}\bigr],
\]
i.e., the norm gap persists in the caches.

\paragraph{Quantization Error and Norm Magnitude.}
For a matrix \( M \in \mathbb{R}^{m \times n} \), the squared Frobenius norm equals the total signal energy and the sum of squared singular values. When quantized with \(b\)-bit uniform scalar quantization, the expected mean-square error satisfies
\[
\mathbb{E}\bigl[\|M - \tilde M\|_{F}^{2}\bigr]
= \Theta\bigl(\|M\|_{F}^{2}\, 2^{-2b}\bigr),
\]
with constants depending only on the quantizer. Since key and value caches have identical shape, minimizing quantization error reduces to allocating bits in proportion to their energy. When \(\|K\|_{F}\gg\|V\|_{F}\) and equal bit-widths are used, key-cache error dominates:
\[
\mathbb{E}\bigl[\|K - \tilde K\|_{F}^{2}\bigr]
\gg
\mathbb{E}\bigl[\|V - \tilde V\|_{F}^{2}\bigr].
\]

This asymmetry in quantization error directly motivates an asymmetric bit allocation strategy, formalized as the following theorem:

\begin{theorembox}[\mn{}]
\label{thm:mn}
Let \((b_K,b_V)\) denote the bit allocations for key and value caches under a uniform scalar quantizer. For any pair with \( b_K > b_V \), the expected inference accuracy is strictly higher than for the swapped allocation \((b_V,b_K)\), provided that
\(\mathbb{E}[\|K\|_{F}^{2}] > \mathbb{E}[\|V\|_{F}^{2}]\).
\end{theorembox}

\begin{figure*}[htbp]
  \centering
  \includegraphics[width=\textwidth]{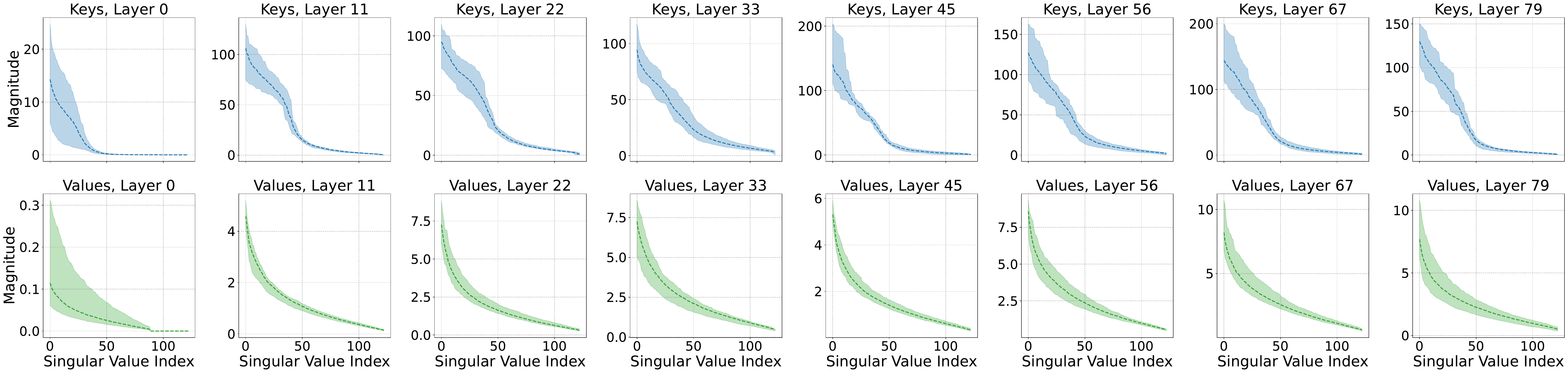}
  \caption{\textbf{Singular value spectra of key and value activations in Llama 3.3-70B on C4 benchmark dataset.} The x-axis shows singular value indices, ordered from the 5th largest onward for cleaner illustration, and the y-axis shows their magnitudes. Shaded regions mark the minimum-maximum range across attention heads within each layer, while dashed lines indicate the mean at each index. Beyond the top singular value (i.e., the spectral norm), key activations consistently exhibit larger singular values than value activations across the spectrum, highlighting their greater representational capacity. Full spectra are provided in Figure~\ref{fig:appendix_singular_value_distribution} of Appendix~\ref{sec:appendix-sing-all}.}
  \label{fig:svd}
\end{figure*}

Figure~\ref{fig:svd} provides empirical evidence for this effect. For Llama 3.3-70B on C4, the singular value spectra of the key caches consistently exceed those of the value caches beyond the top singular mode, indicating higher representational significance throughout the spectrum. Appendix~\ref{sec:appendix-sing-all} extends this analysis to the full singular value range, where Figure~\ref{fig:append_norms_frob_spectral} shows layer-wise Frobenius and spectral norms, all consistently revealing larger magnitudes for keys than for values.

The theoretical underpinnings of this phenomenon are established in Theorems~\ref{append:tq1} of Appendix~\ref{sec:spectral_norm_error_bound} and \ref{append:tq2} of Appendix~\ref{sec:Frobenius_norm_error_bound}, which derive a norm-dependent upper bound on the quantization error of an arbitrary matrix M. Appendix~\ref{append:tq3} then applies this bound to the key and value caches, showing that the larger norms of the key caches translate into proportionally higher quantization errors under equal bit allocations.

\section{Results}
\label{sec:results}

\subsection{Experimental Setup}
\label{sec:exp_setup_det}

Three evaluations are conducted: \emph{(i)} a quantization-error analysis, which measures reconstruction error across different bit-widths; \emph{(ii)} a downstream task accuracy evaluation, which assesses how mixed-precision KV cache quantization affects model accuracy on practical benchmarks; and \emph{(iii)} an integration case study with rotation-based outlier distribution methods, which investigates the effect of combining mixed precisions with rotation strategies. Full details on compute resources and software configurations are supplied in Appendix~\ref{appendix:compute_resources}.

\begin{table*}[t]
\centering
\setlength{\tabcolsep}{8pt}
\caption{\textbf{Quantization error (lower is better) of \(K\) and \(V\) caches under matched bit-widths.} 
Each cell reports the mean \(\pm\) standard deviation of the reconstruction MSE between the \emph{dequantized} cache and its BF16 reference, averaged over layers, heads, and tokens (10 sequences per dataset up to 1{,}000 tokens each). 
Both caches are quantized to the \emph{same} precision, i.e., \(\mathbf{K}_2\mathbf{V}_2\), \(\mathbf{K}_3\mathbf{V}_3\), and \(\mathbf{K}_4\mathbf{V}_4\), to isolate their intrinsic sensitivity at equal bit budgets. 
Across model families (Llama, Phi, Mistral, Qwen, DeepSeek) and datasets (C4, MMLU, GSM8K), \(\mathbf{K}\) consistently exhibits larger reconstruction error than \(\mathbf{V}\) at the same bit-width, and the error decreases monotonically with increasing precision, indicating that keys are the dominant source of quantization distortion.}
\label{tab:merged_quant_err}
\footnotesize
\resizebox{0.98\linewidth}{!}{
\begin{tabular}{ll c >{\columncolor[gray]{1}}c c >{\columncolor[gray]{1}}c c >{\columncolor[gray]{1}}c}
\toprule
\textbf{Dataset} & \textbf{Model} & \multicolumn{2}{c}{\textbf{2-bit}} & \multicolumn{2}{c}{\textbf{3-bit}} & \multicolumn{2}{c}{\textbf{4-bit}} \\
\cmidrule(lr){3-4} \cmidrule(lr){5-6} \cmidrule(lr){7-8}
& & $\mathbf{K}_\mathbf{2}$ & $\mathbf{V}_\mathbf{2}$ & $\mathbf{K}_\mathbf{3}$ & $\mathbf{V}_\mathbf{3}$ & $\mathbf{K}_\mathbf{4}$ & $\mathbf{V}_\mathbf{4}$ \\
\midrule
\multirow{7}{*}{\textbf{MMLU}} 
& Llama3.2-1B & 4.851 {\color{gray}{\scriptsize ± 1.037}} & 0.127 {\color{gray}{\scriptsize ± 0.101}} & 1.037 {\color{gray}{\scriptsize ± 0.265}} & 0.021 {\color{gray}{\scriptsize ± 0.015}} & 0.227 {\color{gray}{\scriptsize ± 0.059}} & 0.005 {\color{gray}{\scriptsize ± 0.003}} \\
& Llama3.1-8B-it & 6.003 {\color{gray}{\scriptsize ± 1.782}} & 0.187 {\color{gray}{\scriptsize ± 0.127}} & 1.082 {\color{gray}{\scriptsize ± 0.244}} & 0.028 {\color{gray}{\scriptsize ± 0.019}} & 0.235 {\color{gray}{\scriptsize ± 0.055}} & 0.006 {\color{gray}{\scriptsize ± 0.004}} \\
& Llama3.3-70B-it & 4.883 {\color{gray}{\scriptsize ± 1.106}} & 0.112 {\color{gray}{\scriptsize ± 0.093}} & 0.942 {\color{gray}{\scriptsize ± 0.198}} & 0.016 {\color{gray}{\scriptsize ± 0.012}} & 0.206 {\color{gray}{\scriptsize ± 0.043}} & 0.003 {\color{gray}{\scriptsize ± 0.003}} \\
& Phi4 & 5.929 {\color{gray}{\scriptsize ± 1.545}} & 0.657 {\color{gray}{\scriptsize ± 0.472}} & 1.306 {\color{gray}{\scriptsize ± 0.231}} & 0.103 {\color{gray}{\scriptsize ± 0.070}} & 0.286 {\color{gray}{\scriptsize ± 0.050}} & 0.022 {\color{gray}{\scriptsize ± 0.015}} \\
& Mistral0.3-7B & 4.718 {\color{gray}{\scriptsize ± 1.340}} & 0.398 {\color{gray}{\scriptsize ± 0.405}} & 0.941 {\color{gray}{\scriptsize ± 0.240}} & 0.059 {\color{gray}{\scriptsize ± 0.059}} & 0.206 {\color{gray}{\scriptsize ± 0.053}} & 0.013 {\color{gray}{\scriptsize ± 0.013}} \\
& Qwen2.5-14B & 5.184 {\color{gray}{\scriptsize ± 2.241}} & 1.270 {\color{gray}{\scriptsize ± 1.547}} & 1.005 {\color{gray}{\scriptsize ± 0.288}} & 0.182 {\color{gray}{\scriptsize ± 0.221}} & 0.223 {\color{gray}{\scriptsize ± 0.067}} & 0.040 {\color{gray}{\scriptsize ± 0.052}} \\
& R1Q-14B & 5.126 {\color{gray}{\scriptsize ± 2.375}} & 1.406 {\color{gray}{\scriptsize ± 1.609}} & 0.900 {\color{gray}{\scriptsize ± 0.269}} & 0.198 {\color{gray}{\scriptsize ± 0.226}} & 0.199 {\color{gray}{\scriptsize ± 0.062}} & 0.044 {\color{gray}{\scriptsize ± 0.052}} \\
\midrule
\multirow{7}{*}{\textbf{C4}} 
& Llama3.2-1B & 4.885 {\color{gray}{\scriptsize ± 1.056}} & 0.207 {\color{gray}{\scriptsize ± 0.166}} & 1.074 {\color{gray}{\scriptsize ± 0.289}} & 0.030 {\color{gray}{\scriptsize ± 0.024}} & 0.233 {\color{gray}{\scriptsize ± 0.062}} & 0.006 {\color{gray}{\scriptsize ± 0.005}} \\
& Llama3.1-8B-it & 6.262 {\color{gray}{\scriptsize ± 1.789}} & 0.254 {\color{gray}{\scriptsize ± 0.185}} & 1.128 {\color{gray}{\scriptsize ± 0.249}} & 0.036 {\color{gray}{\scriptsize ± 0.026}} & 0.247 {\color{gray}{\scriptsize ± 0.056}} & 0.008 {\color{gray}{\scriptsize ± 0.005}} \\
& Llama3.3-70B-it & 4.391 {\color{gray}{\scriptsize ± 1.027}} & 0.121 {\color{gray}{\scriptsize ± 0.097}} & 0.847 {\color{gray}{\scriptsize ± 0.175}} & 0.017 {\color{gray}{\scriptsize ± 0.013}} & 0.186 {\color{gray}{\scriptsize ± 0.038}} & 0.004 {\color{gray}{\scriptsize ± 0.003}} \\
& Phi4 & 5.715 {\color{gray}{\scriptsize ± 1.442}} & 0.850 {\color{gray}{\scriptsize ± 0.684}} & 1.316 {\color{gray}{\scriptsize ± 0.245}} & 0.124 {\color{gray}{\scriptsize ± 0.093}} & 0.291 {\color{gray}{\scriptsize ± 0.056}} & 0.027 {\color{gray}{\scriptsize ± 0.020}} \\
& Mistral0.3-7B & 5.027 {\color{gray}{\scriptsize ± 1.332}} & 0.543 {\color{gray}{\scriptsize ± 0.493}} & 1.014 {\color{gray}{\scriptsize ± 0.269}} & 0.079 {\color{gray}{\scriptsize ± 0.068}} & 0.223 {\color{gray}{\scriptsize ± 0.060}} & 0.017 {\color{gray}{\scriptsize ± 0.015}} \\
& Qwen2.5-14B & 4.382 {\color{gray}{\scriptsize ± 2.170}} & 1.544 {\color{gray}{\scriptsize ± 1.872}} & 0.846 {\color{gray}{\scriptsize ± 0.250}} & 0.220 {\color{gray}{\scriptsize ± 0.265}} & 0.187 {\color{gray}{\scriptsize ± 0.060}} & 0.048 {\color{gray}{\scriptsize ± 0.060}} \\
& DeepSeekR1Q-14B & 4.832 {\color{gray}{\scriptsize ± 2.354}} & 1.651 {\color{gray}{\scriptsize ± 1.914}} & 0.927 {\color{gray}{\scriptsize ± 0.283}} & 0.232 {\color{gray}{\scriptsize ± 0.267}} & 0.201 {\color{gray}{\scriptsize ± 0.061}} & 0.051 {\color{gray}{\scriptsize ± 0.060}} \\
\midrule
\multirow{7}{*}{\textbf{GSM8K}} 
& Llama3.2-1B & 5.703 {\color{gray}{\scriptsize ± 1.557}} & 0.179 {\color{gray}{\scriptsize ± 0.136}} & 1.213 {\color{gray}{\scriptsize ± 0.352}} & 0.026 {\color{gray}{\scriptsize ± 0.020}} & 0.266 {\color{gray}{\scriptsize ± 0.078}} & 0.005 {\color{gray}{\scriptsize ± 0.004}} \\
& Llama3.1-8B-it & 6.445 {\color{gray}{\scriptsize ± 1.837}} & 0.213 {\color{gray}{\scriptsize ± 0.161}} & 1.184 {\color{gray}{\scriptsize ± 0.268}} & 0.030 {\color{gray}{\scriptsize ± 0.022}} & 0.257 {\color{gray}{\scriptsize ± 0.060}} & 0.007 {\color{gray}{\scriptsize ± 0.005}} \\
& Llama3.3-70B-it & 4.967 {\color{gray}{\scriptsize ± 1.127}} & 0.113 {\color{gray}{\scriptsize ± 0.091}} & 0.978 {\color{gray}{\scriptsize ± 0.203}} & 0.016 {\color{gray}{\scriptsize ± 0.012}} & 0.214 {\color{gray}{\scriptsize ± 0.044}} & 0.004 {\color{gray}{\scriptsize ± 0.003}} \\
& Phi4 & 6.610 {\color{gray}{\scriptsize ± 1.624}} & 0.785 {\color{gray}{\scriptsize ± 0.598}} & 1.498 {\color{gray}{\scriptsize ± 0.293}} & 0.116 {\color{gray}{\scriptsize ± 0.082}} & 0.330 {\color{gray}{\scriptsize ± 0.064}} & 0.025 {\color{gray}{\scriptsize ± 0.017}} \\
& Mistral0.3-7B & 5.308 {\color{gray}{\scriptsize ± 1.367}} & 0.461 {\color{gray}{\scriptsize ± 0.434}} & 1.065 {\color{gray}{\scriptsize ± 0.288}} & 0.067 {\color{gray}{\scriptsize ± 0.061}} & 0.232 {\color{gray}{\scriptsize ± 0.061}} & 0.015 {\color{gray}{\scriptsize ± 0.013}} \\
& Qwen2.5-14B & 4.829 {\color{gray}{\scriptsize ± 2.179}} & 1.736 {\color{gray}{\scriptsize ± 2.659}} & 0.979 {\color{gray}{\scriptsize ± 0.264}} & 0.241 {\color{gray}{\scriptsize ± 0.372}} & 0.214 {\color{gray}{\scriptsize ± 0.061}} & 0.051 {\color{gray}{\scriptsize ± 0.077}} \\
& DeepSeekR1Q-14B & 4.477 {\color{gray}{\scriptsize ± 2.176}} & 1.424 {\color{gray}{\scriptsize ± 1.752}} & 0.830 {\color{gray}{\scriptsize ± 0.256}} & 0.200 {\color{gray}{\scriptsize ± 0.242}} & 0.181 {\color{gray}{\scriptsize ± 0.058}} & 0.044 {\color{gray}{\scriptsize ± 0.056}} \\
\bottomrule
\end{tabular}
}
\end{table*}

\vspace{1ex}
\noindent\textbf{Quantization-Error Evaluation.} 
PyTorch’s weight-packing quantization is applied without residual buffers or activation grouping to isolate quantization effects. Ten random sequences are sampled from \textsc{C4}~\cite{c4}, \textsc{MMLU}~\cite{mmlu}, and \textsc{GSM8K}~\cite{gsm8k}, padded to the longest length, and generated autoregressively for up to 1{,}000 tokens. Both per-layer and per-head reconstruction errors are computed, along with global averages across all heads, layers, and tokens, to offer a comprehensive view of bit-width sensitivity. This experiment spans seven models, including Llama-3.2-1B, Llama-3.1-8B~\cite{llama3}, Phi-4-14B~\cite{phi4}, Mistral-0.3-7B~\cite{mistrallarge2}, Qwen-2.5-14B, Llama-3.3-70B, DeepSeek-R1-Qwen-14B, Phi-3-Medium-128K~\cite{phi3}, and Llama-3.1-Nemotron-70B~\cite{nemotron}.

\vspace{1ex}
\noindent\textbf{Downstream Task Accuracy.} 
Two representative quantization backends are employed. \emph{Optimum Quanto} applies token-wise (per-row) quantization with mixed 2/4-bit precision on Llama-3.2-1B, Llama-3.1-8B, Phi-4-14B, and DeepSeek-R1-Qwen-14B. \emph{HQQ}~\cite{hqq} applies channel-wise (per-column) quantization with bit-widths \(\{1,2,4,6,8\}\) on Llama-3.1-8B, Llama-3.2-1B, Llama-3.2-3B, and Qwen3-0.6B/1.7B/4B/8B~\cite{qwen3}. A 64-token residual buffer (which stores the most recent tokens in full precision before being periodically flushed) and 128-element activation grouping (which quantizes activations in fixed-size blocks to reduce overhead) are adopted to mirror practical decoding configurations used in KIVI~\cite{kivi}, Flash-Decoding~\cite{flashdecoding}, and Marlin~\cite{marlin}, ensuring that the evaluation reflects realistic deployment rather than idealized settings. Accuracy is measured on three generative benchmarks: \textsc{GSM8K}~\cite{gsm8k}, \textsc{CoQA}~\cite{coqa}, and \textsc{EQ-Bench}~\cite{eqbench}, which collectively probe mathematical reasoning, conversational QA, and structured long-form generation.

\vspace{1ex}
\noindent\textbf{Integration with Rotation-Based Methods.}  
QuaRot~\cite{quarot} is selected for this case study. It applies structured randomized Hadamard rotations to activations before quantization, effectively dispersing outliers and improving the uniformity of the quantization distribution. A three-dimensional design space is explored, spanning \emph{bit-width allocation}, \emph{group size configuration}, and \emph{rotation strategies}. Specifically, mixed-precision settings \(\mathrm{K}_2\mathrm{V}_2\), \(\mathrm{K}_2\mathrm{V}_4\), \(\mathrm{K}_4\mathrm{V}_2\), and \(\mathrm{K}_4\mathrm{V}_4\); key and value group sizes in \(\{32, 64, 128\}\); and four rotation strategies (no rotation, key-only, value-only, and both). The evaluation encompasses generative tasks including \textsc{CoQA}, \textsc{GSM8K}, \textsc{EQ-Bench}, and \textsc{LongBench}, utilizing the same seven models as in quantization-error evaluation.

\subsection{Mixed-Precision Quantization Error}
\label{sec:quanterrorexperiments}

Table~\ref{tab:merged_quant_err} reports reconstruction errors at 2-, 3-, and 4-bit precision for seven representative models spanning multiple model families (Llama, Phi, Mistral, Qwen, DeepSeek) and datasets; full results appear in Appendix~\ref{appendix-d:quant-error}. Figure~\ref{fig:bit_error} shows the complete error curves for Llama-3.3-70B on \textsc{C4}, and Figure~\ref{fig:layer-head-err} provides a per-layer breakdown for Llama-3.1-8B. Across all models, datasets, and bit-widths, key caches consistently incur larger reconstruction errors than value caches, and this gap remains stable across precision levels. These findings empirically support the theoretical prediction that key representations have higher energy and are therefore more sensitive to quantization.

\subsection{Mixed-Precision Downstream Accuracy}
\label{sec:exp-eval}

Table~\ref{tab:quanto} summarizes the Optimum Quanto results on GSM8K under both 1-shot and 8-shot Chain-of-Thought (CoT) prompting. Although CoT prompting can sometimes reduce reasoning accuracy, it is included here to assess the impact of longer contexts on quantized decoding. Across four representative models, including Llama-3.2-1B, Llama-3.1-8B, Phi-4-14B, and DeepSeek-R1-Qwen-14B, a consistent pattern emerges: allocating higher precision to the \emph{key} cache (\(\mathrm{K}_4\mathrm{V}_2\)) preserves accuracy substantially better than the inverse (\(\mathrm{K}_2\mathrm{V}_4\)). On average, \(\mathrm{K}_4\mathrm{V}_2\) recovers approximately \textbf{94\%} of the full-precision baseline, whereas value-favored allocations incur losses of up to 30 percentage points (\textbf{\textit{pp}}). The performance gap widens with model scale; for instance, under 1-shot GSM8K, the \(\mathrm{K}_4\mathrm{V}_2\) configuration outperforms \(\mathrm{K}_2\mathrm{V}_4\) by 30\,pp on Llama-3.2-1B and by 16\,pp on Phi-4-14B. Notably, \(\mathrm{K}_4\mathrm{V}_2\) nearly matches the symmetric \(\mathrm{K}_4\mathrm{V}_4\) baseline despite halving the value bit budget, indicating that downstream performance is primarily constrained by key precision.

\begin{table}[htbp]
  \centering
  \caption{\textbf{\textsc{GSM8K} – Downstream accuracy with Optimum Quanto (token-wise) mixed-precision KV quantization.} 
  Token-wise quantization is applied with supported precisions \(i,j \in \{2,4\}\), where \(\mathrm{K}_i\mathrm{V}_j\) denotes \(i\)-bit keys and \(j\)-bit values. 
  Results are shown for both 1-shot and 8-shot settings. Across models, the key-favored \(\mathrm{K}_4\mathrm{V}_2\) consistently outperforms the value-favored \(\mathrm{K}_2\mathrm{V}_4\) and approaches the \(\mathrm{K}_4\mathrm{V}_4\) baseline, demonstrating the benefits of prioritizing key precision.}
  \label{tab:quanto}
  \footnotesize
  \resizebox{\linewidth}{!}{
  \begin{tabular}{llrrrr}
    \toprule
    \textbf{Model} & \textbf{Shots} & \(\mathbf{K}_2\mathbf{V}_2\) & \(\mathbf{K}_2\mathbf{V}_4\) & \(\mathbf{K}_4\mathbf{V}_2\) & \(\mathbf{K}_4\mathbf{V}_4\) \\
    \midrule
    \multirow{2}{*}{\shortstack{Llama\\3.2-1B-it}}
      & 1-shot  & 0.033 & 0.035 & 0.338 & 0.357 \\
      & 8-shot  & 0.031 & 0.031 & 0.289 & 0.369 \\
    \midrule
    \multirow{2}{*}{\shortstack{Llama\\3.1-8B-it}}
      & 1-shot  & 0.511 & 0.547 & 0.752 & 0.754 \\
      & 8-shot  & 0.408 & 0.441 & 0.770 & 0.782 \\
    \midrule
    \multirow{2}{*}{\shortstack{Phi\\4-14B}}
      & 1-shot  & 0.759 & 0.783 & 0.913 & 0.923 \\
      & 8-shot  & 0.771 & 0.815 & 0.927 & 0.931 \\
    \midrule
    \multirow{2}{*}{\shortstack{DeepSeek\\R1Q-14B}}
      & 1-shot  & 0.772 & 0.775 & 0.865 & 0.867 \\
      & 8-shot  & 0.763 & 0.792 & 0.876 & 0.875 \\
    \bottomrule
  \end{tabular}}
\end{table}

\begin{table*}[t]
  \centering
  \scriptsize
  \setlength{\tabcolsep}{10pt}
  \caption{\textbf{Downstream accuracy with HQQ (channel-wise) mixed-precision KV quantization across GSM8K, EQ-Parseable, and CoQA.} 
  Each cell compares \(\mathbf{K}_i\mathbf{V}_x\) and \(\mathbf{K}_x\mathbf{V}_i\) for \(i \in \{1,2,4,6,8\}\), where \(x\) represents the mean accuracy averaged over the other cache’s bit-widths \(B=\{1,2,4,6,8\}\). 
  This corresponds to contrasting ``allocate \(i\) bits to keys (values averaged over \(B\))'' versus ``allocate \(i\) bits to values (keys averaged over \(B\)).'' 
  Across model scales from 0.6B to 32B and multiple tasks, key-favored allocations consistently yield higher accuracy, with more pronounced gains observed at lower precision and persistent benefits at higher precision, demonstrating that the key-first advantage generalizes beyond a single model, task, and quantization backend.}
  \label{tab:hqq}
  \begin{tabularx}{\textwidth}{lXXXXX}
    \toprule
    \textbf{GSM8K} &
      {\scriptsize \(\mathbf{K}_1\mathbf{V}_x\) vs.\ \(\mathbf{K}_x\mathbf{V}_1\)} &
      {\scriptsize \(\mathbf{K}_2\mathbf{V}_x\) vs.\ \(\mathbf{K}_x\mathbf{V}_2\)} &
      {\scriptsize \(\mathbf{K}_4\mathbf{V}_x\) vs.\ \(\mathbf{K}_x\mathbf{V}_4\)} &
      {\scriptsize \(\mathbf{K}_6\mathbf{V}_x\) vs.\ \(\mathbf{K}_x\mathbf{V}_6\)} &
      {\scriptsize \(\mathbf{K}_8\mathbf{V}_x\) vs.\ \(\mathbf{K}_x\mathbf{V}_8\)} \\
    \midrule
    Qwen3-0.6B   & \(0.00\stackrel{\scriptscriptstyle+3\%}{<}\mathbf{0.03}\)  & \(0.00\stackrel{\scriptscriptstyle+16\%}{<}\mathbf{0.16}\) & \(0.04\stackrel{\scriptscriptstyle+13\%}{<}\mathbf{0.17}\) & \(\mathbf{0.34}\stackrel{\scriptscriptstyle+16\%}{>}0.18\) & \(\mathbf{0.34}\stackrel{\scriptscriptstyle+16\%}{>}0.18\) \\
    Llama-3.2-1B & \(0.00\stackrel{\scriptscriptstyle+5\%}{<}\mathbf{0.05}\)  & \(0.01\stackrel{\scriptscriptstyle+18\%}{<}\mathbf{0.19}\) & \(\mathbf{0.27}\stackrel{\scriptscriptstyle+7\%}{>}0.20\) & \(\mathbf{0.28}\stackrel{\scriptscriptstyle+8\%}{>}0.20\) & \(\mathbf{0.28}\stackrel{\scriptscriptstyle+7\%}{>}0.21\) \\
    Llama-3.2-3B & \(0.00\stackrel{\scriptscriptstyle+19\%}{<}\mathbf{0.19}\) & \(0.27\stackrel{\scriptscriptstyle+17\%}{<}\mathbf{0.44}\) & \(\mathbf{0.57}\stackrel{\scriptscriptstyle+11\%}{>}0.46\) & \(\mathbf{0.58}\stackrel{\scriptscriptstyle+12\%}{>}0.46\) & \(\mathbf{0.59}\stackrel{\scriptscriptstyle+13\%}{>}0.46\) \\
    Qwen3-4B     & \(0.00\stackrel{\scriptscriptstyle+41\%}{<}\mathbf{0.41}\) & \(0.01\stackrel{\scriptscriptstyle+49\%}{<}\mathbf{0.50}\) & \(\mathbf{0.80}\stackrel{\scriptscriptstyle+29\%}{>}0.51\) & \(\mathbf{0.82}\stackrel{\scriptscriptstyle+31\%}{>}0.51\) & \(\mathbf{0.82}\stackrel{\scriptscriptstyle+31\%}{>}0.51\) \\
    Llama-3.1-8B & \(0.00\stackrel{\scriptscriptstyle+31\%}{<}\mathbf{0.31}\) & \(0.35\stackrel{\scriptscriptstyle+17\%}{<}\mathbf{0.52}\) & \(\mathbf{0.70}\stackrel{\scriptscriptstyle+16\%}{>}0.54\) & \(\mathbf{0.70}\stackrel{\scriptscriptstyle+16\%}{>}0.54\) & \(\mathbf{0.70}\stackrel{\scriptscriptstyle+16\%}{>}0.54\) \\
    Qwen3-8B     & \(0.00\stackrel{\scriptscriptstyle+48\%}{<}\mathbf{0.48}\) & \(0.08\stackrel{\scriptscriptstyle+46\%}{<}\mathbf{0.54}\) & \(\mathbf{0.86}\stackrel{\scriptscriptstyle+31\%}{>}0.55\) & \(\mathbf{0.87}\stackrel{\scriptscriptstyle+32\%}{>}0.55\) & \(\mathbf{0.86}\stackrel{\scriptscriptstyle+31\%}{>}0.55\) \\
    Qwen3-32B    & \(0.00\stackrel{\scriptscriptstyle+42\%}{<}\mathbf{0.42}\) & \(0.25\stackrel{\scriptscriptstyle+23\%}{<}\mathbf{0.48}\) & \(\mathbf{0.73}\stackrel{\scriptscriptstyle+23\%}{>}0.50\) & \(\mathbf{0.71}\stackrel{\scriptscriptstyle+21\%}{>}0.50\) & \(\mathbf{0.72}\stackrel{\scriptscriptstyle+21\%}{>}0.51\) \\
    \bottomrule
    \midrule
    \textbf{EQ-Parseable} & & & & & \\
    \midrule
    Qwen3-0.6B   & $0.00\stackrel{\scriptscriptstyle+10\%}{<}\mathbf{0.10}$ & $0.00\stackrel{\scriptscriptstyle+53\%}{<}\mathbf{0.53}$ & $0.51\stackrel{\scriptscriptstyle+2\%}{<}\mathbf{0.53}$  & $\mathbf{0.84}\stackrel{\scriptscriptstyle+32\%}{>}0.53$ & $\mathbf{0.85}\stackrel{\scriptscriptstyle+33\%}{>}0.52$ \\
    Llama-3.2-1B & $0.00\stackrel{\scriptscriptstyle+34\%}{<}\mathbf{0.34}$ & $0.26\stackrel{\scriptscriptstyle+37\%}{<}\mathbf{0.62}$ & $\mathbf{0.87}\stackrel{\scriptscriptstyle+22\%}{>}0.65$  & $\mathbf{0.90}\stackrel{\scriptscriptstyle+24\%}{>}0.66$ & $\mathbf{0.90}\stackrel{\scriptscriptstyle+25\%}{>}0.66$ \\
    Llama-3.2-3B & $0.00\stackrel{\scriptscriptstyle+36\%}{<}\mathbf{0.35}$ & $0.68\stackrel{\scriptscriptstyle+5\%}{<}\mathbf{0.73}$  & $\mathbf{0.90}\stackrel{\scriptscriptstyle+14\%}{>}0.76$ & $\mathbf{0.89}\stackrel{\scriptscriptstyle+13\%}{>}0.76$ & $\mathbf{0.90}\stackrel{\scriptscriptstyle+14\%}{>}0.76$ \\
    Qwen3-4B     & $0.00\stackrel{\scriptscriptstyle+54\%}{<}\mathbf{0.48}$ & $0.29\stackrel{\scriptscriptstyle+33\%}{<}\mathbf{0.58}$ & $\mathbf{0.84}\stackrel{\scriptscriptstyle+28\%}{>}0.60$  & $\mathbf{0.86}\stackrel{\scriptscriptstyle+31\%}{>}0.59$ & $\mathbf{0.85}\stackrel{\scriptscriptstyle+29\%}{>}0.60$ \\
    Llama-3.1-8B & $0.00\stackrel{\scriptscriptstyle+61\%}{<}\mathbf{0.61}$ & $\mathbf{0.78}\stackrel{\scriptscriptstyle+2\%}{>}0.76$  & $\mathbf{0.96}\stackrel{\scriptscriptstyle+19\%}{>}0.77$ & $\mathbf{0.97}\stackrel{\scriptscriptstyle+20\%}{>}0.77$ & $\mathbf{0.97}\stackrel{\scriptscriptstyle+21\%}{>}0.77$ \\
    Qwen3-8B     & $0.00\stackrel{\scriptscriptstyle+62\%}{<}\mathbf{0.58}$ & $0.54\stackrel{\scriptscriptstyle+17\%}{<}\mathbf{0.70}$ & $\mathbf{0.95}\stackrel{\scriptscriptstyle+25\%}{>}0.71$ & $\mathbf{0.96}\stackrel{\scriptscriptstyle+27\%}{>}0.71$ & $\mathbf{0.96}\stackrel{\scriptscriptstyle+26\%}{>}0.71$ \\
    Qwen3-32B    & $0.00\stackrel{\scriptscriptstyle+60\%}{<}\mathbf{0.47}$ & $0.56\stackrel{\scriptscriptstyle+7\%}{<}\mathbf{0.62}$  & $\mathbf{0.80}\stackrel{\scriptscriptstyle+22\%}{>}0.62$ & $\mathbf{0.80}\stackrel{\scriptscriptstyle+23\%}{>}0.62$ & $\mathbf{0.80}\stackrel{\scriptscriptstyle+22\%}{>}0.63$ \\
    \bottomrule
    \midrule
    \textbf{CoQA} & & & & & \\
    \midrule
    Qwen3-0.6B     & $0.20\stackrel{\scriptscriptstyle+33\%}{<}\mathbf{0.44}$ & $0.32\stackrel{\scriptscriptstyle+29\%}{<}\mathbf{0.52}$ & $\mathbf{0.65}\stackrel{\scriptscriptstyle+16\%}{>}0.53$ & $\mathbf{0.69}\stackrel{\scriptscriptstyle+23\%}{>}0.53$ & $\mathbf{0.69}\stackrel{\scriptscriptstyle+23\%}{>}0.53$ \\
    Llama-3.2-1B   & $0.22\stackrel{\scriptscriptstyle+29\%}{<}\mathbf{0.42}$ & $0.48\stackrel{\scriptscriptstyle+13\%}{<}\mathbf{0.57}$ & $\mathbf{0.67}\stackrel{\scriptscriptstyle+13\%}{>}0.58$ & $\mathbf{0.68}\stackrel{\scriptscriptstyle+14\%}{>}0.58$ & $\mathbf{0.68}\stackrel{\scriptscriptstyle+14\%}{>}0.58$ \\
    Llama-3.2-3B   & $0.21\stackrel{\scriptscriptstyle+50\%}{<}\mathbf{0.60}$ & $\mathbf{0.73}\stackrel{\scriptscriptstyle+6\%}{>}0.68$  & $\mathbf{0.79}\stackrel{\scriptscriptstyle+14\%}{>}0.68$ & $\mathbf{0.79}\stackrel{\scriptscriptstyle+14\%}{>}0.68$ & $\mathbf{0.79}\stackrel{\scriptscriptstyle+15\%}{>}0.67$ \\
    Qwen3-4B       & $0.34\stackrel{\scriptscriptstyle+34\%}{<}\mathbf{0.62}$ & $0.68\stackrel{\scriptscriptstyle+3\%}{<}\mathbf{0.70}$  & $\mathbf{0.81}\stackrel{\scriptscriptstyle+12\%}{>}0.71$ & $\mathbf{0.81}\stackrel{\scriptscriptstyle+12\%}{>}0.71$ & $\mathbf{0.81}\stackrel{\scriptscriptstyle+13\%}{>}0.70$ \\
    Llama-3.1-8B   & $0.21\stackrel{\scriptscriptstyle+50\%}{<}\mathbf{0.60}$ & $\mathbf{0.72}\stackrel{\scriptscriptstyle+7\%}{>}0.66$  & $\mathbf{0.78}\stackrel{\scriptscriptstyle+14\%}{>}0.67$ & $\mathbf{0.78}\stackrel{\scriptscriptstyle+15\%}{>}0.67$ & $\mathbf{0.78}\stackrel{\scriptscriptstyle+14\%}{>}0.67$ \\
    Qwen3-8B       & $0.38\stackrel{\scriptscriptstyle+37\%}{<}\mathbf{0.68}$ & $\mathbf{0.73}\stackrel{\scriptscriptstyle+1\%}{>}0.72$  & $\mathbf{0.82}\stackrel{\scriptscriptstyle+12\%}{>}0.72$ & $\mathbf{0.82}\stackrel{\scriptscriptstyle+12\%}{>}0.72$ & $\mathbf{0.82}\stackrel{\scriptscriptstyle+12\%}{>}0.72$ \\
    Qwen3-32B      & $0.32\stackrel{\scriptscriptstyle+44\%}{<}\mathbf{0.68}$ & $\mathbf{0.79}\stackrel{\scriptscriptstyle+9\%}{>}0.72$  & $\mathbf{0.82}\stackrel{\scriptscriptstyle+12\%}{>}0.73$ & $\mathbf{0.82}\stackrel{\scriptscriptstyle+12\%}{>}0.73$ & $\mathbf{0.82}\stackrel{\scriptscriptstyle+12\%}{>}0.73$ \\
    \bottomrule
  \end{tabularx}
\end{table*}

The HQQ results, shown in Table~\ref{tab:hqq}, extend these observations to a broader range of bit-widths, model scales, and tasks. The analysis systematically compares \(\mathbf{K}_i\mathbf{V}_x\) against \(\mathbf{K}_x\mathbf{V}_i\) for \(i \in \{1,2,4,6,8\}\), where \(x\) denotes the mean accuracy over all bit-widths of the other cache. This directly addresses the question: ``If \(i\) bits are available, should they be allocated to keys or values?'' Across all models (0.6B-32B) and datasets (GSM8K, CoQA, EQ-Parseable), the answer is consistently ``keys''. At ultra-low precision (1-2 bits), the advantage is especially pronounced; for instance, on GSM8K, allocating a single extra bit to keys with 1-bit quantization yields gains of +48\,pp for Qwen3-8B. Similar trends hold on EQ-Bench and CoQA: for Qwen3-0.6B on EQ-Parseable, prioritizing keys delivers up to +62\,pp, and key-first allocations never underperform value-first ones in any configuration. Even at moderate precisions (4-6 bits), key-centric allocation continues to offer 7-12\,pp improvements for 8-14B models, indicating that the advantage persists well beyond extreme compression.
\textbf{On average, \(\mathbf{K}_4\mathbf{V}_2\) retains 98.3\% accuracy of \(\mathbf{K}_4\mathbf{V}_4\)} (CoQA: 99.2\%, EQ-Bench: 99.35\%, GSM8K: 97.7\%; worst: 88.3\%, best: 103.5\%).

\begin{figure}[htbp]
    \centering
    \includegraphics[width=\linewidth]{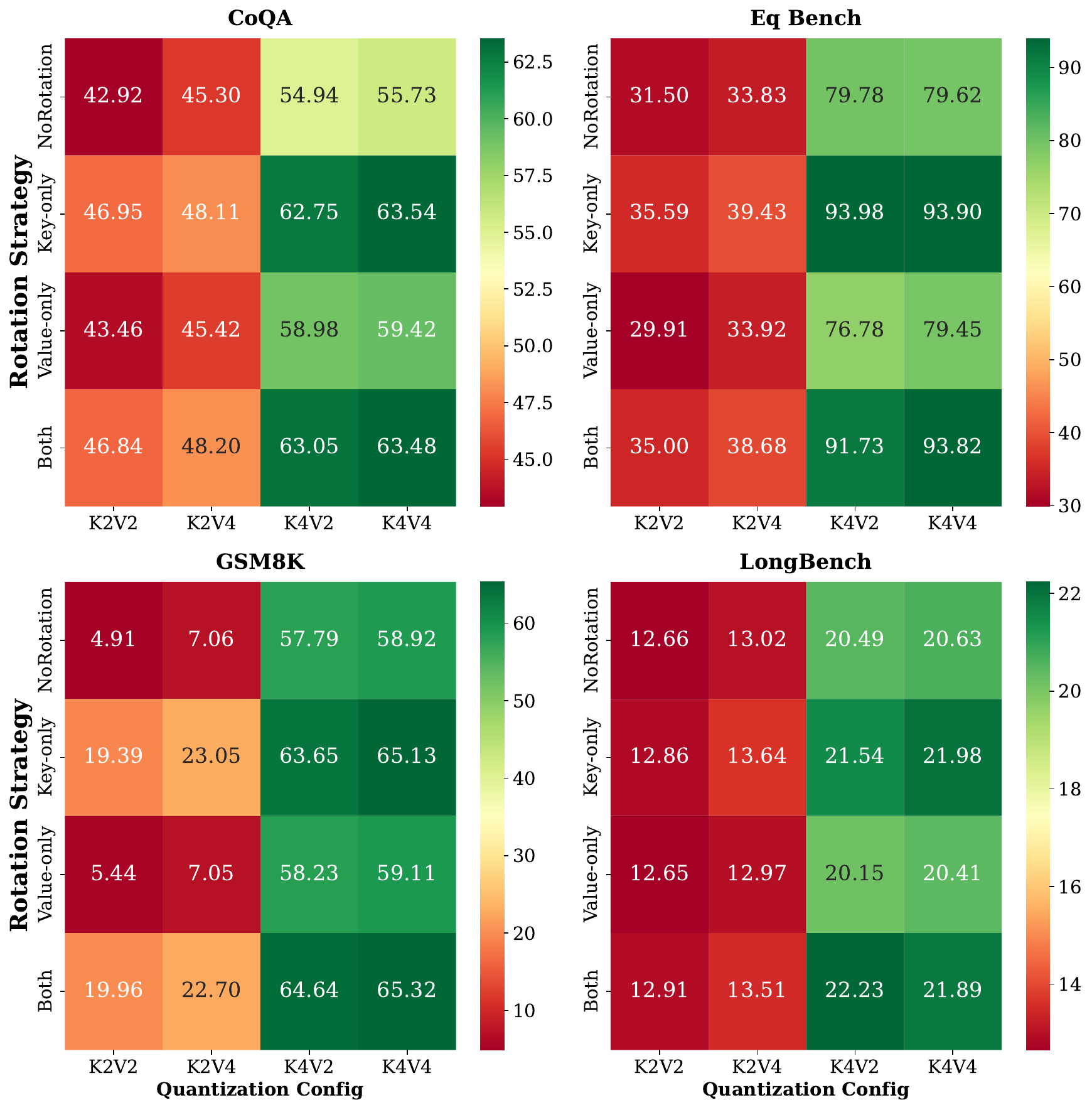}
    \caption{\textbf{Integration of rotation and mixed-precision quantization.} 
    Downstream accuracy is shown for four quantization configurations 
    (\(\mathrm{K}_2\mathrm{V}_2\), \(\mathrm{K}_2\mathrm{V}_4\), \(\mathrm{K}_4\mathrm{V}_2\), \(\mathrm{K}_4\mathrm{V}_4\)) 
    combined with four rotation strategies (none, key-only, value-only, both), 
    using a fixed group size of 64 for both keys and values. 
    Results are reported on \textsc{CoQA}, \textsc{GSM8K}, \textsc{EQ-Bench}, and \textsc{LongBench}, 
    enabling a controlled comparison of precision-rotation interactions.}
    \label{fig:rot}
\end{figure}

More detailed downstream accuracy results are provided in Appendix~\ref{sec:downstream_accuracy_full_hqq}. Overall, downstream accuracy is far more sensitive to key precision than to value precision. Across both token-wise and channel-wise quantization schemes, model scales, and task types, assigning the higher bit-width to \(\mathrm{K}\) consistently yields near-baseline accuracy while substantially reducing KV memory. This establishes a simple, backend-agnostic design principle for mixed-precision KV cache quantization: \emph{More for keys, less for values}.

\subsection{Integrating with Rotation-Based Methods}
\label{sec:rotation_integration}

Figure~\ref{fig:rot} visualizes how rotation interacts with key-value bit allocations.  
Across all tasks, applying rotation to \emph{keys} consistently yields larger gains than applying it to values.  
Notably, applying key-only rotation to the \(\mathrm{K}_4\mathrm{V}_2\) configuration achieves accuracy that closely matches the full \(\mathrm{K}_4\mathrm{V}_4\) baseline, indicating that the primary benefits of rotation arise from mitigating key outliers. These trends are consistent across tasks and model scales, reinforcing that rotation is most effective when applied in conjunction with key-favored bit allocation. 

Appendix~\ref{sec:rotation_group_results} presents detailed downstream results illustrating the synergistic effects of integrating rotation with mixed-precision quantization across tasks and models. It also examines the impact of key and value group sizes, showing that smaller sizes are beneficial for \(\mathrm{K}\) due to higher information density, whereas larger group sizes suffice for  \(\mathrm{V}\) given lower sensitivity to quantization.

\section{Conclusion}
\label{sec:conclusion}

As large language models increasingly devote most of their inference cost to KV-cache storage and access, effective cache compression has become essential for practical deployment.  
This work provides a theoretically grounded justification and solution to the bit-allocation problem by bridging model geometry and quantization design. We theoretically establish that key projections consistently carry higher information density than value projections. Building on this, we show that allocating higher precision to keys and lower precision to values minimizes quantization error. Extensive experiments across nine model families, six benchmarks, and two hardware-aligned backends validate this principle: a \(\mathrm{K}_4\mathrm{V}_2\) precision split reliably recovers up to \(98.3\%\) of \(\mathrm{K}_4\mathrm{V}_4\) accuracy while significantly reducing memory consumption. Moreover, we demonstrate that our geometry-driven strategy is \emph{orthogonal} to rotation-based outlier redistribution methods, enabling seamless integration and further accuracy gains. These findings elevate bit allocation from empirical tuning to a theoretically grounded geometry-driven design principle, providing clear guidance for efficient deployment and future hardware-algorithm co-design.

\section*{Limitations}
While our geometry-driven mixed-precision quantization framework demonstrates both theoretical soundness and practical effectiveness through rigorous analysis and extensive experiments, several limitations remain. 
First, all evaluations are conducted with a maximum context length of 2{,}000 tokens, which reflects common inference-time configurations but does not fully capture the behavior of models operating at much larger context windows. 
Extending the approach to longer contexts may expose additional challenges, including increased quantization sensitivity and higher memory management overheads. 
Addressing these factors is an important direction for future work.

\section*{Ethical Considerations}
This research aims to reduce the memory and computational costs of large language model inference by utilizing efficient KV-cache quantization. 
Such improvements have the potential to lower energy consumption and broaden access to language models in resource-constrained settings, promoting more sustainable and inclusive deployment. 
However, any compression technique entails accuracy trade-offs, which must be carefully monitored to avoid disproportionate impacts in high-stakes domains such as healthcare, law, or finance. 
Responsible deployment requires thorough evaluation of model behavior under mixed-precision settings, particularly for safety-critical applications.

\section*{Acknowledgment}

This research was supported in part by NSF awards 2117439, 2112606, and 2320952.

\bibliography{references}

\clearpage
\appendix

\section*{Appendix}
\section{Dynamics of the Norms of Key and Value Weight Matrices}\label{appendix:theorem1}

\subsection{Preliminaries and Notation}
\label{sec:dimensions}

\[
\begin{tabular}{@{}ll@{}}
  $c$      & Sequence length\\
  $d_{m}$  & Embedding dimension\\
  $d_{v}=d_{k}=d_{m}/n_{h}$ 
           & Single head dimension\\
  $n_{h}$  & Number of attention heads
\end{tabular}
\]

\[
\begin{array}{@{}ll@{}}
  X        & \text{Input matrix }(c\times d_{m})\\
  X^{O}    & \text{Output matrix }(c\times d_{m})\\
  W^{V}    & \text{Value weights }(d_{m}\times d_{m})\\
  W^{K}    & \text{Key weights }(d_{m}\times d_{m})\\
  W^{Q}    & \text{Query weights }(d_{m}\times d_{m})\\
  K=XW^{K} & (c\times d_{m})\\
  Q=XW^{Q} & (c\times d_{m})\\
  V=XW^{V} & (c\times d_{m})\\
  S=QK^{\top}/\sqrt{d_{m}}
           & (c\times c)\\
  A=\operatorname{softmax}(S)
           & (c\times c)\\
  \mathcal{L} & \text{loss function.}
\end{array}
\]

\subsection{Training Dynamics of Frobenius Norms}\label{append:theorem1}

We compare the long-time behavior of the Frobenius norms of the key ($W^{K}$) and value ($W^{V}$) weight matrices of a single-head self-attention layer trained with stochastic gradient descent (SGD). We show that, under standard isotropic assumptions, $\|W^{K}\|_{F}$ grows faster than $\|W^{V}\|_{F}$.

\vspace{1ex}
\noindent\textbf{Update rule.}
At step $i$, the weights follow
\begin{equation*}
  W^{m}_{i+1} \;=\; W^{m}_{i} - \eta \,
  \frac{\partial \mathcal{L}}{\partial W^{m}_{i}}, 
  \qquad m\in\{K,V\},
\end{equation*}
with learning rate $\eta$. Squaring the Frobenius norm of both sides gives

\begin{align*}\label{eq:norm-dynamics}
  \|W^{m}_{i+1}\|_{F}^{2}
  &= \|W^{m}_{i}\|_{F}^{2}
   + \eta^{2}
     \Bigl\|
       \frac{\partial \mathcal{L}}{\partial W^{m}_{i}}
     \Bigr\|_{F}^{2}\\
  &\quad
   - 2\eta\,
     \Bigl\langle
       W^{m}_{i},
       \frac{\partial \mathcal{L}}{\partial W^{m}_{i}}
     \Bigr\rangle_{F}.
\end{align*}
where
\begin{align*}
  \|A\|_{F} &\equiv \sqrt{\operatorname{tr}(A^{\top}A)},\\
  \langle A,B\rangle_{F} &\equiv \operatorname{tr}(A^{\top}B).
\end{align*}

\vspace{1ex}
\noindent\textbf{Expectation over mini-batches.}
Taking an expectation over mini-batches yields

\begin{equation*} \label{eq:expected-dynamics}
\begin{aligned}
  \Delta
  \mathbb{E}\bigl[\|W^{m}_{i}\|_{F}^{2}\bigr]
  &\equiv
    \mathbb{E}\bigl[\|W^{m}_{i+1}\|_{F}^{2}\bigr]
    -\mathbb{E}\bigl[\|W^{m}_{i}\|_{F}^{2}\bigr]\\[4pt]
  &=
    \eta^{2}\,
    \mathbb{E}\!\Bigl[
      \Bigl\|
        \frac{\partial \mathcal{L}}{\partial W^{m}_{i}}
      \Bigr\|_{F}^{2}
    \Bigr]\\
  &\quad
    - 2\eta\,
    \mathbb{E}\!\Bigl[
      \bigl\langle
        W^{m}_{i},
        \frac{\partial \mathcal{L}}{\partial W^{m}_{i}}
      \bigr\rangle_{F}
    \Bigr].
\end{aligned}
\end{equation*}

In high-dimensional weight space, the gradient is almost orthogonal to the current weights, making the second term negligible. Hence,

\begin{equation*}\label{eq:dE}
  \Delta
  \mathbb{E}\bigl[\|W^{m}_{i}\|_{F}^{2}\bigr]
  \;\approx\;
  \eta^{2}\,
  \mathbb{E}\Bigl[
    \Bigl\|
      \frac{\partial \mathcal{L}}{\partial W^{m}_{i}}
    \Bigr\|_{F}^{2}
  \Bigr].
\end{equation*}

Throughout, we assume Xavier initialization: each entry of $W^{K}$ and $W^{V}$ is drawn i.i.d.\ from a zero-mean distribution whose variance preserves the input scale \cite{pmlr-v9-glorot10a}.

The attention outputs are
\begin{equation*}
  X^{O} \;=\; A\,V\,W^{O},
\end{equation*}
where $A=\operatorname{softmax}\!\bigl(QK^{\top}/\sqrt{d_{m}}\bigr)$ and $V=XW^{V}$. Differentials give

\begin{align*}
  d\mathcal{L}(X^{O})
  &= \operatorname{tr}\!\bigl[
       (\partial\mathcal{L}/\partial X^{O})^{\!\top}
       \,AX\,dW^{V}W^{O}
     \bigr]\\
  &= \operatorname{tr}\!\bigl[
       (AX)^{\top}\,
       \frac{\partial\mathcal{L}}{\partial X^{O}}\,
       (W^{O})^{\top}
       \,dW^{V}
     \bigr].
\end{align*}
so that
\begin{equation*}\label{eq:grad-v}
  \frac{\partial\mathcal{L}}{\partial W^{V}}
  =(AX)^{\top}\,
  \frac{\partial\mathcal{L}}{\partial X^{O}}\,
  (W^{O})^{\top}.
\end{equation*}
Similarly, writing $S=QK^{\top}/\sqrt{d_{m}}$,

\begin{align*}
  d\mathcal{L}(S)
  &= \operatorname{tr}\!\bigl[
       (\partial\mathcal{L}/\partial S)^{\!\top}
       \,dS
     \bigr]\\
  &= \frac{1}{\sqrt{d_{m}}}\,
     \operatorname{tr}\!\bigl[
       X^{\top}\,
       (\partial\mathcal{L}/\partial S)^{\!\top}\,
       Q\,dW^{K}
     \bigr].
\end{align*}
yielding
\begin{equation*}\label{eq:grad-k}
  \frac{\partial\mathcal{L}}{\partial W^{K}}
  =\frac{1}{\sqrt{d_{m}}}\,
   X^{\top}\,
   (\partial\mathcal{L}/\partial S)^{\!\top}\,
   Q.
\end{equation*}

Squaring the Frobenius norm of~\eqref{eq:grad-v} and taking an expectation under the isotropic-input assumption,
\begin{align*}
  \mathbb{E}\!\bigl[\|
    \partial\mathcal{L}/\partial W^{V}
  \|^{2}_{F}\bigr]
  &= c\,\sigma_{x}^{2}\,\sigma_{o}^{2},
\end{align*}
where
$\mathbb{E}[X^{\top}X]=c\sigma_{x}^{2}I$
and
$\mathbb{E}\Bigl[\Bigl(A^{\top} \tfrac{\partial \mathcal{L}}{\partial X^O}(W^O)^{\top} \Bigr)  \Bigl(A^{\top} \tfrac{\partial \mathcal{L}}{\partial X^O}(W^O)^{\top} \Bigr)^{\top} \Bigr] = \sigma_o^{2}I $.

For the key weights,
\begin{align*}
  \mathbb{E}\!\bigl[\|
    \partial\mathcal{L}/\partial W^{K}
  \|^{2}_{F}\bigr]
  =\frac{\sigma_{x}^{2}\,\sigma_{s}^{2}}{d_{m}}\,
    \|Q\|_{F}^{2},
\end{align*}
with
$\sigma_{s}^{2}$ the entry-wise variance of
$\partial\mathcal{L}/\partial S$.

Substituting these expectations into~\eqref{eq:dE} shows that

\(\mathbb{E}\!\bigl[\|W^{K}\|^{2}_{F}\bigr]\) grows as
\(\eta^{2}\,\frac{\sigma_{x}^{2}\sigma_{s}^{2}}{d_{m}}\,\|Q\|_{F}^{2}\), whereas
\(\mathbb{E}\!\bigl[\|W^{V}\|^{2}_{F}\bigr]\) grows as
\(\eta^{2}\,c\sigma_{x}^{2}\sigma_{o}^{2}.\)

\noindent Because $\|Q\|_{F}^{2}$ itself increases during training, $\|W^{K}\|_{F}$ eventually dominates:
\begin{equation*}\label{t2}
   \mathbb{E}\!\bigl[\|
     W^{K}
  \|_{F}\bigr]
  \;>\;
  \mathbb{E}\!\bigl[\|
     W^{V}
  \|_{F}\bigr].
\end{equation*}

As shown in Equation \ref{t2}, the expectation of the Frobenius norm of \( W^k \) is greater than that of \( W^v \).\footnote{The analysis considers a single-head decoder-only Transformer without loss of generality; the conclusions extend directly to multi-head, group-query, and multi-query attention.}

\section{Quantization Error Bounds}
\label{appendix:proofs}

We establish upper bounds on the quantization error incurred when a matrix is represented using a finite bit-width in two's complement format. Two theorems are presented: one that characterizes the error in terms of the spectral norm and another in terms of the Frobenius norm.

The first theorem demonstrates that the spectral norm of the quantization error is bounded by
\[
\|A - \widehat{A}\|_2 \;\lesssim\; \frac{\sqrt{m\,n}}{2^b}\,\|A\|_2.
\]
This result implies that, for a given bit-width \(b\), matrices with larger spectral norms incur proportionally larger quantization errors. Consequently, a matrix that exhibits a larger spectral norm is more susceptible to quantization errors. To control error propagation in such matrices, a higher bit width is necessary.

The second theorem provides an analogous bound for the Frobenius norm,
\[
\|A - \widehat{A}\|_F \;\lesssim\; \frac{\sqrt{m\,n}}{2^b}\,\|A\|_F.
\]
Similar to the spectral norm result, this bound indicates that the quantization error, measured in the Frobenius norm, is directly proportional to the norm of the original matrix. Hence, matrices with larger Frobenius norms are also more vulnerable to quantization errors and would benefit from a higher precision during quantization.

\subsection{Preliminaries}

Let \(A \in \mathbb{R}^{m \times n}\) denote a real matrix whose entries are to be stored using a fixed number of bits in two's complement representation. The following notation is adopted:
\begin{itemize}
  \item \textit{Bit Depth.} Given \(b\) bits in two's-complement format, each representable integer \(q\) lies in the interval
  \[
    q \in \{-2^{b-1}, -2^{b-1}+1, \dots, 2^{b-1}-1\}.
  \]
  \item \textit{Maximum Entry Magnitude.} Define
  \[
    M \;=\; \max_{1 \leq i \leq m,\, 1 \leq j \leq n} \;|A_{ij}|.
  \]
  \item \textit{Scale Factor.} Set
  \[
    \alpha \;=\; \frac{M}{2^{b-1}-1}.
  \]
  This choice ensures that the scaled entries \(A_{ij}/\alpha\) lie within the representable range.
\noindent
\item \textit{Quantization.} Define the integer matrix \(Q \in \mathbb{Z}^{m \times n}\) by

\begin{equation*}
  \begin{aligned}
    Q_{ij} &= \mathrm{round}\!\Bigl(\frac{A_{ij}}{\alpha}\Bigr), \\
    \text{with} \quad Q_{ij} &\in \{-2^{b-1}, \dots, 2^{b-1}-1\}.
  \end{aligned}
\end{equation*}

\noindent
\item \textit{Dequantization (Reconstruction).} The reconstructed matrix \(\widehat{A}\) is given by
\[
  \widehat{A}_{ij} \;=\; \alpha \, Q_{ij}.
\]

\end{itemize}
\medskip

\noindent
\textbf{Objective.} The aim is to bound the errors
\[
  \|A - \widehat{A}\|_2 \quad \text{and} \quad \|A - \widehat{A}\|_F,
\]
in terms of \(b\), \(m\), \(n\), and the norms of \(A\).

\subsection{Spectral Norm Error Bound}
\label{sec:spectral_norm_error_bound}
\medskip
\begin{theorembox}[Spectral Norm Error Bound for Uniform Quantization] 
\label{append:tq1}
Let \(A \in \mathbb{R}^{m \times n}\) and \(b \in \mathbb{N}\) be given. Consider two's complement quantization with a scale factor of
\[
  \alpha \;=\; \frac{M}{2^{b-1} - 1}, \quad M \;=\; \max_{i,j} |A_{ij}|.
\]
Define 
\[
  Q_{ij} \;=\; \mathrm{round}\!\Bigl(\frac{A_{ij}}{\alpha}\Bigr)
  \quad \text{and} \quad
  \widehat{A}_{ij} \;=\; \alpha\,Q_{ij}.
\]
Then, the following bound holds:
\begin{equation*}
  \begin{aligned}
    M &\le \|A - \widehat{A}\|_2 \le \sqrt{m\,n} \, \frac{M}{2(2^{b-1} - 1)} \\
      &\le \sqrt{m\,n} \, \frac{\|A\|_2}{2(2^{b-1} - 1)}.
  \end{aligned}
\end{equation*}
In approximate form for large \(b\),
\[
  \|A - \widehat{A}\|_2 
  \;\lesssim\; 
  \frac{\sqrt{m\,n}}{2^b}\,\|A\|_2.
\]
\end{theorembox}

\begin{proof}
\textbf{Entrywise Bound.} By construction,
\[
  \left|\frac{A_{ij}}{\alpha} - Q_{ij}\right| \;\le\; \frac{1}{2}.
\]
Multiplying by \(\alpha\) gives
\[
  |A_{ij} - \widehat{A}_{ij}| \;\le\; \frac{\alpha}{2} \;=\; \frac{M}{2\,(2^{b-1} - 1)}.
\]
Thus, 
\[
  \max_{i,j}\,|A_{ij} - \widehat{A}_{ij}| \;\le\; \frac{M}{2\,(2^{b-1} - 1)}.
\]

\medskip

\textbf{Conversion to the Spectral Norm.} Using the inequality
\[
  \|B\|_2 \;\le\; \sqrt{m\,n}\,\max_{i,j}|B_{ij}|,
\]
with \(B = A - \widehat{A}\), it follows that
\[
  \|A - \widehat{A}\|_2 \;\le\; \sqrt{m\,n}\,\frac{M}{2\,(2^{b-1} - 1)}.
\]

\medskip

\textbf{Relating \(M\) to \(\|A\|_2\).} Since
\[
  M \;=\; \max_{i,j}|A_{ij}| \;\le\; \|A\|_2,
\]
the bound can be written as
\[
  \|A - \widehat{A}\|_2 \;\le\; \sqrt{m\,n}\,\frac{\|A\|_2}{2\,(2^{b-1} - 1)}.
\]
For large \(b\), where \(2^{b-1} - 1 \approx 2^{b-1}\), the bound becomes
\[
  \|A - \widehat{A}\|_2 \;\lesssim\; \frac{\sqrt{m\,n}}{2^b}\,\|A\|_2.
\]
\end{proof}

\subsection{Frobenius Norm Error Bound}
\label{sec:Frobenius_norm_error_bound}

\medskip
\begin{theorembox}[Frobenius Norm Error Bound for Uniform Quantization] 
\label{append:tq2}
Under the same setup as Theorem~\ref{append:tq1}, the Frobenius norm of the quantization error satisfies
\begin{equation*}
\begin{split}
  \|A - \widehat{A}\|_F
  &\;\le\;
  \sqrt{m\,n}\,{\textstyle\frac{M}{2(2^{b-1}-1)}} \\
  &\;\le\;
  \sqrt{m\,n}\,{\textstyle\frac{\|A\|_F}{2(2^{b-1}-1)}}.
\end{split}
\end{equation*}
In approximate form,
\[
  \|A - \widehat{A}\|_F
  \;\lesssim\;
  \frac{\sqrt{m\,n}}{2^b}\,\|A\|_F.
\]
\end{theorembox}

\begin{proof}
\textbf{Entrywise Bound.} As established,
\[
  |A_{ij} - \widehat{A}_{ij}| \;\le\; \frac{M}{2\,(2^{b-1}-1)} \quad \text{for all } i,j.
\]

\medskip

\textbf{Conversion to the Frobenius Norm.} By definition,
\[
  \|A - \widehat{A}\|_F^2 \;=\; \sum_{i=1}^m \sum_{j=1}^n (A_{ij} - \widehat{A}_{ij})^2,
\]
which yields
\[
  \|A - \widehat{A}\|_F^2 \;\le\; m\,n \left(\frac{M}{2\,(2^{b-1}-1)}\right)^2.
\]
Taking square roots leads to
\[
  \|A - \widehat{A}\|_F \;\le\; \sqrt{m\,n}\,\frac{M}{2\,(2^{b-1}-1)}.
\]

\medskip

\textbf{Relating \(M\) to \(\|A\|_F\).} Since
\[
  M^2 \;\le\; \sum_{i=1}^m \sum_{j=1}^n A_{ij}^2 \;=\; \|A\|_F^2,
\]
it follows that \(M \leq \|A\|_F\) and hence
\[
  \|A - \widehat{A}\|_F \;\le\; \sqrt{m\,n}\,\frac{\|A\|_F}{2\,(2^{b-1}-1)}.
\]
For large \(b\), this simplifies to
\[
  \|A - \widehat{A}\|_F \;\lesssim\; \frac{\sqrt{m\,n}}{2^b}\,\|A\|_F.
\]
\end{proof}

\medskip

\noindent
\textbf{Remark.} The results indicate that both the spectral norm and Frobenius norm errors satisfy similar approximate bounds:
\begin{equation*}
  \begin{aligned}
    \|A - \widehat{A}\|_2 &\lesssim \frac{\sqrt{m\,n}}{2^b}\,\|A\|_2,\\
    \|A - \widehat{A}\|_F &\lesssim \frac{\sqrt{m\,n}}{2^b}\,\|A\|_F.
  \end{aligned}
\end{equation*}

\bigskip

\subsection{Implications for KV Cache Quantization}
\label{append:tq3}

Consider the key and value cache
\[
  \mathbf{V} \in \mathbb{R}^{L \times d_{\text{head}}}, \quad
  \mathbf{K} \in \mathbb{R}^{L \times d_{\text{head}}},
\]
with quantization bit-widths denoted by \(b_V\) and \(b_K\), respectively. Let \(\widehat{\mathbf{V}}\) and \(\widehat{\mathbf{K}}\) denote the dequantized matrices, and define the quantization errors as
\[
  \mathbf{E}_V \;=\; \mathbf{V} - \widehat{\mathbf{V}}, \quad
  \mathbf{E}_K \;=\; \mathbf{K} - \widehat{\mathbf{K}}.
\]
An empirical observation is that
\[
  \|\mathbf{V}\|_{\ast} \;<\; \|\mathbf{K}\|_{\ast},
\]
where \(\ast\) denotes either the spectral norm (\(\|\cdot\|_2\)) or the Frobenius norm (\(\|\cdot\|_F\)). In practice, \(\mathbf{K}\) typically exhibits a larger norm than \(\mathbf{V}\).

\noindent\textbf{Spectral-Norm Perspective.}
Standard quantization error bounds yield
\begin{equation*}
  \begin{aligned}
    \|\mathbf{E}_V\|_2 &\lesssim \frac{\sqrt{L \, d_{\text{head}}}}{2^{b_V}}\;\|\mathbf{V}\|_2, \\
    \|\mathbf{E}_K\|_2 &\lesssim \frac{\sqrt{L \, d_{\text{head}}}}{2^{b_K}}\;\|\mathbf{K}\|_2.
  \end{aligned}
\end{equation*}
To achieve comparable spectral-norm errors (i.e., \(\|\mathbf{E}_V\|_2 \approx \|\mathbf{E}_K\|_2\)), it is necessary that
\[
   \frac{\sqrt{L \, d_{\text{head}}}}{2^{b_V}}\;\|\mathbf{V}\|_2 
   \;\approx\;
   \frac{\sqrt{L \, d_{\text{head}}}}{2^{b_K}}\;\|\mathbf{K}\|_2.
\]
Cancelling the common factor \(\sqrt{L\, d_{\text{head}}}\) yields
\[
  2^{b_V}\,\|\mathbf{V}\|_2 \;\approx\; 2^{b_K}\,\|\mathbf{K}\|_2,
\]
or equivalently,
\[
  2^{\,b_K - b_V} \;\approx\; \frac{\|\mathbf{V}\|_2}{\|\mathbf{K}\|_2}.
\]
Since \(\|\mathbf{V}\|_2 < \|\mathbf{K}\|_2\), it follows that \(b_K > b_V\).

\vspace{1ex}
\noindent\textbf{Frobenius Norm (MSE) Perspective.} 
The Frobenius norm of the quantization error corresponds directly to the mean-squared error (MSE) when normalized by the number of elements. Specifically, for a matrix \(\mathbf{A}\) with quantized approximation \(\widehat{\mathbf{A}}\), the MSE is
\[
  \mathrm{MSE}(\mathbf{A}, \widehat{\mathbf{A}}) 
  \;=\; \frac{1}{\mathrm{nnz}(\mathbf{A})}\,\|\mathbf{A} - \widehat{\mathbf{A}}\|_F^2,
\]
where \(\mathrm{nnz}(\mathbf{A})\) denotes the number of entries. Thus, controlling the Frobenius norm is equivalent to controlling the MSE up to a scaling factor.

The quantization error bounds under the Frobenius norm are given by
\begin{equation*}
  \begin{aligned}
    \|\mathbf{E}_V\|_F 
    &\;\lesssim\;
    \frac{\sqrt{L\, d_{\text{head}}}}{2^{b_V}}\;\|\mathbf{V}\|_F, \\
    \|\mathbf{E}_K\|_F 
    &\;\lesssim\;
    \frac{\sqrt{L\, d_{\text{head}}}}{2^{b_K}}\;\|\mathbf{K}\|_F,
  \end{aligned}
\end{equation*}
where \(L\) is the sequence length and \(d_{\text{head}}\) is the head dimension.

To ensure comparable Frobenius (or MSE) errors between keys and values, we require
\[
   \frac{\sqrt{L\, d_{\text{head}}}}{2^{b_V}}\;\|\mathbf{V}\|_F 
   \;\approx\;
   \frac{\sqrt{L\, d_{\text{head}}}}{2^{b_K}}\;\|\mathbf{K}\|_F,
\]
which simplifies to
\[
   2^{b_V}\,\|\mathbf{V}\|_F \;\approx\; 2^{b_K}\,\|\mathbf{K}\|_F,
\]
and therefore
\[
   2^{\,b_K - b_V} \;\approx\; \frac{\|\mathbf{V}\|_F}{\|\mathbf{K}\|_F}.
\]
Since typically \(\|\mathbf{V}\|_F < \|\mathbf{K}\|_F\), it follows that \(b_K > b_V\). This reinforces the earlier spectral-norm result: keys should be allocated more bits than values to achieve balanced quantization error under an MSE criterion.

\section{Supplemental Results}
\label{sec:supplemental}

\subsection{Focus on Generative Tasks}
\label{sec:why_generative_tasks}

We validate and operationalize our theorems across a diverse set of datasets, including C4, MMLU, GSM8K, EQ-Bench, CoQA, and LongBench.
Our evaluation purposefully focuses on \emph{generative} tasks, i.e., open-ended generation and free-form responses, because KV-cache quantization primarily affects the \emph{decoding phase} rather than the \emph{prefill phase}. To assess quantization error in isolation, we use C4 under a free-form decoding setup that mirrors pretraining usage, enabling us to analyze how compression directly distorts activations. To quantify downstream impact, we select GSM8K, CoQA, EQ-Bench, and LongBench, all of which rely on \texttt{generate\_until}-style evaluation, where the model must produce extended, structured responses. In contrast, commonsense reasoning benchmarks (e.g., BoolQ, PIQA, HellaSwag, or DoRA) use log-likelihood scoring over candidate options and do not engage KV-cache quantization unless the input prompt itself is compressed. Consequently, such discriminative evaluations are orthogonal to our focus and were intentionally excluded.%

Within LongBench, we surveyed all subtasks (gov\_report, lcc, lsht, multi\_news, narrativeqa, qasper, qmsum, repobench-p, samsum) and found that only \texttt{gov\_report} and \texttt{qmsum} require substantial generation, with \texttt{gov\_report} involving the longest outputs. Throughout this work, “LongBench” refers specifically to \texttt{gov\_report}.

\subsection{Key-Value Weight Norm}
\label{appendix:kvw}

\begin{figure}[t]
  \centering
  \includegraphics[width=\linewidth]{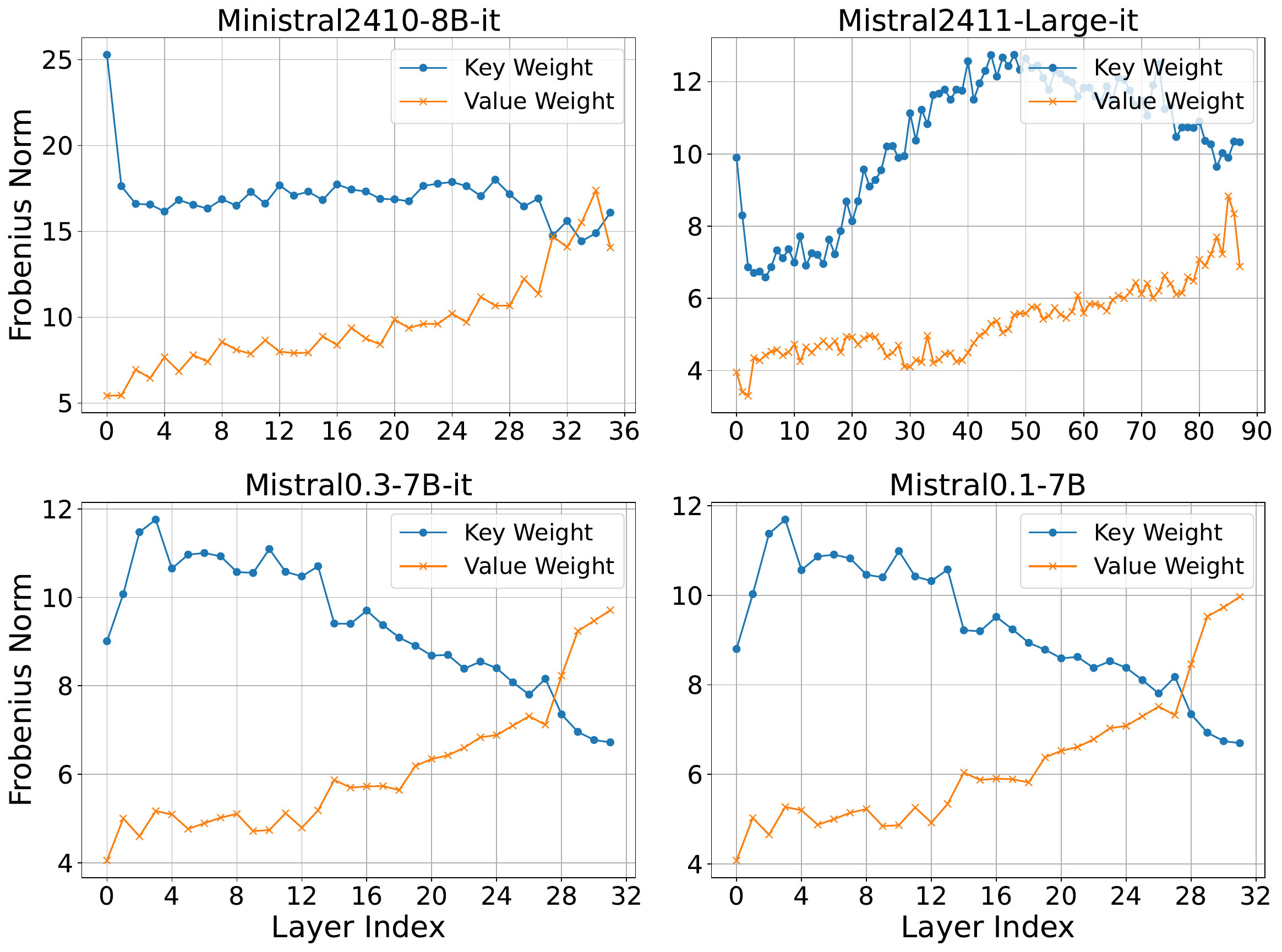}
  \caption{\textbf{Frobenius norm plot for Mistral Family.} The x-axis represents the layer index in the model, while the y-axis represents the Frobenius norm magnitude. The spectral norms are higher for the key weights than for the value weights across layers.}
  \label{fig:frobweights_mistral}
\end{figure}

Figure~\ref{fig:frobweights_mistral} presents the Frobenius norms of key and value weights for the Mistral family, exhibiting the same pattern (keys consistently having higher norms than values) as shown in Figure~\ref{fig:frobweights_llama} for the Llama family, further corroborating the Key-Value Norm Disparity theorem established in Section~\ref{sec:kvnd}.

\subsection{Singular Value Distributions}
\label{sec:appendix-sing-all}

\begin{figure*}[htbp]
\centering
\includegraphics[width=\textwidth]{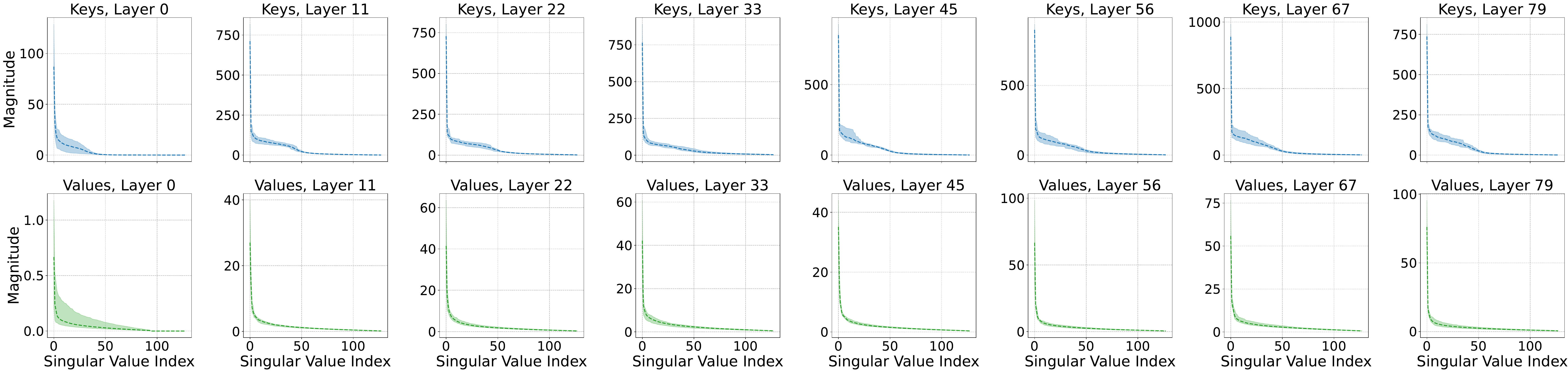}
\caption{\textbf{Complete singular value distribution of key and value cache} for Llama 3.3-70B on the C4 dataset. The x-axis denotes the singular value indices, ordered from the largest (spectral norm) to the smallest, while the y-axis represents the corresponding magnitudes. The shaded region illustrates the range between the minimum and maximum singular values across attention heads within each layer, and the solid lines indicate the mean singular value magnitude at each index. This full-spectrum view highlights that key matrices consistently maintain significantly higher singular values throughout the entire distribution, further reinforcing their dominant representational capacity compared to value matrices.}
\label{fig:appendix_singular_value_distribution}
\end{figure*}

Figure~\ref{fig:appendix_singular_value_distribution} illustrates the full-spectrum singular value distribution of key and value caches across layers of the Llama~3.3~70B model on the C4 dataset. The horizontal axis indexes singular values in descending order, starting from the largest (i.e., the spectral norm), while the vertical axis shows their magnitudes. The shaded region represents the minimum-maximum range of singular values across attention heads within each layer, and the solid curves indicate the mean singular value at each rank.

\begin{figure}[htbp]
    \centering
    \begin{subfigure}[t]{0.8\linewidth}
        \centering
        \includegraphics[width=\linewidth]{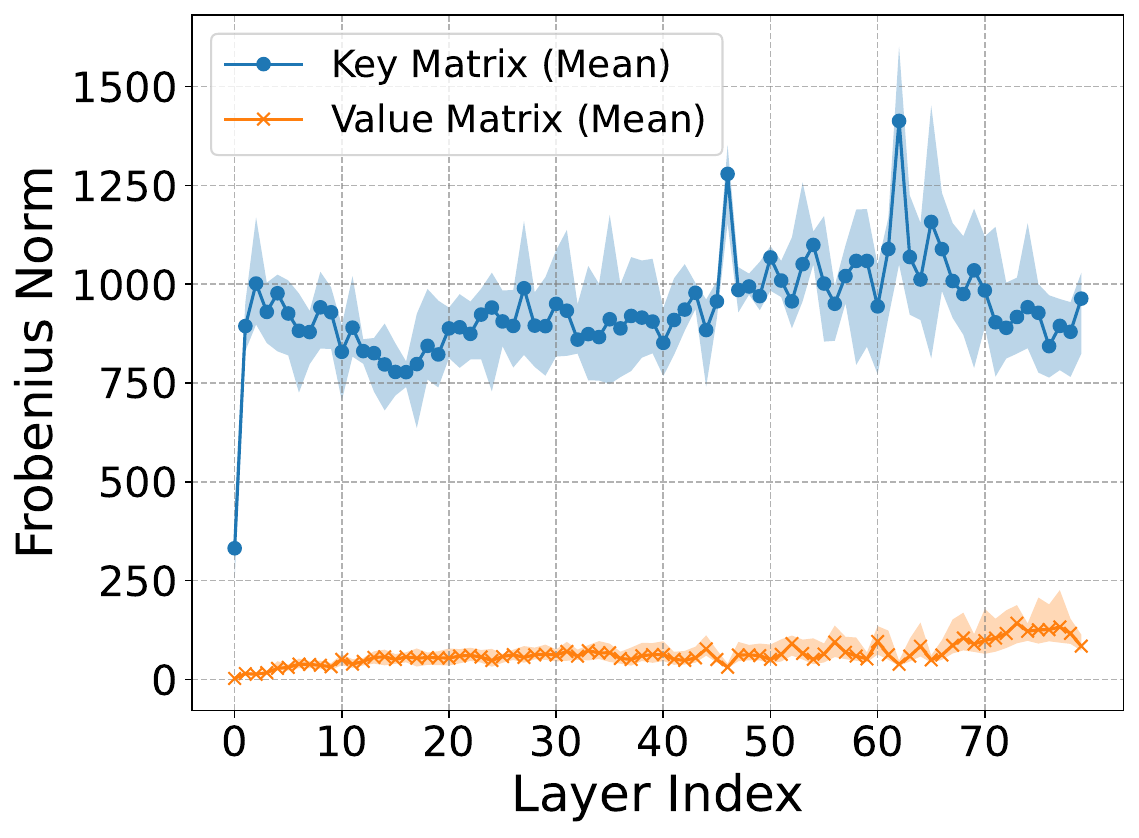}
        \caption{Frobenius Norm}
        \label{fig:append_frobenius_norm_sub}
    \end{subfigure}\hfill
    \begin{subfigure}[t]{0.8\linewidth}
        \centering
        \includegraphics[width=\linewidth]{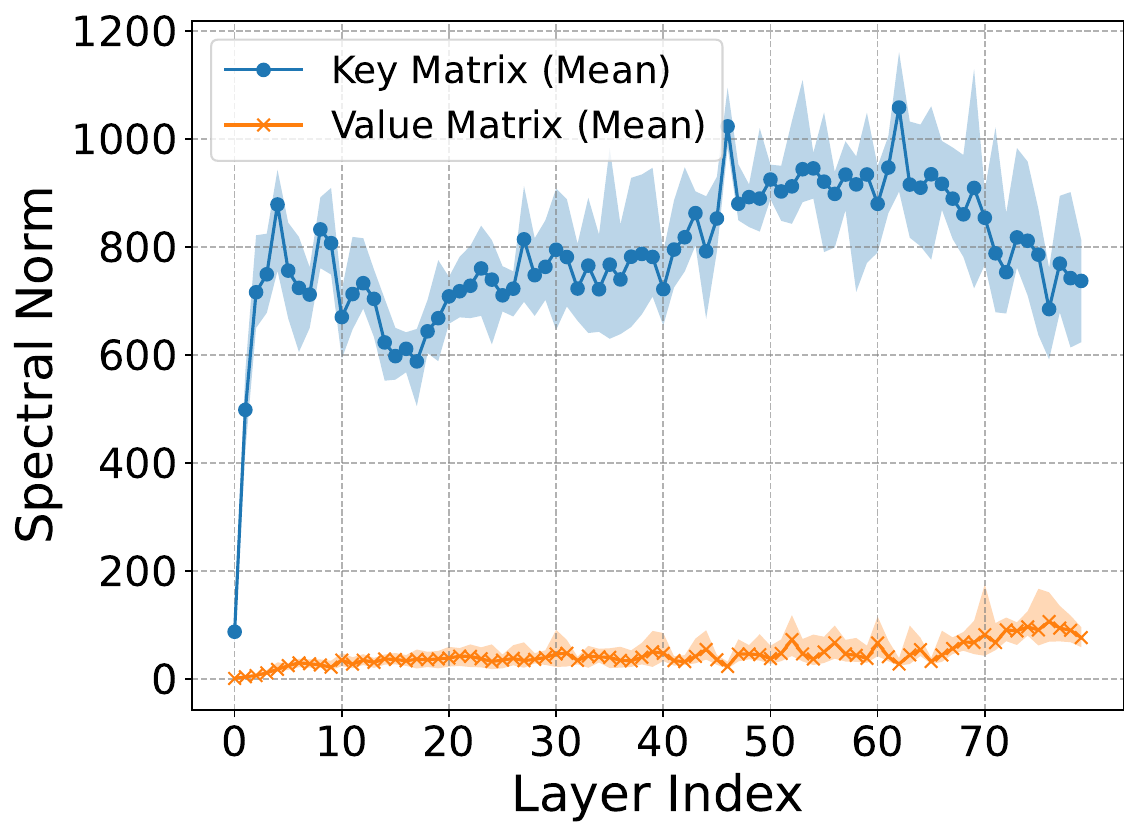}
        \caption{Spectral Norm}
        \label{fig:append_spectral_norm_sub}
    \end{subfigure}
    \caption{
    \textbf{Frobenius and spectral norms of key and value caches across layers for the Llama~3.3~70B model on the C4 dataset.} 
    The x-axis represents the layer index. The shaded regions indicate the min-max range across attention heads within each layer, while the solid curves represent the mean norm per layer. 
    In both cases, key matrices consistently exhibit substantially higher norms than value matrices, reflecting their stronger representational capacity.
    }
    \label{fig:append_norms_frob_spectral}
\end{figure}

\subsection{Quantization Error}
\label{appendix-d:quant-error}

By using quantization error, we refer to evaluating how closely the quantized-then-dequantized key/value (KV) caches reconstruct their full-precision counterparts, measured as the reconstruction mean-squared error (MSE), which is exactly the Frobenius norm error normalized by the number of elements. This serves as a practical proxy for downstream quality because uniform \( b \)-bit quantization admits norm-based bounds in which reconstruction error scales with the cache norm and decays roughly like \( 2^{-b} \); thus, at a fixed bit budget, higher-norm caches incur larger distortion, and bit allocations that minimize reconstruction error are directly targeting the quantity most affected by quantization.

\begin{table*}[htbp]
\centering
\setlength{\tabcolsep}{5pt} 
\caption{Quantization error (mean ± std) for the key and value caches (\(\mathbf{K}_i\) and \(\mathbf{V}_i\)) at 2-bit, 3-bit, and 4-bit quantization, evaluated on the \textbf{MMLU} dataset.}
\label{tab:mmlu_loss_bits}
\begin{tabular}{l>{\columncolor[gray]{0.9}}c c >{\columncolor[gray]{0.9}}c c >{\columncolor[gray]{0.9}}c c}
\toprule
& \(\mathbf{K}_2\) & \(\mathbf{V}_2\) & \(\mathbf{K}_3\) & \(\mathbf{V}_3\) & \(\mathbf{K}_4\) & \(\mathbf{V}_4\) \\
\midrule
Llama3.2-1B         & 4.851 {\color{gray}{\scriptsize ± 1.037}} & 0.127 {\color{gray}{\scriptsize ± 0.101}} & 1.037 {\color{gray}{\scriptsize ± 0.265}} & 0.021 {\color{gray}{\scriptsize ± 0.015}} & 0.227 {\color{gray}{\scriptsize ± 0.059}} & 0.005 {\color{gray}{\scriptsize ± 0.003}} \\
Llama3.2-1B-it      & 4.373 {\color{gray}{\scriptsize ± 1.034}} & 0.124 {\color{gray}{\scriptsize ± 0.090}} & 0.879 {\color{gray}{\scriptsize ± 0.218}} & 0.019 {\color{gray}{\scriptsize ± 0.013}} & 0.192 {\color{gray}{\scriptsize ± 0.047}} & 0.004 {\color{gray}{\scriptsize ± 0.003}} \\
Llama3.2-3B         & 3.943 {\color{gray}{\scriptsize ± 0.924}} & 0.193 {\color{gray}{\scriptsize ± 0.096}} & 0.849 {\color{gray}{\scriptsize ± 0.150}} & 0.030 {\color{gray}{\scriptsize ± 0.015}} & 0.183 {\color{gray}{\scriptsize ± 0.031}} & 0.007 {\color{gray}{\scriptsize ± 0.003}} \\
Llama3.2-3B-it      & 4.487 {\color{gray}{\scriptsize ± 1.180}} & 0.202 {\color{gray}{\scriptsize ± 0.100}} & 0.894 {\color{gray}{\scriptsize ± 0.180}} & 0.030 {\color{gray}{\scriptsize ± 0.015}} & 0.193 {\color{gray}{\scriptsize ± 0.037}} & 0.007 {\color{gray}{\scriptsize ± 0.003}} \\
Llama2-7B           & 3.190 {\color{gray}{\scriptsize ± 0.783}} & 0.259 {\color{gray}{\scriptsize ± 0.184}} & 0.769 {\color{gray}{\scriptsize ± 0.194}} & 0.042 {\color{gray}{\scriptsize ± 0.030}} & 0.168 {\color{gray}{\scriptsize ± 0.042}} & 0.009 {\color{gray}{\scriptsize ± 0.006}} \\
Llama3.1-8B-it      & 6.003 {\color{gray}{\scriptsize ± 1.782}} & 0.187 {\color{gray}{\scriptsize ± 0.127}} & 1.082 {\color{gray}{\scriptsize ± 0.244}} & 0.028 {\color{gray}{\scriptsize ± 0.019}} & 0.235 {\color{gray}{\scriptsize ± 0.055}} & 0.006 {\color{gray}{\scriptsize ± 0.004}} \\
Llama3.3-70B-it     & 4.883 {\color{gray}{\scriptsize ± 1.106}} & 0.112 {\color{gray}{\scriptsize ± 0.093}} & 0.942 {\color{gray}{\scriptsize ± 0.198}} & 0.016 {\color{gray}{\scriptsize ± 0.012}} & 0.206 {\color{gray}{\scriptsize ± 0.043}} & 0.003 {\color{gray}{\scriptsize ± 0.003}} \\
Nemotron3.1-it      & 5.125 {\color{gray}{\scriptsize ± 1.284}} & 0.114 {\color{gray}{\scriptsize ± 0.094}} & 0.985 {\color{gray}{\scriptsize ± 0.207}} & 0.016 {\color{gray}{\scriptsize ± 0.012}} & 0.216 {\color{gray}{\scriptsize ± 0.046}} & 0.003 {\color{gray}{\scriptsize ± 0.003}} \\
Phi3-Medium-128K-it & 5.063 {\color{gray}{\scriptsize ± 1.914}} & 0.584 {\color{gray}{\scriptsize ± 0.559}} & 1.000 {\color{gray}{\scriptsize ± 0.319}} & 0.087 {\color{gray}{\scriptsize ± 0.083}} & 0.217 {\color{gray}{\scriptsize ± 0.068}} & 0.019 {\color{gray}{\scriptsize ± 0.018}} \\
Phi4                & 5.929 {\color{gray}{\scriptsize ± 1.545}} & 0.657 {\color{gray}{\scriptsize ± 0.472}} & 1.306 {\color{gray}{\scriptsize ± 0.231}} & 0.103 {\color{gray}{\scriptsize ± 0.070}} & 0.286 {\color{gray}{\scriptsize ± 0.050}} & 0.022 {\color{gray}{\scriptsize ± 0.015}} \\
Mistral0.3-7B       & 4.718 {\color{gray}{\scriptsize ± 1.340}} & 0.398 {\color{gray}{\scriptsize ± 0.405}} & 0.941 {\color{gray}{\scriptsize ± 0.240}} & 0.059 {\color{gray}{\scriptsize ± 0.059}} & 0.206 {\color{gray}{\scriptsize ± 0.053}} & 0.013 {\color{gray}{\scriptsize ± 0.013}} \\
Qwen2.5-14B         & 5.184 {\color{gray}{\scriptsize ± 2.241}} & 1.270 {\color{gray}{\scriptsize ± 1.547}} & 1.005 {\color{gray}{\scriptsize ± 0.288}} & 0.182 {\color{gray}{\scriptsize ± 0.221}} & 0.223 {\color{gray}{\scriptsize ± 0.067}} & 0.040 {\color{gray}{\scriptsize ± 0.052}} \\
DeepSeekR1L-8B      & 5.502 {\color{gray}{\scriptsize ± 1.549}} & 0.189 {\color{gray}{\scriptsize ± 0.118}} & 0.955 {\color{gray}{\scriptsize ± 0.204}} & 0.028 {\color{gray}{\scriptsize ± 0.017}} & 0.209 {\color{gray}{\scriptsize ± 0.046}} & 0.006 {\color{gray}{\scriptsize ± 0.004}} \\
DeepSeekR1Q-14B     & 5.126 {\color{gray}{\scriptsize ± 2.375}} & 1.406 {\color{gray}{\scriptsize ± 1.609}} & 0.900 {\color{gray}{\scriptsize ± 0.269}} & 0.198 {\color{gray}{\scriptsize ± 0.226}} & 0.199 {\color{gray}{\scriptsize ± 0.062}} & 0.044 {\color{gray}{\scriptsize ± 0.052}} \\
\bottomrule
\end{tabular}
\end{table*}

\begin{table*}[htbp]
\centering
\setlength{\tabcolsep}{5pt}
\caption{Quantization error (mean ± std) for the key and value caches (\(\mathbf{K}_i\) and \(\mathbf{V}_i\)) at 2-bit, 3-bit, and 4-bit quantization, evaluated on the \textbf{C4} dataset.}
\label{tab:c4_loss_bits}
\begin{tabular}{l>{\columncolor[gray]{0.9}}c c >{\columncolor[gray]{0.9}}c c >{\columncolor[gray]{0.9}}c c}
\toprule
& \(\mathbf{K}_2\) & \(\mathbf{V}_2\) & \(\mathbf{K}_3\) & \(\mathbf{V}_3\) & \(\mathbf{K}_4\) & \(\mathbf{V}_4\) \\
\midrule
Llama3.2-1B         & 4.885 {\color{gray}{\scriptsize ± 1.056}} & 0.207 {\color{gray}{\scriptsize ± 0.166}} & 1.074 {\color{gray}{\scriptsize ± 0.289}} & 0.030 {\color{gray}{\scriptsize ± 0.024}} & 0.233 {\color{gray}{\scriptsize ± 0.062}} & 0.006 {\color{gray}{\scriptsize ± 0.005}} \\
Llama3.2-1B-it      & 4.524 {\color{gray}{\scriptsize ± 1.108}} & 0.193 {\color{gray}{\scriptsize ± 0.137}} & 0.925 {\color{gray}{\scriptsize ± 0.235}} & 0.028 {\color{gray}{\scriptsize ± 0.020}} & 0.201 {\color{gray}{\scriptsize ± 0.050}} & 0.006 {\color{gray}{\scriptsize ± 0.005}} \\
Llama3.2-3B         & 3.885 {\color{gray}{\scriptsize ± 0.777}} & 0.282 {\color{gray}{\scriptsize ± 0.150}} & 0.909 {\color{gray}{\scriptsize ± 0.168}} & 0.042 {\color{gray}{\scriptsize ± 0.023}} & 0.194 {\color{gray}{\scriptsize ± 0.032}} & 0.009 {\color{gray}{\scriptsize ± 0.005}} \\
Llama3.2-3B-it      & 4.135 {\color{gray}{\scriptsize ± 1.088}} & 0.274 {\color{gray}{\scriptsize ± 0.137}} & 0.912 {\color{gray}{\scriptsize ± 0.176}} & 0.039 {\color{gray}{\scriptsize ± 0.020}} & 0.195 {\color{gray}{\scriptsize ± 0.036}} & 0.009 {\color{gray}{\scriptsize ± 0.004}} \\
Llama2-7B           & 6.337 {\color{gray}{\scriptsize ± 1.710}} & 0.456 {\color{gray}{\scriptsize ± 0.247}} & 1.054 {\color{gray}{\scriptsize ± 0.263}} & 0.071 {\color{gray}{\scriptsize ± 0.038}} & 0.213 {\color{gray}{\scriptsize ± 0.052}} & 0.015 {\color{gray}{\scriptsize ± 0.008}} \\
Llama3.1-8B-it      & 6.262 {\color{gray}{\scriptsize ± 1.789}} & 0.254 {\color{gray}{\scriptsize ± 0.185}} & 1.128 {\color{gray}{\scriptsize ± 0.249}} & 0.036 {\color{gray}{\scriptsize ± 0.026}} & 0.247 {\color{gray}{\scriptsize ± 0.056}} & 0.008 {\color{gray}{\scriptsize ± 0.005}} \\
Llama3.3-70B-it     & 4.391 {\color{gray}{\scriptsize ± 1.027}} & 0.121 {\color{gray}{\scriptsize ± 0.097}} & 0.847 {\color{gray}{\scriptsize ± 0.175}} & 0.017 {\color{gray}{\scriptsize ± 0.013}} & 0.186 {\color{gray}{\scriptsize ± 0.038}} & 0.004 {\color{gray}{\scriptsize ± 0.003}} \\
Nemotron3.1-it      & 5.367 {\color{gray}{\scriptsize ± 1.332}} & 0.127 {\color{gray}{\scriptsize ± 0.105}} & 1.049 {\color{gray}{\scriptsize ± 0.222}} & 0.018 {\color{gray}{\scriptsize ± 0.013}} & 0.231 {\color{gray}{\scriptsize ± 0.049}} & 0.004 {\color{gray}{\scriptsize ± 0.003}} \\
Phi3-Medium-128K-it & 4.831 {\color{gray}{\scriptsize ± 1.759}} & 0.788 {\color{gray}{\scriptsize ± 0.726}} & 1.022 {\color{gray}{\scriptsize ± 0.306}} & 0.109 {\color{gray}{\scriptsize ± 0.097}} & 0.220 {\color{gray}{\scriptsize ± 0.064}} & 0.023 {\color{gray}{\scriptsize ± 0.021}} \\
Phi4                & 5.715 {\color{gray}{\scriptsize ± 1.442}} & 0.850 {\color{gray}{\scriptsize ± 0.684}} & 1.316 {\color{gray}{\scriptsize ± 0.245}} & 0.124 {\color{gray}{\scriptsize ± 0.093}} & 0.291 {\color{gray}{\scriptsize ± 0.056}} & 0.027 {\color{gray}{\scriptsize ± 0.020}} \\
Mistral0.3-7B       & 5.027 {\color{gray}{\scriptsize ± 1.332}} & 0.543 {\color{gray}{\scriptsize ± 0.493}} & 1.014 {\color{gray}{\scriptsize ± 0.269}} & 0.079 {\color{gray}{\scriptsize ± 0.068}} & 0.223 {\color{gray}{\scriptsize ± 0.060}} & 0.017 {\color{gray}{\scriptsize ± 0.015}} \\
Qwen2.5-14B         & 4.382 {\color{gray}{\scriptsize ± 2.170}} & 1.544 {\color{gray}{\scriptsize ± 1.872}} & 0.846 {\color{gray}{\scriptsize ± 0.250}} & 0.220 {\color{gray}{\scriptsize ± 0.265}} & 0.187 {\color{gray}{\scriptsize ± 0.060}} & 0.048 {\color{gray}{\scriptsize ± 0.060}} \\
DeepSeekR1L-8B      & 4.575 {\color{gray}{\scriptsize ± 1.122}} & 0.204 {\color{gray}{\scriptsize ± 0.134}} & 0.817 {\color{gray}{\scriptsize ± 0.141}} & 0.030 {\color{gray}{\scriptsize ± 0.019}} & 0.179 {\color{gray}{\scriptsize ± 0.033}} & 0.006 {\color{gray}{\scriptsize ± 0.004}} \\
DeepSeekR1Q-14B     & 4.832 {\color{gray}{\scriptsize ± 2.354}} & 1.651 {\color{gray}{\scriptsize ± 1.914}} & 0.927 {\color{gray}{\scriptsize ± 0.283}} & 0.232 {\color{gray}{\scriptsize ± 0.267}} & 0.201 {\color{gray}{\scriptsize ± 0.061}} & 0.051 {\color{gray}{\scriptsize ± 0.060}} \\
\bottomrule
\end{tabular}
\end{table*}

\begin{table*}[htbp]
\centering
\setlength{\tabcolsep}{5pt}
\caption{Quantization error (mean ± std) for the key and value caches ($K_i$ and $V_i$) at 2-bit, 3-bit, and 4-bit quantization, evaluated on the GSM8K dataset.}
\label{tab:gsm8k_loss_qbits}
\begin{tabular}{l>{\columncolor[gray]{0.9}}l l >{\columncolor[gray]{0.9}}l l >{\columncolor[gray]{0.9}}l l}
\toprule
& \(\mathbf{K}_2\) & \(\mathbf{V}_2\) & \(\mathbf{K}_3\) & \(\mathbf{V}_3\) & \(\mathbf{K}_4\) & \(\mathbf{V}_4\) \\
\midrule
Llama3.2-1B          & 5.703 {\color{gray}{\scriptsize ± 1.557}} & 0.179 {\color{gray}{\scriptsize ± 0.136}} & 1.213 {\color{gray}{\scriptsize ± 0.352}} & 0.026 {\color{gray}{\scriptsize ± 0.020}} & 0.266 {\color{gray}{\scriptsize ± 0.078}} & 0.005 {\color{gray}{\scriptsize ± 0.004}} \\
Llama3.2-1B-it       & 5.002 {\color{gray}{\scriptsize ± 1.383}} & 0.171 {\color{gray}{\scriptsize ± 0.130}} & 1.024 {\color{gray}{\scriptsize ± 0.287}} & 0.025 {\color{gray}{\scriptsize ± 0.020}} & 0.223 {\color{gray}{\scriptsize ± 0.061}} & 0.006 {\color{gray}{\scriptsize ± 0.004}} \\
Llama3.2-3B          & 4.840 {\color{gray}{\scriptsize ± 1.396}} & 0.261 {\color{gray}{\scriptsize ± 0.136}} & 1.045 {\color{gray}{\scriptsize ± 0.211}} & 0.038 {\color{gray}{\scriptsize ± 0.021}} & 0.226 {\color{gray}{\scriptsize ± 0.044}} & 0.008 {\color{gray}{\scriptsize ± 0.005}} \\
Llama3.2-3B-it       & 3.604 {\color{gray}{\scriptsize ± 0.850}} & 0.226 {\color{gray}{\scriptsize ± 0.129}} & 0.790 {\color{gray}{\scriptsize ± 0.135}} & 0.034 {\color{gray}{\scriptsize ± 0.019}} & 0.171 {\color{gray}{\scriptsize ± 0.028}} & 0.007 {\color{gray}{\scriptsize ± 0.004}} \\
Llama2-7B            & 5.081 {\color{gray}{\scriptsize ± 1.396}} & 0.405 {\color{gray}{\scriptsize ± 0.231}} & 0.969 {\color{gray}{\scriptsize ± 0.238}} & 0.065 {\color{gray}{\scriptsize ± 0.037}} & 0.205 {\color{gray}{\scriptsize ± 0.050}} & 0.014 {\color{gray}{\scriptsize ± 0.008}} \\
Llama3.1-8B-it       & 6.445 {\color{gray}{\scriptsize ± 1.837}} & 0.213 {\color{gray}{\scriptsize ± 0.161}} & 1.184 {\color{gray}{\scriptsize ± 0.268}} & 0.030 {\color{gray}{\scriptsize ± 0.022}} & 0.257 {\color{gray}{\scriptsize ± 0.060}} & 0.007 {\color{gray}{\scriptsize ± 0.005}} \\
Llama3.3-70B-it      & 4.967 {\color{gray}{\scriptsize ± 1.127}} & 0.113 {\color{gray}{\scriptsize ± 0.091}} & 0.978 {\color{gray}{\scriptsize ± 0.203}} & 0.016 {\color{gray}{\scriptsize ± 0.012}} & 0.214 {\color{gray}{\scriptsize ± 0.044}} & 0.004 {\color{gray}{\scriptsize ± 0.003}} \\
Nemotron3.1-it       & 4.752 {\color{gray}{\scriptsize ± 1.124}} & 0.113 {\color{gray}{\scriptsize ± 0.089}} & 0.940 {\color{gray}{\scriptsize ± 0.194}} & 0.016 {\color{gray}{\scriptsize ± 0.012}} & 0.206 {\color{gray}{\scriptsize ± 0.042}} & 0.004 {\color{gray}{\scriptsize ± 0.003}} \\
Phi3-Medium-128K-it  & 4.940 {\color{gray}{\scriptsize ± 1.834}} & 0.605 {\color{gray}{\scriptsize ± 0.579}} & 1.042 {\color{gray}{\scriptsize ± 0.320}} & 0.088 {\color{gray}{\scriptsize ± 0.082}} & 0.227 {\color{gray}{\scriptsize ± 0.069}} & 0.019 {\color{gray}{\scriptsize ± 0.018}} \\
Phi4                 & 6.610 {\color{gray}{\scriptsize ± 1.624}} & 0.785 {\color{gray}{\scriptsize ± 0.598}} & 1.498 {\color{gray}{\scriptsize ± 0.293}} & 0.116 {\color{gray}{\scriptsize ± 0.082}} & 0.330 {\color{gray}{\scriptsize ± 0.064}} & 0.025 {\color{gray}{\scriptsize ± 0.017}} \\
Mistral0.3-7B        & 5.308 {\color{gray}{\scriptsize ± 1.367}} & 0.461 {\color{gray}{\scriptsize ± 0.434}} & 1.065 {\color{gray}{\scriptsize ± 0.288}} & 0.067 {\color{gray}{\scriptsize ± 0.061}} & 0.232 {\color{gray}{\scriptsize ± 0.061}} & 0.015 {\color{gray}{\scriptsize ± 0.013}} \\
Qwen2.5-14B          & 4.829 {\color{gray}{\scriptsize ± 2.179}} & 1.736 {\color{gray}{\scriptsize ± 2.659}} & 0.979 {\color{gray}{\scriptsize ± 0.264}} & 0.241 {\color{gray}{\scriptsize ± 0.372}} & 0.214 {\color{gray}{\scriptsize ± 0.061}} & 0.051 {\color{gray}{\scriptsize ± 0.077}} \\
DeepSeekR1L-8B       & 5.547 {\color{gray}{\scriptsize ± 1.517}} & 0.193 {\color{gray}{\scriptsize ± 0.129}} & 1.000 {\color{gray}{\scriptsize ± 0.212}} & 0.028 {\color{gray}{\scriptsize ± 0.018}} & 0.218 {\color{gray}{\scriptsize ± 0.049}} & 0.006 {\color{gray}{\scriptsize ± 0.004}} \\
DeepSeekR1Q-14B      & 4.477 {\color{gray}{\scriptsize ± 2.176}} & 1.424 {\color{gray}{\scriptsize ± 1.752}} & 0.830 {\color{gray}{\scriptsize ± 0.256}} & 0.200 {\color{gray}{\scriptsize ± 0.242}} & 0.181 {\color{gray}{\scriptsize ± 0.058}} & 0.044 {\color{gray}{\scriptsize ± 0.056}} \\
\bottomrule
\end{tabular}
\end{table*}

\begin{figure}[htbp]
    \centering
    \includegraphics[width=0.8\linewidth]{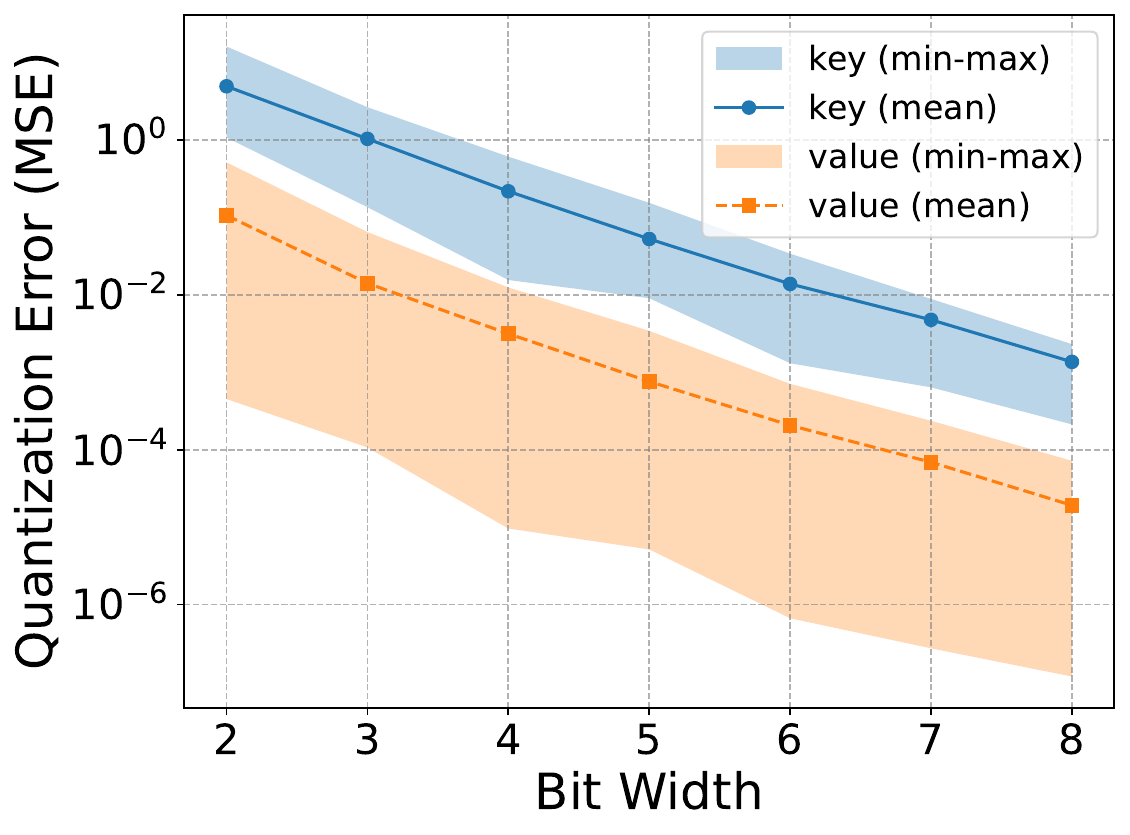}
    \caption{\textbf{MSE of KV quantization for Llama 3.3-70B on the C4 dataset.} We use quantization bit-widths ranging from 2 to 8. The x-axis represents the quantization bit-width, while the y-axis shows the MSE on a logarithmic scale. A logarithmic scale is used to highlight differences at higher bit-widths, where MSE values decrease significantly and approach zero, particularly at 8-bit. Solid lines indicate the mean MSE across layers, while the shaded regions represent the min-max range of errors.}
    \label{fig:bit_error}
\end{figure}

\begin{figure*}[h]
    \centering
    \begin{subfigure}{0.9\linewidth}
        \centering
        \includegraphics[width=\linewidth]{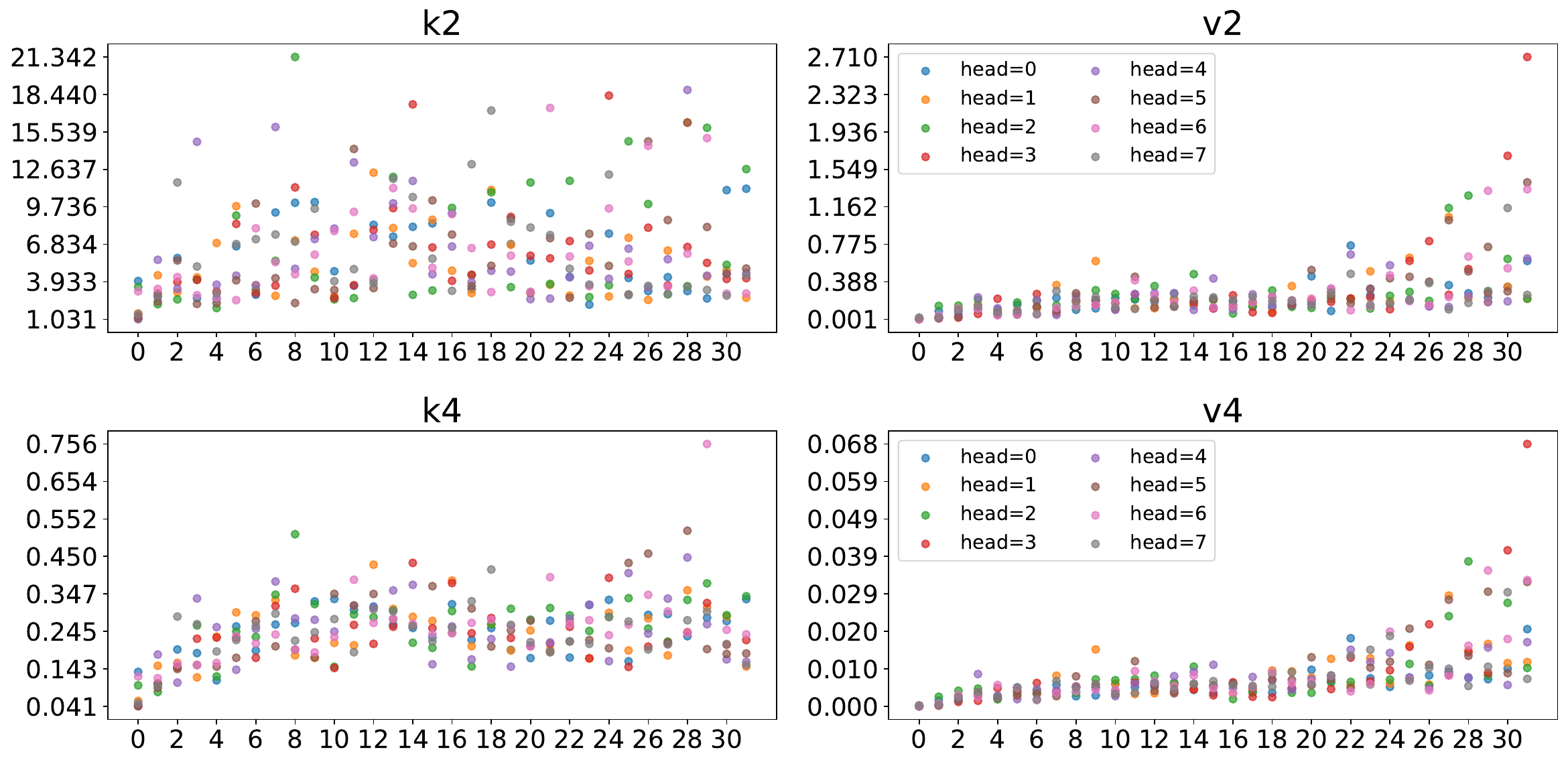}
        
        \caption{C4}
        \label{fig:c4-plot}
    \end{subfigure}
    \begin{subfigure}{0.9\linewidth}
        \centering
        \includegraphics[width=\linewidth]{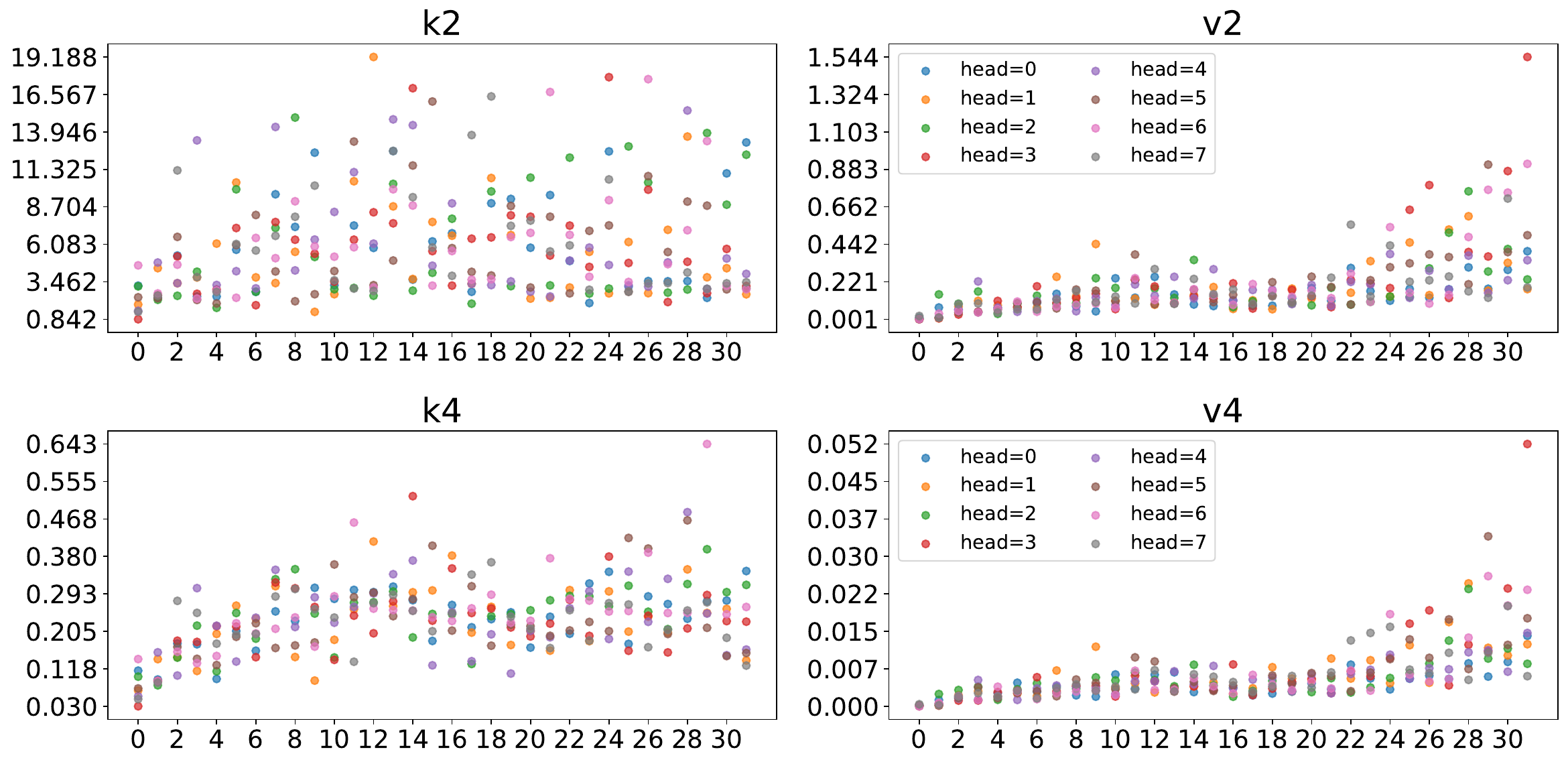}
        \caption{MMLU}
        \label{fig:mmlu-plot}
    \end{subfigure}
    \begin{subfigure}{0.9\linewidth}
        \centering
        \includegraphics[width=\linewidth]{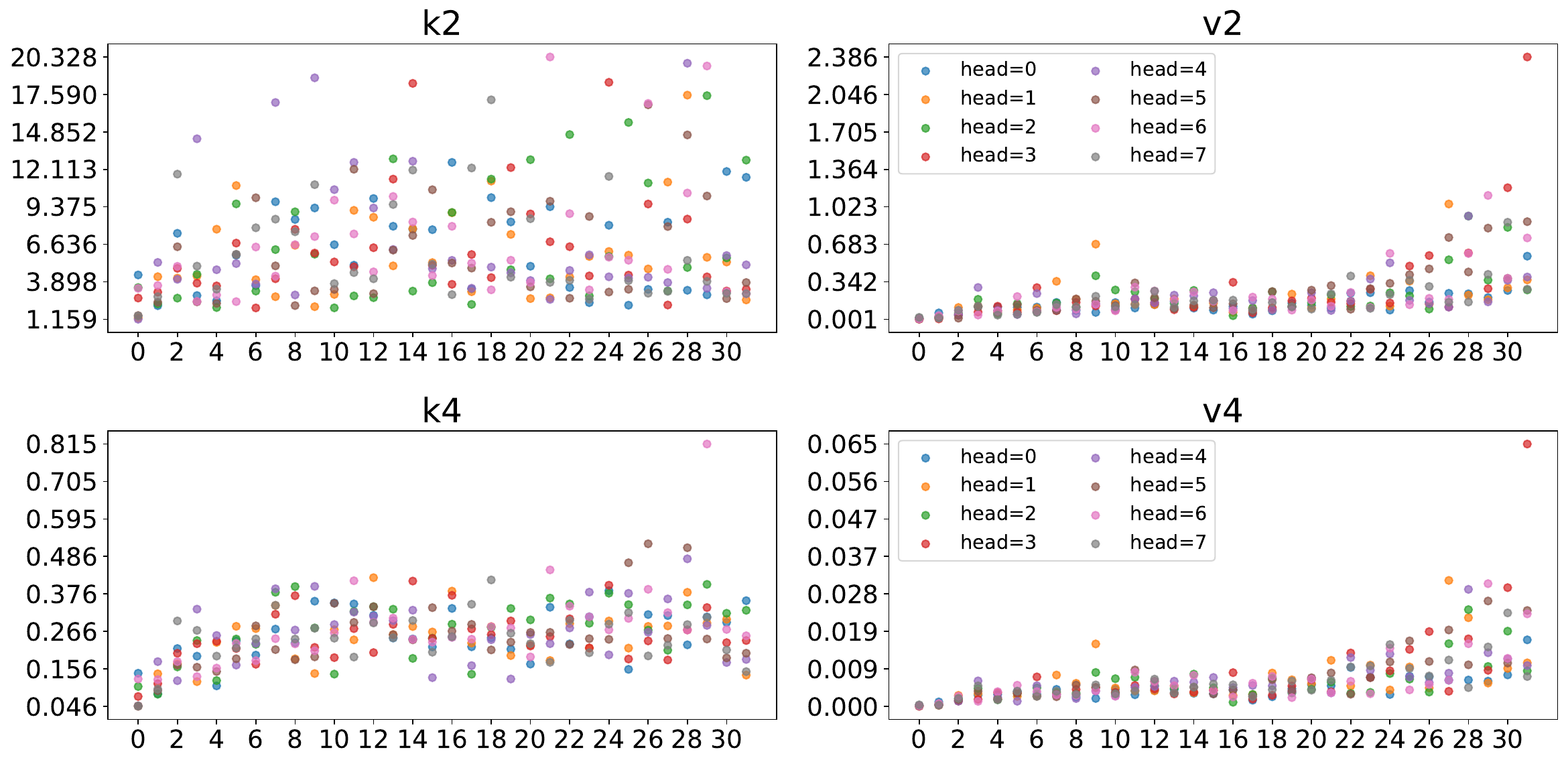}
        
        \caption{GSM8k}
        \label{fig:gsm8k-plot}
    \end{subfigure}
    \caption{\textbf{Quantization error (MSE) of key and value caches in the Llama~3.1~8B model across 32 layers for (a)~C4, (b)~MMLU, and (c)~GSM8K.} 
    The top row shows 2-bit quantization (\(\mathrm{K}_2\), \(\mathrm{V}_2\)), and the bottom row shows 4-bit quantization (\(\mathrm{K}_4\), \(\mathrm{V}_4\)). 
    Each point corresponds to an attention head within the respective layer. 
    The x-axis denotes the layer index, and the y-axis indicates the quantization error (MSE).}
    \label{fig:layer-head-err}
\end{figure*}

\begin{figure*}[H]
    \centering
    \begin{subfigure}{.9\textwidth}
        \centering
        \includegraphics[width=\linewidth]{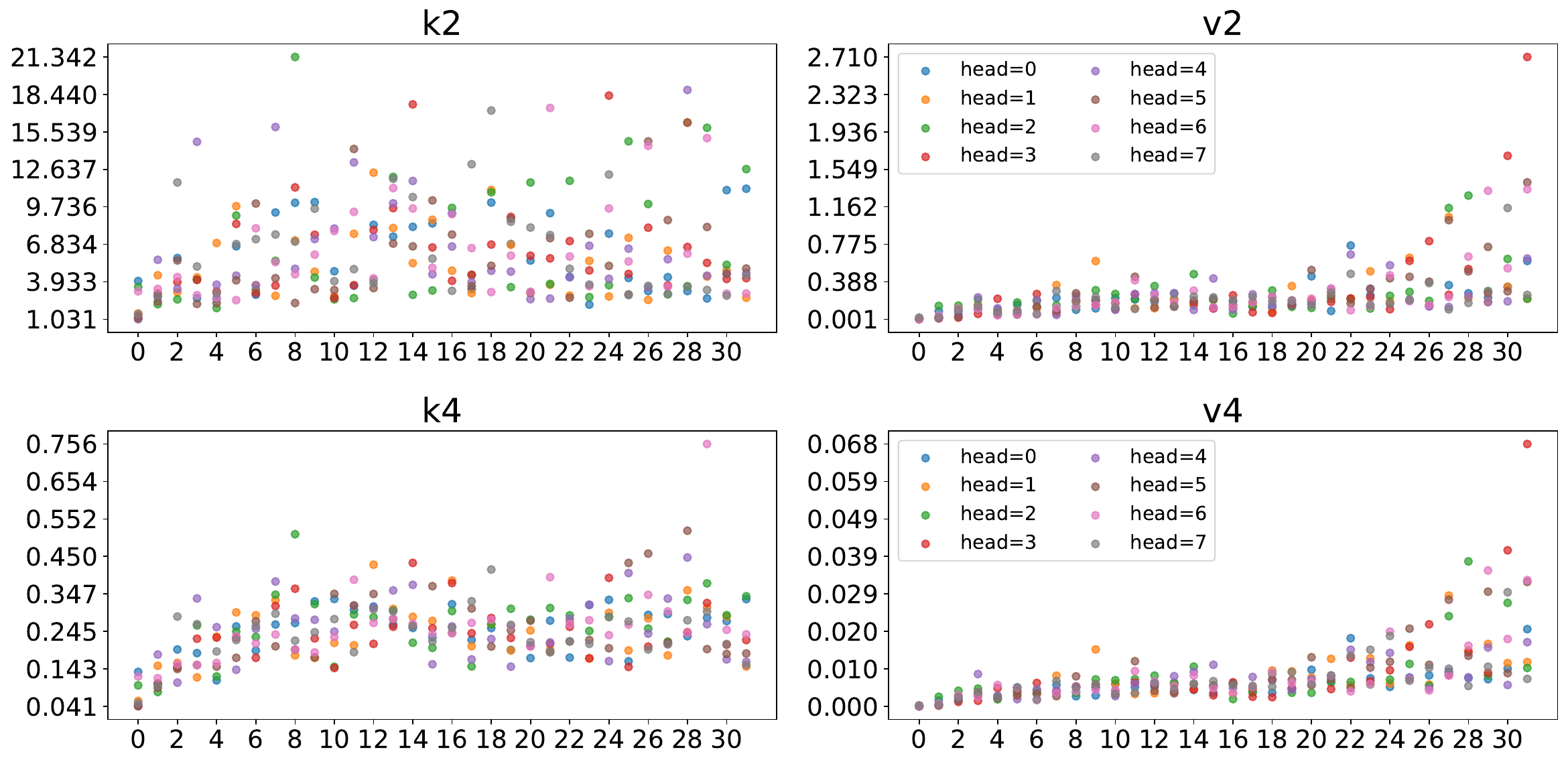}
        \caption{C4 dataset}
        \label{fig:c4-plot}
    \end{subfigure}
    \begin{subfigure}{.9\textwidth}
        \centering
        \includegraphics[width=\linewidth]{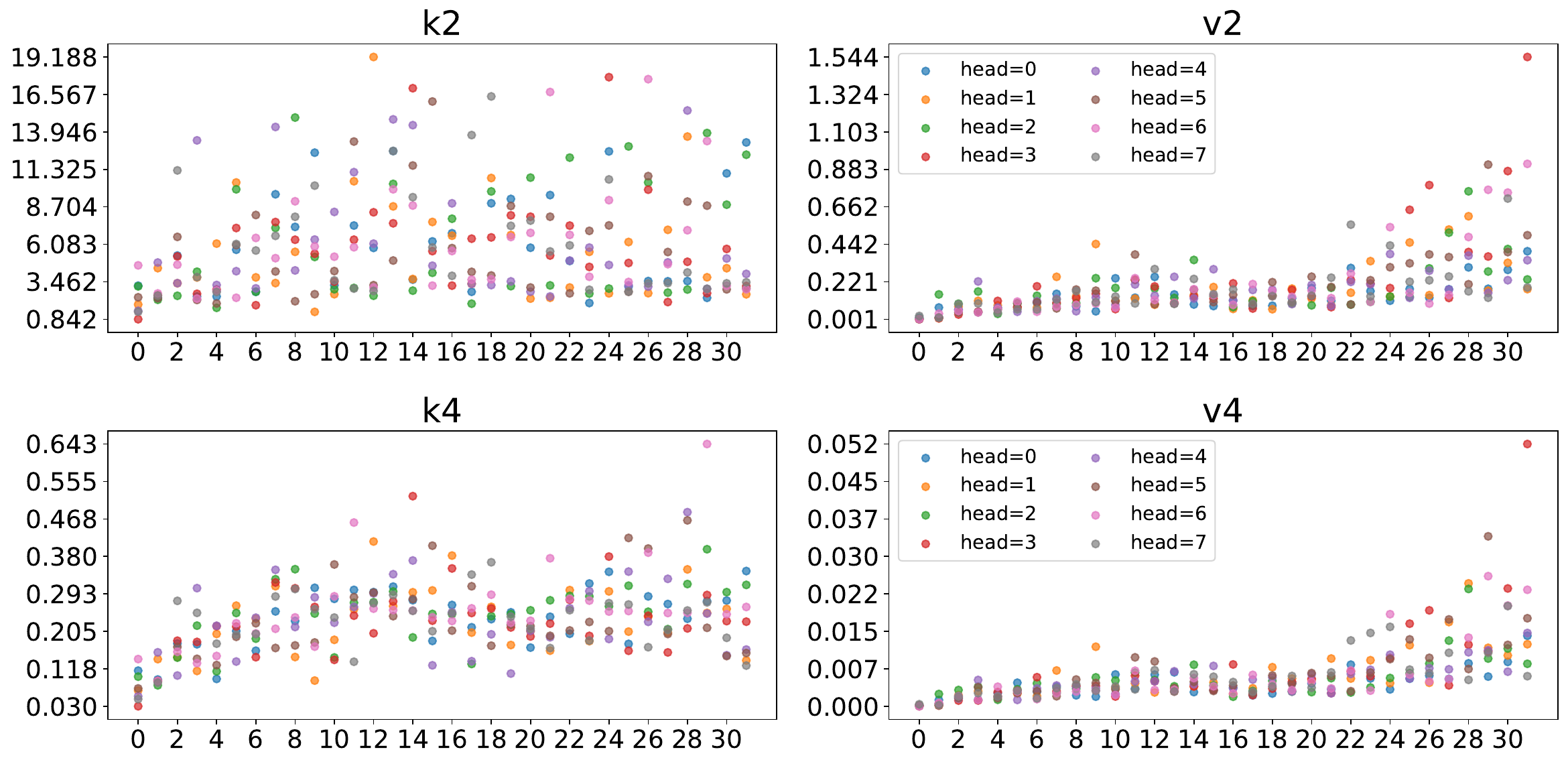}
        \caption{MMLU dataset}
        \label{fig:mmlu-plot}
    \end{subfigure}
    \begin{subfigure}{.9\textwidth}
        \centering
        \includegraphics[width=\linewidth]{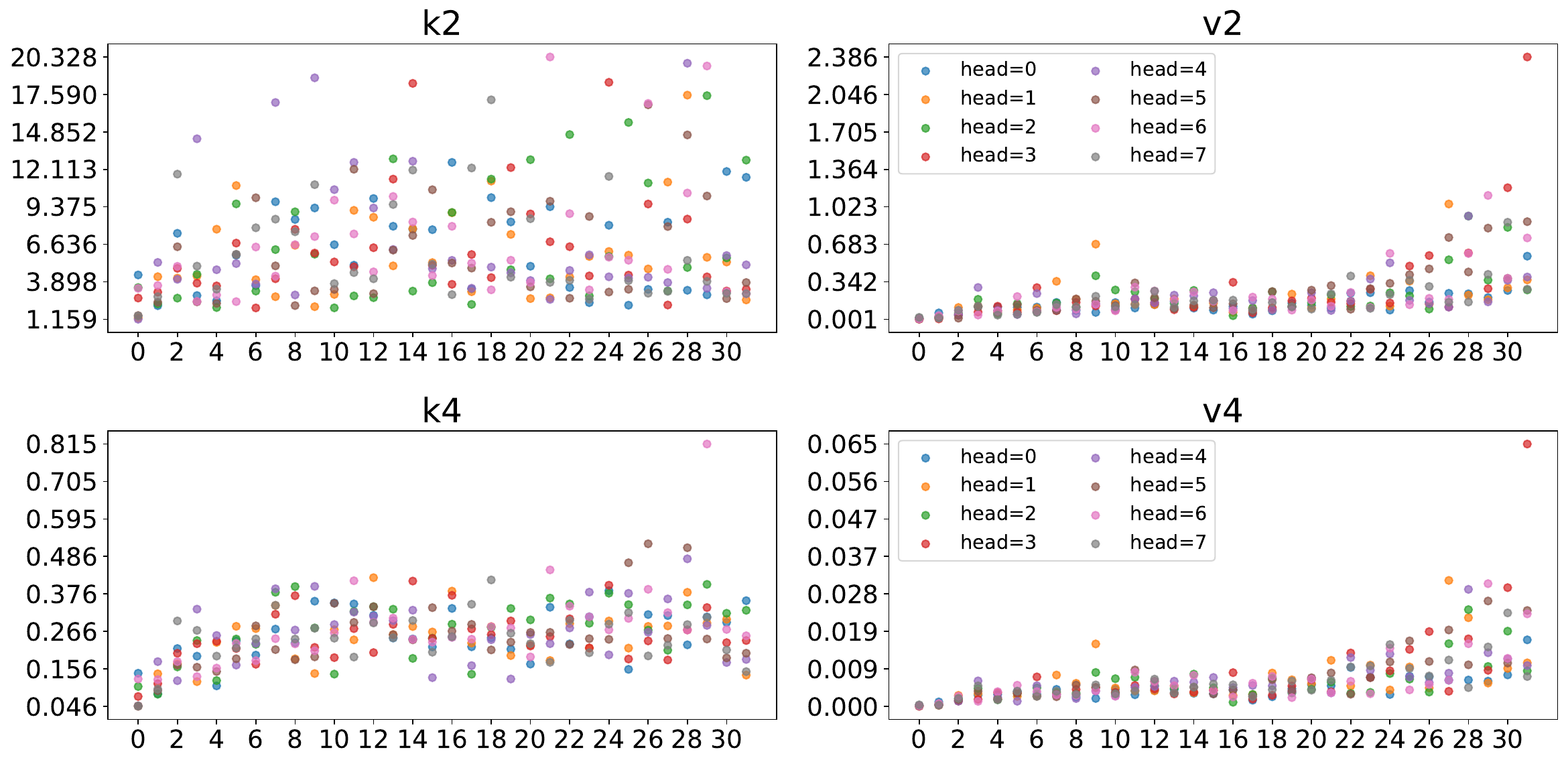}
        \caption{GSM8k dataset}
        \label{fig:gsm8k-plot}
    \end{subfigure}
    \caption{Quantization error (MSE) of K and V caches in the Llama 3.1 8B model across 32 layers for the (a)~C4, (b)~MMLU, and (c)~GSM8k datasets. 
    The top row in each plot shows 2-bit quantization (k2, v2), while the bottom row shows 4-bit quantization (k4, v4). 
    Each point represents a different attention head in the corresponding layer. 
    The x-axis indicates the layer index, and the y-axis shows the quantization error in MSE.}
    \label{fig:layer-head-err}
\end{figure*}

\subsection{Downstream Accuracy Across Quantization Precisions}
\label{sec:downstream_accuracy_full_hqq}

Comprehensive downstream accuracy results for KV cache quantization using the HQQ backend are provided for CoQA (F1 scores), EQ-Parseable (parseable-pass rates and exact-match scores), and GSM8K (flexible and strict accuracy) in Tables~\ref{tab:coqa_accuracies_f1_full}, ~\ref{tab:eqbench_parseable_full}, ~\ref{tab:gsm8k_accuracies_flexible_full}, and~\ref{tab:gsm8k_accuracies_strict_full}, respectively.

\begin{table*}[htbp]
\centering
\setlength{\tabcolsep}{5pt}
\caption{\textbf{Downstream Accuracy on \textsc{CoQA} (Word-Overlap F1) Across Quantization Precisions}. (K\textsubscript{x}V\textsubscript{y}) denotes x-bit keys and y-bit values; higher is better. Results show the keys are the bottleneck while values can be compressed more aggressively: moving from 1-bit to 2-bit \textbf{keys} with \textbf{values fixed at 1-bit} yields large gains (e.g., Qwen3-32B: 0.231~$\rightarrow$~0.732; Llama-3.2-3B: 0.128~$\rightarrow$~0.577), whereas at \textbf{fixed keys $\geq$4-bit}, sweeping values from 1$\rightarrow$8 bits changes F1 only marginally (e.g., Qwen3-8B (K\textsubscript{6}V\textsubscript{1}) 0.813 vs.\ (K\textsubscript{6}V\textsubscript{8}) 0.819; Qwen3-32B (K\textsubscript{8}V\textsubscript{1}) 0.818 vs.\ (K\textsubscript{8}V\textsubscript{8}) 0.827). With sufficiently precise keys (6-8 bits), even \textbf{2-bit values} nearly match BF16: Llama-3.2-1B (K\textsubscript{6}V\textsubscript{2}) 0.700 vs.\ BF16 0.701; Qwen3-32B (K\textsubscript{6}V\textsubscript{2}) 0.826 vs.\ BF16 0.826. In contrast, raising \textbf{values} at 1-bit keys barely helps (e.g., Qwen3-4B (K\textsubscript{1}V\textsubscript{2}) 0.361 vs.\ (K\textsubscript{1}V\textsubscript{8}) 0.364).}
\label{tab:coqa_accuracies_f1_full}
\begin{tabular}{lccccccc}
\toprule
& \multicolumn{1}{c}{\textbf{Qwen3}}
& \multicolumn{1}{c}{\textbf{Llama-3.2}}
& \multicolumn{1}{c}{\textbf{Llama-3.2}}
& \multicolumn{1}{c}{\textbf{Qwen3}}
& \multicolumn{1}{c}{\textbf{Llama-3.1}}
& \multicolumn{1}{c}{\textbf{Qwen3}}
& \multicolumn{1}{c}{\textbf{Qwen3}} \\
& 0.6B & 1B & 3B & 4B & 8B & 8B & 32B \\
\midrule
\(\mathbf{K}_1\mathbf{V}_1\) & 0.130 & 0.148 & 0.128 & 0.222 & 0.127 & 0.359 & 0.231 \\
\(\mathbf{K}_1\mathbf{V}_2\) & 0.227 & 0.236 & 0.223 & 0.361 & 0.207 & 0.384 & 0.328 \\
\(\mathbf{K}_1\mathbf{V}_4\) & 0.226 & 0.237 & 0.229 & 0.376 & 0.242 & 0.386 & 0.349 \\
\(\mathbf{K}_1\mathbf{V}_6\) & 0.215 & 0.243 & 0.230 & 0.379 & 0.221 & 0.379 & 0.352 \\
\(\mathbf{K}_1\mathbf{V}_8\) & 0.220 & 0.240 & 0.222 & 0.364 & 0.242 & 0.382 & 0.355 \\
\(\mathbf{K}_2\mathbf{V}_1\) & 0.253 & 0.190 & 0.577 & 0.464 & 0.565 & 0.599 & 0.732 \\
\(\mathbf{K}_2\mathbf{V}_2\) & 0.334 & 0.526 & 0.760 & 0.721 & 0.757 & 0.767 & 0.809 \\
\(\mathbf{K}_2\mathbf{V}_4\) & 0.333 & 0.559 & 0.766 & 0.732 & 0.763 & 0.761 & 0.809 \\
\(\mathbf{K}_2\mathbf{V}_6\) & 0.340 & 0.565 & 0.764 & 0.730 & 0.764 & 0.766 & 0.807 \\
\(\mathbf{K}_2\mathbf{V}_8\) & 0.332 & 0.568 & 0.764 & 0.733 & 0.766 & 0.770 & 0.804 \\
\(\mathbf{K}_4\mathbf{V}_1\) & 0.497 & 0.575 & 0.769 & 0.803 & 0.773 & 0.815 & 0.818 \\
\(\mathbf{K}_4\mathbf{V}_2\) & 0.666 & 0.693 & 0.797 & 0.806 & 0.786 & 0.822 & 0.820 \\
\(\mathbf{K}_4\mathbf{V}_4\) & 0.692 & 0.701 & 0.798 & 0.807 & 0.788 & 0.819 & 0.825 \\
\(\mathbf{K}_4\mathbf{V}_6\) & 0.688 & 0.702 & 0.797 & 0.809 & 0.784 & 0.817 & 0.824 \\
\(\mathbf{K}_4\mathbf{V}_8\) & 0.687 & 0.700 & 0.797 & 0.807 & 0.782 & 0.818 & 0.823 \\
\(\mathbf{K}_6\mathbf{V}_1\) & 0.650 & 0.597 & 0.771 & 0.801 & 0.765 & 0.813 & 0.817 \\
\(\mathbf{K}_6\mathbf{V}_2\) & 0.693 & 0.700 & 0.798 & 0.803 & 0.785 & 0.821 & 0.826 \\
\(\mathbf{K}_6\mathbf{V}_4\) & 0.702 & 0.705 & 0.796 & 0.807 & 0.786 & 0.821 & 0.825 \\
\(\mathbf{K}_6\mathbf{V}_6\) & 0.701 & 0.706 & 0.795 & 0.808 & 0.790 & 0.819 & 0.825 \\
\(\mathbf{K}_6\mathbf{V}_8\) & 0.694 & 0.703 & 0.796 & 0.807 & 0.789 & 0.819 & 0.825 \\
\(\mathbf{K}_8\mathbf{V}_1\) & 0.651 & 0.599 & 0.769 & 0.802 & 0.767 & 0.815 & 0.818 \\
\(\mathbf{K}_8\mathbf{V}_2\) & 0.691 & 0.696 & 0.798 & 0.803 & 0.788 & 0.823 & 0.825 \\
\(\mathbf{K}_8\mathbf{V}_4\) & 0.701 & 0.704 & 0.795 & 0.807 & 0.789 & 0.821 & 0.826 \\
\(\mathbf{K}_8\mathbf{V}_6\) & 0.699 & 0.705 & 0.795 & 0.809 & 0.787 & 0.819 & 0.826 \\
\(\mathbf{K}_8\mathbf{V}_8\) & 0.697 & 0.704 & 0.795 & 0.808 & 0.786 & 0.821 & 0.827 \\
\textbf{BF16} & 0.708 & 0.701 & 0.795 & 0.808 & 0.785 & 0.820 & 0.826 \\
\bottomrule
\end{tabular}
\end{table*}

\begin{table*}[htbp]
\centering
\setlength{\tabcolsep}{5pt}
\caption{\textbf{Downstream Accuracy on \textsc{EQ-Bench parseable} Across Quantization Precisions}. (K\textsubscript{x}V\textsubscript{y}) denotes x-bit keys and y-bit values; the higher the value, the better. \emph{Parseable accuracy} is the share of prompts where the model's output follows the required structured format so the scorer can automatically extract the four 0-10 emotion ratings. Results consistently show that keys are the bottleneck, while values can be compressed more aggressively. With 1-bit keys, outputs are never parseable across all models regardless of value precision ((K\textsubscript{1}V\textsubscript{y}=0) everywhere). Upgrading to \textbf{2-bit keys} yields large jumps even with very low-precision values. For instance, Llama-3.2-3B rises from 0 to \textbf{80.702} at (K\textsubscript{2}V\textsubscript{2}), Llama-3.1-8B from 0 to \textbf{85.965}, and Qwen3-32B from 0 to \textbf{67.836}, while further increasing \textbf{values} from 2$\rightarrow$8 bits at fixed (K{=}2) brings only modest gains (e.g., Llama-3.2-3B \textbf{80.702$\rightarrow$84.211}, Llama-3.1-8B \textbf{85.965$\rightarrow$90.059}, Qwen3-8B \textbf{65.497$\rightarrow$68.421}). Once \textbf{keys are $\geq$4-6 bits}, even \textbf{1-2-bit values} nearly saturate parseability and often match or exceed BF16 (e.g., Qwen3-8B (K\textsubscript{4}V\textsubscript{1}) \textbf{98.830} vs.\ BF16 \textbf{94.737}; Qwen3-32B (K\textsubscript{6}V\textsubscript{1}) \textbf{79.532} $\approx$ BF16 \textbf{79.532}; Llama-3.2-1B (K\textsubscript{6}V\textsubscript{2}) \textbf{97.076} vs.\ BF16 \textbf{97.661}).}
\label{tab:eqbench_parseable_full}
\begin{tabular}{lccccccc}
\toprule
& \multicolumn{1}{c}{\textbf{Qwen3}} 
& \multicolumn{1}{c}{\textbf{Llama-3.2}} 
& \multicolumn{1}{c}{\textbf{Llama-3.2}} 
& \multicolumn{1}{c}{\textbf{Qwen3}} 
& \multicolumn{1}{c}{\textbf{Llama-3.1}} 
& \multicolumn{1}{c}{\textbf{Qwen3}} 
& \multicolumn{1}{c}{\textbf{Qwen3}} \\
& 0.6B & 1B & 3B & 4B & 8B & 8B & 32B \\
\midrule
\(\mathbf{K}_1\mathbf{V}_1\) & 0 & 0 & 0 & 0 & 0 & 0 & 0 \\
\(\mathbf{K}_1\mathbf{V}_2\) & 0 & 0 & 0 & 0 & 0 & 0 & 0 \\
\(\mathbf{K}_1\mathbf{V}_4\) & 0 & 0 & 0 & 0 & 0 & 0 & 0 \\
\(\mathbf{K}_1\mathbf{V}_6\) & 0 & 0 & 0 & 0 & 0 & 0 & 0 \\
\(\mathbf{K}_1\mathbf{V}_8\) & 0 & 0 & 0 & 0 & 0 & 0 & 0 \\
\(\mathbf{K}_2\mathbf{V}_1\) & 0 & 0 & 8.187 & 0 & 35.673 & 0 & 2.339 \\
\(\mathbf{K}_2\mathbf{V}_2\) & 0 & 16.374 & 80.702 & 27.485 & 85.965 & 65.497 & 67.836 \\
\(\mathbf{K}_2\mathbf{V}_4\) & 0 & 34.503 & 84.795 & 39.766 & 88.889 & 70.175 & 70.175 \\
\(\mathbf{K}_2\mathbf{V}_6\) & 0 & 39.181 & 83.041 & 36.842 & 90.059 & 67.836 & 69.591 \\
\(\mathbf{K}_2\mathbf{V}_8\) & 0 & 38.012 & 84.211 & 40.936 & 88.889 & 68.421 & 70.760 \\
\(\mathbf{K}_4\mathbf{V}_1\) & 0 & 47.953 & 58.480 & 77.778 & 83.626 & 98.830 & 77.778 \\
\(\mathbf{K}_4\mathbf{V}_2\) & 63.743 & 97.076 & 95.322 & 87.719 & 97.661 & 95.322 & 80.702 \\
\(\mathbf{K}_4\mathbf{V}_4\) & 65.497 & 97.076 & 98.830 & 84.795 & 99.415 & 95.322 & 80.702 \\
\(\mathbf{K}_4\mathbf{V}_6\) & 64.328 & 97.076 & 98.830 & 85.380 & 98.830 & 93.567 & 80.702 \\
\(\mathbf{K}_4\mathbf{V}_8\) & 62.573 & 97.076 & 98.830 & 84.795 & 98.830 & 93.567 & 80.702 \\
\(\mathbf{K}_6\mathbf{V}_1\) & 23.392 & 60.234 & 53.801 & 81.287 & 90.643 & 96.491 & 79.532 \\
\(\mathbf{K}_6\mathbf{V}_2\) & 99.415 & 97.076 & 94.737 & 87.719 & 97.661 & 94.737 & 80.702 \\
\(\mathbf{K}_6\mathbf{V}_4\) & 99.415 & 97.661 & 99.415 & 87.135 & 98.830 & 95.906 & 80.702 \\
\(\mathbf{K}_6\mathbf{V}_6\) & 99.415 & 97.661 & 98.830 & 86.550 & 98.830 & 96.491 & 80.702 \\
\(\mathbf{K}_6\mathbf{V}_8\) & 99.415 & 97.661 & 98.830 & 87.135 & 98.830 & 97.661 & 80.702 \\
\(\mathbf{K}_8\mathbf{V}_1\) & 26.901 & 60.234 & 56.725 & 78.947 & 92.983 & 97.076 & 77.193 \\
\(\mathbf{K}_8\mathbf{V}_2\) & 99.415 & 97.076 & 94.737 & 87.135 & 97.661 & 96.491 & 79.532 \\
\(\mathbf{K}_8\mathbf{V}_4\) & 99.415 & 97.661 & 99.415 & 87.719 & 98.830 & 95.322 & 80.702 \\
\(\mathbf{K}_8\mathbf{V}_6\) & 99.415 & 97.661 & 99.415 & 86.550 & 98.830 & 95.906 & 80.702 \\
\(\mathbf{K}_8\mathbf{V}_8\) & 99.415 & 97.661 & 99.415 & 86.550 & 98.830 & 95.906 & 80.702 \\
\textbf{BF16} & 100 & 97.661 & 99.415 & 87.719 & 98.830 & 94.737 & 79.532 \\
\bottomrule
\end{tabular}
\end{table*}

\begin{table*}[htbp]
\centering
\setlength{\tabcolsep}{5pt}
\caption{\textbf{Downstream Accuracy on \textsc{GSM8K (Flexible)} Across Quantization Precisions}. (K\textsubscript{x}V\textsubscript{y}) denotes x-bit keys and y-bit values; the higher the value, the better. \emph{Flexible accuracy} counts a prediction as correct if the gold final numeric answer appears anywhere in the model’s output (ignoring extra formatting), rather than requiring an isolated exact-match box. Results consistently show that keys are the bottleneck, while values can be compressed more aggressively. With 1-bit keys, performance is essentially zero across all models (max $=0.028$). Upgrading to \textbf{2-bit keys} yields large jumps even with low-precision values—for example, Llama-3.2-3B rises from (0.015) at (K\textsubscript{1}V\textsubscript{2}) to $\mathbf{0.278}$ at (K\textsubscript{2}V\textsubscript{2}), Llama-3.1-8B from (0.020) to $\mathbf{0.385}$, and Qwen3-32B from (0.018) to $\mathbf{0.385}$. At fixed \textbf{K($\ge$6)}, even \textbf{2-bit values} already match or beat BF16 (e.g., Qwen3-8B (K\textsubscript{6}V\textsubscript{2}) $\mathbf{0.880}$ vs.\ BF16 $\mathbf{0.877}$; Llama-3.1-8B $\mathbf{0.762}$ vs.\ $\mathbf{0.760}$; Qwen3-4B $\mathbf{0.830}$ vs.\ $\mathbf{0.845}$), and increasing values from (2$\rightarrow$8) bits yields only modest gains (Llama-3.2-3B $\mathbf{0.644}$($\rightarrow$)$\mathbf{0.665}$, Qwen3-8B $\mathbf{0.880}$($\rightarrow$)$\mathbf{0.886}$). In several cases, quantized caches even \emph{exceed} BF16 (e.g., Qwen3-8B (K\textsubscript{6}V\textsubscript{8}) $\mathbf{0.886}$ $>$ 0.877; Llama-3.2-3B (K\textsubscript{4}V\textsubscript{6}) $\mathbf{0.668}$ $>$ 0.663; Qwen3-32B (K\textsubscript{4}V\textsubscript{1}) $\mathbf{0.721}$ $>$ 0.603).}
\label{tab:gsm8k_accuracies_flexible_full}
\begin{tabular}{lccccccc}
\toprule
& \multicolumn{1}{c}{\textbf{Qwen3}}
& \multicolumn{1}{c}{\textbf{Llama-3.2}}
& \multicolumn{1}{c}{\textbf{Llama-3.2}}
& \multicolumn{1}{c}{\textbf{Qwen3}}
& \multicolumn{1}{c}{\textbf{Llama-3.1}}
& \multicolumn{1}{c}{\textbf{Qwen3}}
& \multicolumn{1}{c}{\textbf{Qwen3}} \\
& 0.6B & 1B & 3B & 4B & 8B & 8B & 32B \\
\midrule
\(\mathbf{K}_1\mathbf{V}_1\) & 0.011 & 0.012 & 0.016 & 0.010 & 0.016 & 0.011 & 0.020 \\
\(\mathbf{K}_1\mathbf{V}_2\) & 0.008 & 0.014 & 0.015 & 0.010 & 0.020 & 0.013 & 0.018 \\
\(\mathbf{K}_1\mathbf{V}_4\) & 0.009 & 0.020 & 0.025 & 0.014 & 0.020 & 0.008 & 0.011 \\
\(\mathbf{K}_1\mathbf{V}_6\) & 0.010 & 0.011 & 0.027 & 0.017 & 0.020 & 0.008 & 0.016 \\
\(\mathbf{K}_1\mathbf{V}_8\) & 0.008 & 0.015 & 0.028 & 0.009 & 0.015 & 0.014 & 0.011 \\
\(\mathbf{K}_2\mathbf{V}_1\) & 0.003 & 0.018 & 0.015 & 0.008 & 0.050 & 0.013 & 0.014 \\
\(\mathbf{K}_2\mathbf{V}_2\) & 0.003 & 0.029 & 0.278 & 0.027 & 0.385 & 0.129 & 0.385 \\
\(\mathbf{K}_2\mathbf{V}_4\) & 0.003 & 0.020 & 0.351 & 0.043 & 0.455 & 0.171 & 0.434 \\
\(\mathbf{K}_2\mathbf{V}_6\) & 0.004 & 0.026 & 0.344 & 0.040 & 0.446 & 0.169 & 0.419 \\
\(\mathbf{K}_2\mathbf{V}_8\) & 0.004 & 0.028 & 0.355 & 0.035 & 0.460 & 0.167 & 0.449 \\
\(\mathbf{K}_4\mathbf{V}_1\) & 0.023 & 0.073 & 0.281 & 0.577 & 0.517 & 0.779 & 0.721 \\
\(\mathbf{K}_4\mathbf{V}_2\) & 0.040 & 0.294 & 0.639 & 0.808 & 0.757 & 0.877 & 0.649 \\
\(\mathbf{K}_4\mathbf{V}_4\) & 0.040 & 0.333 & 0.653 & 0.830 & 0.778 & 0.877 & 0.629 \\
\(\mathbf{K}_4\mathbf{V}_6\) & 0.055 & 0.332 & 0.668 & 0.832 & 0.763 & 0.881 & 0.619 \\
\(\mathbf{K}_4\mathbf{V}_8\) & 0.042 & 0.341 & 0.658 & 0.826 & 0.758 & 0.878 & 0.632 \\
\(\mathbf{K}_6\mathbf{V}_1\) & 0.093 & 0.096 & 0.308 & 0.632 & 0.522 & 0.795 & 0.701 \\
\(\mathbf{K}_6\mathbf{V}_2\) & 0.368 & 0.312 & 0.644 & 0.830 & 0.762 & 0.880 & 0.598 \\
\(\mathbf{K}_6\mathbf{V}_4\) & 0.405 & 0.334 & 0.654 & 0.837 & 0.770 & 0.878 & 0.610 \\
\(\mathbf{K}_6\mathbf{V}_6\) & 0.420 & 0.337 & 0.663 & 0.844 & 0.772 & 0.884 & 0.596 \\
\(\mathbf{K}_6\mathbf{V}_8\) & 0.414 & 0.342 & 0.665 & 0.841 & 0.771 & 0.886 & 0.608 \\
\(\mathbf{K}_8\mathbf{V}_1\) & 0.088 & 0.096 & 0.304 & 0.637 & 0.530 & 0.798 & 0.708 \\
\(\mathbf{K}_8\mathbf{V}_2\) & 0.390 & 0.313 & 0.647 & 0.826 & 0.750 & 0.882 & 0.606 \\
\(\mathbf{K}_8\mathbf{V}_4\) & 0.416 & 0.340 & 0.660 & 0.832 & 0.770 & 0.875 & 0.621 \\
\(\mathbf{K}_8\mathbf{V}_6\) & 0.419 & 0.342 & 0.667 & 0.843 & 0.767 & 0.880 & 0.608 \\
\(\mathbf{K}_8\mathbf{V}_8\) & 0.413 & 0.334 & 0.666 & 0.843 & 0.759 & 0.880 & 0.603 \\
\textbf{BF16} & 0.412 & 0.328 & 0.663 & 0.845 & 0.760 & 0.877 & 0.603 \\
\bottomrule
\end{tabular}
\end{table*}

\begin{table*}[htbp]
\centering
\setlength{\tabcolsep}{5pt}
\caption{\textbf{Downstream Accuracy on \textsc{GSM8K (Strict)} Across Quantization Precisions}. (K\textsubscript{x}V\textsubscript{y}) denotes x-bit keys and y-bit values; higher is better. Same dataset as \autoref{tab:gsm8k_accuracies_flexible_full}, but \emph{Strict accuracy} only counts a prediction as correct when the scorer can extract a single final numeric answer that exactly matches the gold. As with the Flexible metric, keys are the bottleneck, while values can be compressed more aggressively. With 1-bit keys, accuracy is essentially zero across all models (max $=0.002$). Upgrading to \textbf{2-bit keys} yields large jumps even with low-precision values—for example, Llama-3.2-3B rises from (0) to $\mathbf{0.276}$ at (K\textsubscript{2}V\textsubscript{2}), Llama-3.1-8B from (0) to $\mathbf{0.380}$, and Qwen3-32B from (0) to $\mathbf{0.234}$. Once \textbf{keys are $\mathbf{\ge}$4-6 bits}, even \textbf{2-bit values} are near or above BF16 (e.g., Qwen3-8B (K\textsubscript{6}V\textsubscript{2}) $\mathbf{0.879}$ vs.\ BF16 $\mathbf{0.874}$; Llama-3.2-3B (K\textsubscript{8}V\textsubscript{6}) $\mathbf{0.661}$ $>$ $\mathbf{0.657}$; Qwen3-32B (K\textsubscript{8}V\textsubscript{6}) $\mathbf{0.723}$ $>$ $\mathbf{0.718}$). At fixed high key precision, increasing values from (2$\rightarrow$8) bits brings only modest gains (e.g., Qwen3-8B at (K$=6$): $\mathbf{0.879}$($\rightarrow$)$\mathbf{0.885}$; Llama-3.2-3B at (K$=6$): $\mathbf{0.639}$($\rightarrow$)$\mathbf{0.657}$).}

\label{tab:gsm8k_accuracies_strict_full}

\begin{tabular}{lccccccc}
\toprule
& \multicolumn{1}{c}{\textbf{Qwen3}}
& \multicolumn{1}{c}{\textbf{Llama-3.2}}
& \multicolumn{1}{c}{\textbf{Llama-3.2}}
& \multicolumn{1}{c}{\textbf{Qwen3}}
& \multicolumn{1}{c}{\textbf{Llama-3.1}}
& \multicolumn{1}{c}{\textbf{Qwen3}}
& \multicolumn{1}{c}{\textbf{Qwen3}} \\
& 0.6B & 1B & 3B & 4B & 8B & 8B & 32B \\
\midrule
\(\mathbf{K}_1\mathbf{V}_1\) & 0.000 & 0.000 & 0.000 & 0.000 & 0.000 & 0.000 & 0.000 \\
\(\mathbf{K}_1\mathbf{V}_2\) & 0.000 & 0.000 & 0.000 & 0.000 & 0.000 & 0.000 & 0.000 \\
\(\mathbf{K}_1\mathbf{V}_4\) & 0.000 & 0.000 & 0.000 & 0.000 & 0.000 & 0.000 & 0.000 \\
\(\mathbf{K}_1\mathbf{V}_6\) & 0.000 & 0.000 & 0.002 & 0.000 & 0.000 & 0.001 & 0.000 \\
\(\mathbf{K}_1\mathbf{V}_8\) & 0.000 & 0.000 & 0.000 & 0.000 & 0.000 & 0.000 & 0.000 \\
\(\mathbf{K}_2\mathbf{V}_1\) & 0.000 & 0.001 & 0.005 & 0.000 & 0.028 & 0.000 & 0.002 \\
\(\mathbf{K}_2\mathbf{V}_2\) & 0.000 & 0.011 & 0.276 & 0.004 & 0.380 & 0.067 & 0.234 \\
\(\mathbf{K}_2\mathbf{V}_4\) & 0.000 & 0.010 & 0.350 & 0.009 & 0.444 & 0.104 & 0.334 \\
\(\mathbf{K}_2\mathbf{V}_6\) & 0.000 & 0.012 & 0.344 & 0.010 & 0.439 & 0.115 & 0.330 \\
\(\mathbf{K}_2\mathbf{V}_8\) & 0.000 & 0.014 & 0.353 & 0.005 & 0.449 & 0.109 & 0.347 \\
\(\mathbf{K}_4\mathbf{V}_1\) & 0.006 & 0.061 & 0.293 & 0.647 & 0.500 & 0.785 & 0.701 \\
\(\mathbf{K}_4\mathbf{V}_2\) & 0.042 & 0.297 & 0.628 & 0.825 & 0.751 & 0.876 & 0.726 \\
\(\mathbf{K}_4\mathbf{V}_4\) & 0.044 & 0.331 & 0.643 & 0.844 & 0.759 & 0.876 & 0.734 \\
\(\mathbf{K}_4\mathbf{V}_6\) & 0.055 & 0.331 & 0.657 & 0.839 & 0.748 & 0.880 & 0.733 \\
\(\mathbf{K}_4\mathbf{V}_8\) & 0.049 & 0.342 & 0.649 & 0.835 & 0.745 & 0.877 & 0.737 \\
\(\mathbf{K}_6\mathbf{V}_1\) & 0.083 & 0.083 & 0.326 & 0.698 & 0.506 & 0.803 & 0.702 \\
\(\mathbf{K}_6\mathbf{V}_2\) & 0.369 & 0.308 & 0.639 & 0.839 & 0.750 & 0.879 & 0.709 \\
\(\mathbf{K}_6\mathbf{V}_4\) & 0.405 & 0.334 & 0.644 & 0.845 & 0.756 & 0.876 & 0.719 \\
\(\mathbf{K}_6\mathbf{V}_6\) & 0.422 & 0.335 & 0.656 & 0.853 & 0.751 & 0.882 & 0.717 \\
\(\mathbf{K}_6\mathbf{V}_8\) & 0.414 & 0.341 & 0.657 & 0.848 & 0.753 & 0.885 & 0.721 \\
\(\mathbf{K}_8\mathbf{V}_1\) & 0.085 & 0.084 & 0.324 & 0.704 & 0.513 & 0.805 & 0.695 \\
\(\mathbf{K}_8\mathbf{V}_2\) & 0.388 & 0.314 & 0.640 & 0.829 & 0.739 & 0.882 & 0.717 \\
\(\mathbf{K}_8\mathbf{V}_4\) & 0.417 & 0.340 & 0.652 & 0.845 & 0.756 & 0.875 & 0.721 \\
\(\mathbf{K}_8\mathbf{V}_6\) & 0.419 & 0.340 & 0.661 & 0.848 & 0.749 & 0.879 & 0.723 \\
\(\mathbf{K}_8\mathbf{V}_8\) & 0.415 & 0.332 & 0.660 & 0.851 & 0.744 & 0.876 & 0.723 \\
\textbf{BF16} & 0.413 & 0.328 & 0.657 & 0.852 & 0.741 & 0.874 & 0.718 \\
\bottomrule
\end{tabular}
\end{table*}

\subsection{Rotation and Grouping Results}
\label{sec:rotation_group_results}

Tables~\ref{tab:rotation-gsm8k}-\ref{tab:rotation-coqa-em} 
present the full set of downstream evaluation results for integrating rotation-based outlier redistribution with mixed-precision KV quantization across multiple models and tasks, including \textsc{GSM8K} (flexible and strict exact match), \textsc{CoQA} (F1), and \textsc{CoQA} (exact match). Each table reports accuracy for four key-value bit allocations (\(\mathbf{K}_2\mathbf{V}_2\), \(\mathbf{K}_2\mathbf{V}_4\), \(\mathbf{K}_4\mathbf{V}_2\), \(\mathbf{K}_4\mathbf{V}_4\)) under four Rotations: none, key-only, value-only, and both key and value. The group size is fixed at 64 for both keys and values in these experiments to isolate the effect of rotation.

The results reveal several consistent trends across model scales and tasks. First, applying rotation to \emph{keys} consistently improves performance at lower bit-widths (\(\mathrm{K}_2\mathrm{V}_2\) and \(\mathrm{K}_2\mathrm{V}_4\)), significantly narrowing the gap to higher-precision baselines. In contrast, rotating \emph{values} alone yields marginal or even negative effects, especially at low precisions. Notably, applying rotation to \emph{keys only} under the \(\mathrm{K}_4\mathrm{V}_2\) configuration achieves accuracy that closely matches the full \(\mathrm{K}_4\mathrm{V}_4\) baseline, indicating that the dominant benefits of rotation stem from mitigating key outliers rather than value outliers. Applying rotation to both keys and values provides little additional gain beyond key-only rotation, reinforcing that keys are the primary target for rotation-based improvements.

\begin{figure*}
    \centering
    \includegraphics[width=\linewidth]{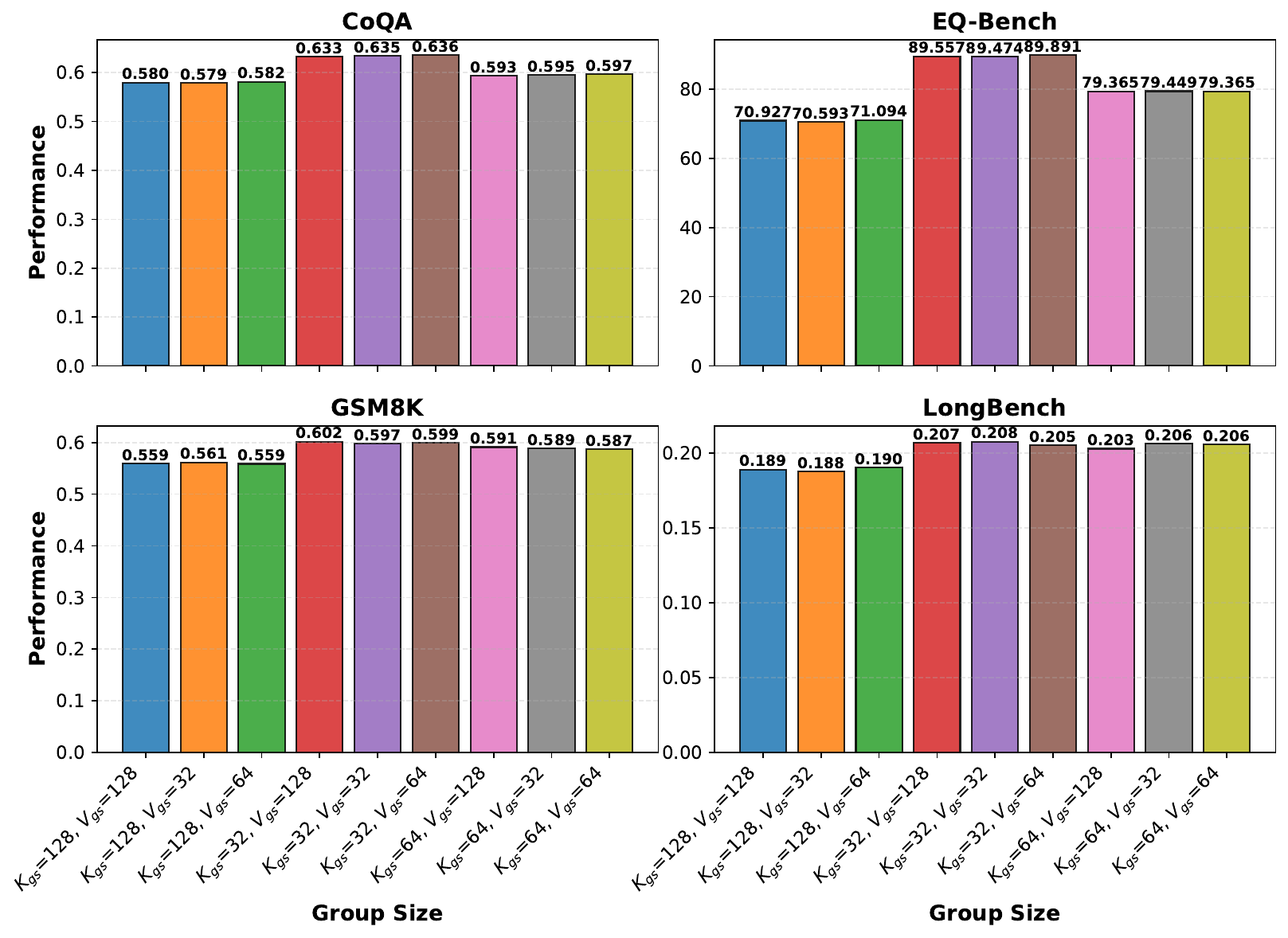}
    \caption{
        \textbf{Effect of group size configurations on downstream accuracy.}
        Each bar shows performance on \textsc{CoQA}, \textsc{GSM8K}, \textsc{EQ-Bench}, and \textsc{LongBench} for different key-value grouping combinations (\( \mathrm{gs}_K, \mathrm{gs}_V \in \{32, 64, 128\}\)) under the \(\mathbf{K}_4\mathbf{V}_2\) mixed-precision setting.
        Smaller group sizes for \textbf{keys} consistently improve accuracy across all tasks, while value group size has a smaller but non-negligible effect.  
        Configurations with \(\mathrm{gs}_K = 32\) achieve the best overall performance, highlighting the benefit of finer quantization granularity for key caches, which are more sensitive to quantization distortion.  
        In contrast, larger group sizes for values (\(\mathrm{gs}_V = 64\) or \(128\)) preserve performance while reducing overhead, aligning with their lower sensitivity.
    }
    \label{fig:group_size_comparison}
\end{figure*}

Figure~\ref{fig:group_size_comparison} examines the effect of \emph{group size configuration} under a fixed \(\mathrm{K}_4\mathrm{V}_2\) mixed-precision setting without rotation. Group size (\(\mathrm{gs}\)) determines the block granularity of quantization: smaller groups offer finer scaling at the cost of increased metadata and computation. The results reveal a clear asymmetry between keys and values. Reducing group size for \emph{keys} consistently improves downstream accuracy across \textsc{CoQA}, \textsc{GSM8K}, \textsc{EQ-Bench}, and \textsc{LongBench}, with \(\mathrm{gs}_K = 32\) achieving the best overall performance. This reflects the higher sensitivity of key caches to quantization distortion. In contrast, increasing value group size to 64 or 128 has a negligible impact on accuracy while reducing overhead, consistent with their lower sensitivity.

Together, these findings provide practical guidance for combining geometry-driven bit allocation with rotation and grouping strategies: rotation should primarily target \emph{keys}, while keys benefit from finer group granularity and higher precision; values, on the other hand, can use coarser grouping and lower precision with minimal accuracy loss.

\begin{table}[htbp]
\centering
\caption{\textbf{Rotation-based Outlier
Redistribution Integration on \textsc{GSM8K} - Exact Match (Flexible).} 
Downstream accuracy under different key-value bit allocations and rotation scopes. Group size is fixed to 64 for both keys and values.}
\label{tab:rotation-gsm8k}
\resizebox{\linewidth}{!}{
\begin{tabular}{llrrrr}
\toprule
\textbf{Model} & \textbf{Rotation} & \(\mathbf{K}_2\mathbf{V}_2\) & \(\mathbf{K}_2\mathbf{V}_4\) & \(\mathbf{K}_4\mathbf{V}_2\) & \(\mathbf{K}_4\mathbf{V}_4\) \\
\midrule
\multirow{4}{*}{\shortstack{Llama\\3.1-8B}} 
 & K+V   & 52.99 & 55.80 & 76.50 & 76.95 \\
 & K only & 52.92 & 57.32 & 75.44 & 76.50 \\
 & V only & 20.70 & 25.70 & 75.28 & 76.95 \\
 & none   & 18.65 & 24.41 & 75.21 & 76.95 \\
\midrule
\multirow{4}{*}{\shortstack{Llama\\3.2-1B}} 
 & K+V   & 1.97 & 2.43 & 30.93 & 33.13 \\
 & K only & 2.05 & 3.03 & 28.43 & 33.13 \\
 & V only & 2.05 & 1.29 & 28.58 & 29.57 \\
 & none   & 1.82 & 2.20 & 27.22 & 29.57 \\
\midrule
\multirow{4}{*}{\shortstack{Llama\\3.2-3B}} 
 & K+V   & 40.11 & 44.12 & 62.85 & 64.37 \\
 & K only & 38.06 & 45.26 & 61.18 & 63.91 \\
 & V only & 16.60 & 19.26 & 63.46 & 63.91 \\
 & none   & 14.86 & 21.08 & 62.93 & 63.53 \\
\midrule
\multirow{4}{*}{\shortstack{Qwen\\0.6B}} 
 & K+V   & 0.76 & 0.61 & 37.83 & 36.47 \\
 & K only & 1.52 & 1.06 & 36.92 & 37.68 \\
 & V only & 1.52 & 1.67 & 1.36 & 2.12 \\
 & none   & 0.83 & 1.90 & 1.36 & 2.05 \\
\midrule
\multirow{4}{*}{\shortstack{Qwen\\32B}} 
 & K+V   & 31.69 & 33.81 & 66.49 & 64.22 \\
 & K only & 30.10 & 33.51 & 63.99 & 65.28 \\
 & V only & 6.90 & 11.07 & 65.81 & 63.68 \\
 & none   & 6.67 & 10.92 & 64.82 & 63.76 \\
\midrule
\multirow{4}{*}{\shortstack{Qwen\\4B}} 
 & K+V   & 0.91 & 0.76 & 83.78 & 85.37 \\
 & K only & 1.14 & 0.68 & 82.26 & 84.76 \\
 & V only & 0.83 & 1.14 & 82.56 & 83.70 \\
 & none   & 1.14 & 0.83 & 81.80 & 83.40 \\
\midrule
\multirow{4}{*}{\shortstack{Qwen\\8B}} 
 & K+V   & 29.34 & 36.01 & 88.17 & 88.17 \\
 & K only & 29.57 & 35.56 & 87.64 & 87.95 \\
 & V only & 1.97 & 1.21 & 86.73 & 87.34 \\
 & none   & 2.27 & 1.67 & 86.58 & 88.02 \\
\bottomrule
\end{tabular}}
\end{table}

\begin{table}[htbp]
\centering
\caption{\textbf{Rotation-based Outlier
Redistribution Integration on \textsc{GSM8K} - Exact Match (Strict).} 
Downstream accuracy under different key-value bit allocations and rotation scopes. Group size is fixed to 64 for both keys and values.}
\label{tab:rotation-gsm8k-strict}
\resizebox{\linewidth}{!}{
\begin{tabular}{llrrrr}
\toprule
\textbf{Model} & \textbf{Rotation} & \(\mathbf{K}_2\mathbf{V}_2\) & \(\mathbf{K}_2\mathbf{V}_4\) & \(\mathbf{K}_4\mathbf{V}_2\) & \(\mathbf{K}_4\mathbf{V}_4\) \\
\midrule
\multirow{4}{*}{\shortstack{Llama\\3.1-8B}}
 & K+V   & 51.71 & 55.27 & 74.37 & 74.45 \\
 & K only & 51.93 & 56.48 & 73.24 & 73.92 \\
 & V only & 19.33 & 24.49 & 73.01 & 74.60 \\
 & none   & 17.97 & 22.74 & 72.93 & 74.60 \\
\midrule
\multirow{4}{*}{\shortstack{Llama\\3.2-1B}}
 & K+V   & 0.83 & 1.21 & 31.01 & 32.98 \\
 & K only & 0.68 & 1.29 & 29.04 & 33.06 \\
 & V only & 0.38 & 0.61 & 28.51 & 29.57 \\
 & none   & 0.45 & 0.61 & 27.22 & 29.34 \\
\midrule
\multirow{4}{*}{\shortstack{Llama\\3.2-3B}}
 & K+V   & 39.65 & 43.67 & 61.94 & 63.84 \\
 & K only & 36.47 & 44.96 & 60.27 & 63.46 \\
 & V only & 16.00 & 19.26 & 63.00 & 63.38 \\
 & none   & 14.40 & 21.00 & 62.02 & 62.85 \\
\midrule
\multirow{4}{*}{\shortstack{Qwen\\0.6B}}
 & K+V   & 0.00 & 0.00 & 38.14 & 36.92 \\
 & K only & 0.00 & 0.00 & 37.38 & 37.83 \\
 & V only & 0.00 & 0.00 & 0.15 & 0.53 \\
 & none   & 0.00 & 0.00 & 0.30 & 0.08 \\
\midrule
\multirow{4}{*}{\shortstack{Qwen\\32B}}
 & K+V   & 21.99 & 28.35 & 74.75 & 75.59 \\
 & K only & 21.46 & 26.38 & 74.75 & 75.28 \\
 & V only & 2.20 & 5.00 & 74.00 & 75.13 \\
 & none   & 1.52 & 4.93 & 73.77 & 74.37 \\
\midrule
\multirow{4}{*}{\shortstack{Qwen\\4B}}
 & K+V   & 0.23 & 0.08 & 84.53 & 85.60 \\
 & K only & 0.00 & 0.08 & 83.40 & 84.99 \\
 & V only & 0.00 & 0.00 & 82.49 & 83.70 \\
 & none   & 0.00 & 0.08 & 81.80 & 83.78 \\
\midrule
\multirow{4}{*}{\shortstack{Qwen\\8B}}
 & K+V   & 25.32 & 30.33 & 87.72 & 87.87 \\
 & K only & 25.17 & 32.15 & 87.49 & 87.34 \\
 & V only & 0.15 & 0.00 & 86.43 & 86.88 \\
 & none   & 0.00 & 0.08 & 86.50 & 87.41 \\
\bottomrule
\end{tabular}}
\end{table}

\begin{table}[htbp]
\centering
\caption{\textbf{Rotation-based Outlier
Redistribution Integration on \textsc{CoQA} - F1 Score.} 
Downstream accuracy under different key-value bit allocations and rotation scopes. Group size is fixed to 64 for both keys and values.}
\label{tab:rotation-coqa-f1}
\resizebox{\linewidth}{!}{
\begin{tabular}{llrrrr}
\toprule
\textbf{Model} & \textbf{Rotation} & \(\mathbf{K}_2\mathbf{V}_2\) & \(\mathbf{K}_2\mathbf{V}_4\) & \(\mathbf{K}_4\mathbf{V}_2\) & \(\mathbf{K}_4\mathbf{V}_4\) \\
\midrule
\multirow{4}{*}{\shortstack{Llama\\3.1-8B}} 
 & K+V   & 77.70 & 77.83 & 78.88 & 78.45 \\
 & K only & 77.39 & 78.10 & 77.93 & 78.75 \\
 & V only & 75.46 & 76.71 & 79.34 & 78.76 \\
 & none   & 74.72 & 76.57 & 78.10 & 79.18 \\
\midrule
\multirow{4}{*}{\shortstack{Llama\\3.2-1B}} 
 & K+V   & 50.66 & 53.55 & 70.44 & 70.47 \\
 & K only & 50.18 & 53.46 & 69.40 & 70.27 \\
 & V only & 39.50 & 44.93 & 69.58 & 70.52 \\
 & none   & 38.82 & 44.33 & 69.53 & 70.31 \\
\midrule
\multirow{4}{*}{\shortstack{Llama\\3.2-3B}} 
 & K+V   & 78.45 & 78.78 & 79.24 & 79.13 \\
 & K only & 78.24 & 78.88 & 78.71 & 79.20 \\
 & V only & 73.25 & 74.48 & 78.86 & 79.27 \\
 & none   & 74.10 & 74.55 & 79.14 & 79.37 \\
\midrule
\multirow{4}{*}{\shortstack{Qwen\\0.6B}} 
 & K+V   & 33.99 & 34.06 & 68.92 & 70.09 \\
 & K only & 33.88 & 33.70 & 69.62 & 70.08 \\
 & V only & 29.00 & 27.39 & 48.19 & 48.06 \\
 & none   & 28.89 & 28.04 & 46.49 & 48.41 \\
\midrule
\multirow{4}{*}{\shortstack{Qwen\\32B}} 
 & K+V   & 64.85 & 67.39 & 82.46 & 82.59 \\
 & K only & 67.16 & 67.18 & 82.62 & 82.55 \\
 & V only & 77.82 & 78.55 & 82.35 & 82.50 \\
 & none   & 76.80 & 78.35 & 82.60 & 82.77 \\
\midrule
\multirow{4}{*}{\shortstack{Qwen\\4B}} 
 & K+V   & 62.43 & 63.82 & 80.16 & 80.56 \\
 & K only & 62.09 & 64.35 & 80.46 & 80.63 \\
 & V only & 56.73 & 59.26 & 80.07 & 80.16 \\
 & none   & 54.93 & 59.68 & 80.41 & 80.19 \\
\midrule
\multirow{4}{*}{\shortstack{Qwen\\8B}} 
 & K+V   & 79.33 & 79.65 & 82.32 & 81.98 \\
 & K only & 78.27 & 79.89 & 81.15 & 82.09 \\
 & V only & 67.51 & 70.41 & 82.23 & 82.01 \\
 & none   & 66.86 & 69.83 & 81.49 & 81.99 \\
\bottomrule
\end{tabular}}
\end{table}

\begin{table}[htbp]
\centering
\caption{\textbf{Rotation-based Outlier
Redistribution Integration on \textsc{CoQA} - Exact Match.} 
Downstream accuracy under different key-value bit allocations and rotation scopes. Group size is fixed to 64 for both keys and values.}
\label{tab:rotation-coqa-em}
\resizebox{\linewidth}{!}{
\begin{tabular}{llrrrr}
\toprule
\textbf{Model} & \textbf{Rotation} & \(\mathbf{K}_2\mathbf{V}_2\) & \(\mathbf{K}_2\mathbf{V}_4\) & \(\mathbf{K}_4\mathbf{V}_2\) & \(\mathbf{K}_4\mathbf{V}_4\) \\
\midrule
\multirow{4}{*}{\shortstack{Llama\\3.1-8B}}
 & K+V   & 62.43 & 63.37 & 63.25 & 63.40 \\
 & K only & 62.12 & 63.32 & 63.03 & 63.65 \\
 & V only & 58.13 & 60.12 & 64.48 & 63.75 \\
 & none   & 58.12 & 59.65 & 62.92 & 64.45 \\
\midrule
\multirow{4}{*}{\shortstack{Llama\\3.2-1B}}
 & K+V   & 33.38 & 37.87 & 54.78 & 56.02 \\
 & K only & 33.72 & 37.30 & 54.22 & 55.32 \\
 & V only & 25.47 & 30.37 & 53.97 & 55.65 \\
 & none   & 24.42 & 30.23 & 54.30 & 55.45 \\
\midrule
\multirow{4}{*}{\shortstack{Llama\\3.2-3B}}
 & K+V   & 62.47 & 62.28 & 62.83 & 63.40 \\
 & K only & 62.12 & 62.72 & 62.93 & 63.53 \\
 & V only & 56.07 & 57.48 & 62.70 & 63.73 \\
 & none   & 56.45 & 57.62 & 63.42 & 63.77 \\
\midrule
\multirow{4}{*}{\shortstack{Qwen\\0.6B}}
 & K+V   & 19.67 & 19.57 & 56.07 & 56.73 \\
 & K only & 18.72 & 18.18 & 55.53 & 56.73 \\
 & V only & 13.97 & 14.27 & 28.18 & 28.17 \\
 & none   & 14.25 & 13.80 & 28.22 & 28.82 \\
\midrule
\multirow{4}{*}{\shortstack{Qwen\\32B}}
 & K+V   & 46.97 & 49.95 & 70.18 & 70.77 \\
 & K only & 50.28 & 49.42 & 70.48 & 70.87 \\
 & V only & 61.77 & 64.08 & 70.02 & 70.37 \\
 & none   & 61.12 & 63.92 & 69.90 & 70.50 \\
\midrule
\multirow{4}{*}{\shortstack{Qwen\\4B}}
 & K+V   & 41.28 & 42.45 & 65.68 & 66.28 \\
 & K only & 41.65 & 43.45 & 66.40 & 66.47 \\
 & V only & 39.22 & 40.03 & 65.43 & 66.20 \\
 & none   & 37.70 & 40.22 & 65.52 & 66.07 \\
\midrule
\multirow{4}{*}{\shortstack{Qwen\\8B}}
 & K+V   & 61.70 & 61.88 & 68.55 & 67.73 \\
 & K only & 60.07 & 62.40 & 66.62 & 68.23 \\
 & V only & 49.62 & 51.58 & 68.10 & 68.07 \\
 & none   & 48.40 & 51.65 & 67.07 & 68.00 \\
\bottomrule
\end{tabular}}
\end{table}

\begin{table*}[htbp]
\centering
\renewcommand{\arraystretch}{1.3}
\small
\caption{\textbf{Performance by rotation strategy across tasks.} 
Values report the mean performance across seven models (0.6B–32B parameters) and four quantization configurations (\(\mathrm{K}_2\mathrm{V}_2\), \(\mathrm{K}_2\mathrm{V}_4\), \(\mathrm{K}_4\mathrm{V}_2\), \(\mathrm{K}_4\mathrm{V}_4\)). 
The $\pm$ values indicate the variability range across model scales and quantization settings, with larger ranges reflecting higher sensitivity to these factors.}

\resizebox{\linewidth}{!}{
\begin{tabular}{llcccc}
\toprule
\textbf{Task} & \textbf{Metric} & \textbf{No Rotation} & \textbf{Value-Only} & \textbf{Key-Only} & \textbf{Both} \\
\midrule
CoQa & Exact Match & $50.10 \pm 17.65$ & $51.82 \pm 17.22$ & $55.34 \pm 14.21$ & $55.39 \pm 14.10$ \\
Eq Bench & Parseable & $56.18 \pm 42.57$ & $55.01 \pm 41.93$ & $65.73 \pm 39.70$ & $64.81 \pm 39.19$ \\
GSM8K & Exact Match & $32.73 \pm 33.42$ & $32.96 \pm 33.48$ & $43.46 \pm 28.98$ & $43.80 \pm 29.28$ \\
Longbench & ROUGE Score & $16.70 \pm 11.75$ & $16.54 \pm 11.82$ & $17.50 \pm 12.83$ & $17.63 \pm 12.81$ \\
\bottomrule
\end{tabular}
}
\label{tab:app:rot}
\end{table*}

\section{Reproducibility and Resources}
\label{appendix:compute_resources}
Inference was performed using the Hugging Face Transformers \cite{wolf-etal-2020-transformers} and Accelerate \cite{accelerate} with FlashAttention \cite{dao2022flashattention, dao2023flashattention2}. We integrated both Quanto and HQQ into the Language Model Evaluation Harness \cite{eval-harness} to enable systematic and reproducible evaluation of model performance under varying quantization schemes. Quantization error is measured by MSE, and confidence intervals use Bayesian variance. \cite{hariri2025don}. Evaluations were executed on two High-performance Computing (HPC) clusters, detailed in Table~\ref{tab:testbeds}.

\begin{table*}[htbp]
\caption{Specifications of Two High-Performance Computing (HPC) Clusters Used in This Study}
\label{tab:testbeds}
\centering
\renewcommand{\arraystretch}{1.2}
\setlength{\tabcolsep}{5pt}        
\begin{tabular}{@{}lcc@{}}
    \toprule
    \textbf{Cluster} & \textbf{Cluster A} & \textbf{Cluster B} \\
    \midrule
    \textbf{Processor}      & AMD EPYC 7742 $\times$ 2          & Intel Xeon Platinum 8468 $\times$ 2 \\
    \textbf{RAM}      & 2048\,GB            & 2048\,GB \\
    \textbf{GPU}            & NVIDIA A100 $\times$ 8            & NVIDIA H200 $\times$ 8 \\
    \textbf{VRAM}            & 80\,GB HBM2e             & 141\,GB HBM3e \\
    \textbf{Scale}          & 5 nodes (40 A100)       & 5 nodes (40 H200) \\
    \bottomrule
\end{tabular}
\end{table*}

\appendix

\pagebreak
\newpage
\clearpage

\end{document}